\documentclass[11pt]{article}
\usepackage[final]{acl}

\usepackage{times}
\usepackage{latexsym}

\usepackage[T1]{fontenc}

\usepackage[utf8]{inputenc}
\usepackage{enumitem}
\usepackage{microtype}

\usepackage{inconsolata}

\usepackage{graphicx}

\usepackage{booktabs}
\usepackage{multirow}
\usepackage{algorithm}
\usepackage{algorithmic}
\usepackage{colortbl}
\usepackage{makecell}
\usepackage{tcolorbox}
\usepackage{rotating}
\usepackage{hhline}
\usepackage{amsmath}
\tcbuselibrary{breakable}
\definecolor{lightestgray}{gray}{0.95}
\usepackage[font=small,labelfont=small]{caption}

\usepackage{xspace}
\newcommand{\datasetName}{{\sc CLAUSE}\xspace}
\usepackage{xfp}

\title{Better Call \datasetName: A Discrepancy Benchmark for Auditing LLMs Legal Reasoning Capabilities}

\author{
\textbf{Manan Roy Choudhury}$^{1}$\thanks{\fontsize{10.5}{12.5}\selectfont The first three authors are recognized for their equal contribution to this work.},
\textbf{Adithya Chandramouli}$^{1}$\footnotemark[1],
\textbf{Mannan Anand}$^{1}$\footnotemark[1],
\textbf{Vivek Gupta}$^{1}$\\[4pt]
$^{1}$Arizona State University\\[2pt]
\texttt{\{mroycho1, achand71, mannan.anand, vgupt140\}@asu.edu}
}

\begin{document}
\maketitle
\begin{abstract}
The rapid integration of large language models (LLMs) into high-stakes legal work has exposed a critical gap: no benchmark exists to systematically stress-test their reliability against the nuanced, adversarial, and often subtle flaws present in real-world contracts. To address this, we introduce CLAUSE, a first-of-its-kind benchmark designed to evaluate the fragility of an LLM's legal reasoning. We study the capabilities of LLMs to detect and reason about fine-grained discrepancies by producing over 7500 real-world perturbed contracts from foundational datasets like CUAD and ContractNLI. Our novel, persona-driven pipeline generates 10 distinct anomaly categories, which are then validated against official statutes using a Retrieval-Augmented Generation (RAG) system to ensure legal fidelity. We use CLAUSE to evaluate leading LLMs’ ability to detect embedded legal flaws and explain their significance. Our analysis shows a key weakness: these models often miss subtle errors and struggle even more to justify them legally. Our work outlines a path to identify and correct such reasoning failures in legal AI.

\end{abstract}

\section{Introduction}

The integration of Large Language Models (LLMs) into high-stakes legal work has outpaced the development of benchmarks that can audit their reliability against nuanced contractual flaws. This gap is critical, as a single undetected defect can have catastrophic consequences. In \textit{Perini Corp. v. Greate Bay Hotel \& Casino, Inc.}, for instance, the simple omission of a consequential damages waiver resulted in a \$14.5 million liability, a sum over twenty times the contract's fee \cite{mdmc-perini-2021}. State-of-the-art models, optimized for fluency, are not trained to detect such legally significant risks, particularly those arising from the absence of text. Existing evaluation methods fail to probe these fragile reasoning capabilities \cite{hu2025multilingual, mishra2025shortcomings}, creating an urgent need for a new class of benchmark to stress-test LLMs against the adversarial and context-aware challenges of legal practice. This research directly confronts the ongoing debate about AI's readiness for legal integration. By evaluating modern LLMs against our \datasetName benchmark, we aim to provide empirical data to help answer the fundamental question: \emph{are these models truly reliable enough for such high-stakes work?}

To answer this question and address this gap by developing \datasetName, a novel benchmark designed to systematically evaluate the fragility of Large Language Models' legal reasoning capabilities. At its core, \datasetName constructs realistic and targeted contradictions by perturbing over 7,500 real-world contracts drawn from foundational sources such as CUAD\cite{hendrycks2021cuad} and ContractNLI\cite{koreeda2021contractnli}. The framework defines ten distinct perturbation categories that span both statutory compliance failures (legal contradictions) and internal document inconsistencies (in-text contradictions). This taxonomy allows researchers to perform fine-grained analysis of how models respond to subtle but consequential changes that could alter contract enforceability or introduce hidden liabilities.

\begin{figure*}[h!]
\centering
\includegraphics[width=0.85\textwidth]{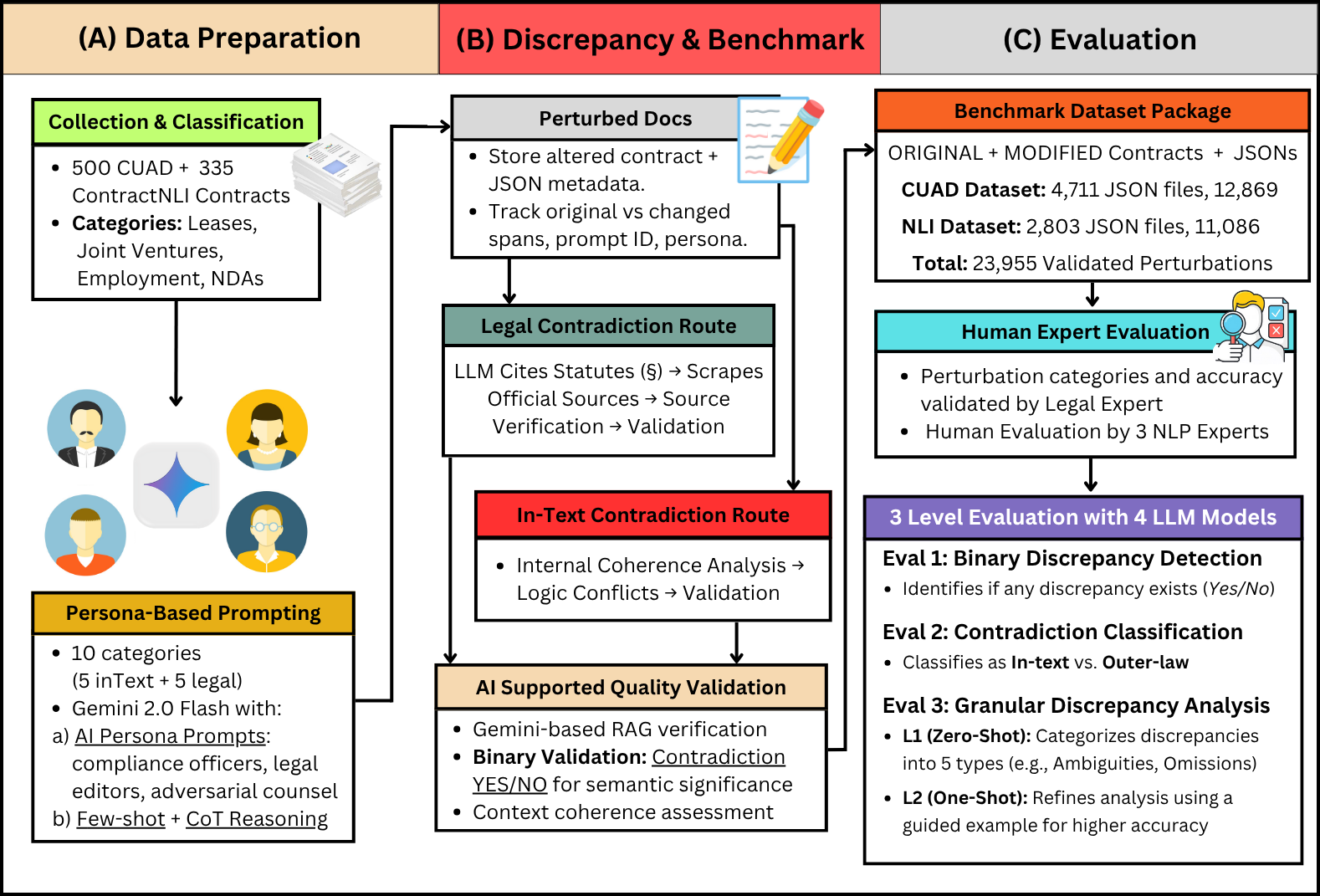}
\caption{\small \datasetName pipeline: Data Generation, AI grounding, and expert validation of legal and Evaluation \& Analysis.}
\label{fig:pipeline}
\end{figure*}

To ensure the benchmark’s fidelity and usefulness, we implement a multi-stage generation and validation pipeline. Each perturbation is first subjected to a retrieval-augmented generation (RAG) stage that grounds the modification in actual statutory language and related legal context, producing initial assessments of semantic significance and coherence. Automated quality checks then filter and classify contradictions, followed by expert human review to verify legal relevance, contradiction strength, and contextual correctness. All artifacts are accompanied by a rich metadata schema capturing provenance, reasoning chains, citation grounding, contradiction type, and confidence measures. This layered process yields high-quality labeled examples that surface both obvious and brittle failure modes in model reasoning.

Beyond creating examples, \datasetName is intended as a diagnostic and comparative platform for the research community. It enables quantification of model brittleness, supports evaluation of prompting and reasoning strategies, and supplies a standard substrate for developing and assessing mitigation techniques. By exposing where and how LLMs fail on realistic legal perturbations, the benchmark helps bridge the gap between fluent generation and reliable legal interpretation, accelerating progress toward safer, more transparent deployment in real-world contractual workflows. Our main contributions are as follows:\\

\vspace{-0.6em}
\noindent\textbf{1)} First benchmark for fine-grained legal contradictions with ten perturbation types grounded in statutory requirements.
    
\noindent\textbf{2)} Automated AI pipeline generating and validating contradictions before expert review.
    
\noindent\textbf{3)} Metadata framework capturing context, reasoning, and strength for reproducibility.
    
\noindent\textbf{4)} Extensive evaluation uncovering systematic legal reasoning weaknesses under varied prompting.

Figure~\ref{fig:pipeline} shows the CLAUSE pipeline: generation, grounding, and expert validation of legal and in-text contradictions.

\section{Our \datasetName Dataset}
The \datasetName benchmark advances legal reasoning evaluation through systematic analysis of contractual discrepancies. Built upon two established legal corpora, the Contract Understanding Atticus Dataset (CUAD) and ContractNLI, our framework represents the first fully AI-generated legal contradiction dataset. Using large language models for both generation and validation, we systematically enrich existing contracts through controlled perturbation. The core innovation lies in our comprehensive taxonomy of modifications, featuring distinct perturbation categories that span two fundamental dimensions:

\begin{itemize}[leftmargin=*,itemsep=0.1em]
\item \textbf{Legal Contradictions}: Modifications that conflict with statutory requirements, regulatory standards, or legal precedents, creating compliance issues that could render clauses unenforceable.
\item \textbf{InText Contradictions}: Modifications that create logical inconsistencies within the document itself, where different sections provide conflicting information or obligations.
\end{itemize}
\vspace{-0.4em}
Each dimension implements five fundamental perturbation types: (a) \textbf{Ambiguity}: Introduction of vague, unclear, or contradictory language that creates uncertainty in interpretation, making contract terms susceptible to multiple conflicting meanings; (b) \textbf{Omission}: Deliberate removal or absence of critical information, clauses, or terms that are necessary for complete understanding or legal enforceability of contract obligations; (c) \textbf{Misaligned Terminology}: Inconsistent use of defined terms throughout the document, or introduction of terminology that conflicts with established legal definitions or industry standards; (d) \textbf{Structural Flaws}: Modifications that disrupt the logical organization, hierarchy, or cross-referencing within the contract, creating navigational confusion or broken dependencies between clauses; and (e) \textbf{Inconsistencies}: Direct contradictions between different sections of the contract where statements, obligations, or terms are mutually exclusive or logically incompatible.

This structured approach, driven by AI generation and verification, enables fine-grained evaluation of LLM capabilities in both external legal compliance verification and internal document coherence assessment.

\paragraph{\datasetName Generation}
Our pipeline applies a unified perturbation framework to the contracts. We extract full text and structural metadata (clause boundaries, section headers, cross-references) to preserve document integrity during modification. Contextually appropriate contradiction locations are identified using prompts across ten perturbation types. Three complementary prompting strategies ensure legal plausibility: persona-based roles (compliance officers, legal editors), few-shot examples, and chain-of-thought reasoning. Each perturbation produces a modified contract with structured JSON metadata containing original/altered text, perturbation type, document location, and model-generated explanations. Legal contradictions include law citations and statutory context; in-text contradictions include conflicting clause references and logical justifications. The unified process is applied to both datasets ensuring consistent output quality and metadata format, supporting downstream detection accuracy and legal reasoning analysis tasks.

\paragraph{\datasetName LLM Verification}
To ensure dataset quality, we implement an AI-driven verification pipeline using specialized Gemini prompts with retrieval-augmented context. For each perturbation, the system analyzes original text, modified clause, and explanation, augmented with scraped legal references or related document clauses.

The verification performs rigorous binary classification (YES/NO) determining whether modifications represent strong, meaningful contradictions meeting our quality threshold. It evaluates semantic significance, contradiction strength, and contextual coherence. Only perturbations receiving YES classification are retained through automatic filtering, producing a clean, high-quality benchmark where each perturbation is explicitly validated for contradiction strength and interpretability.

\begin{table}[h]
\small
\centering
\setlength{\aboverulesep}{0pt}
\setlength{\belowrulesep}{0pt}
\setlength{\tabcolsep}{6pt}
\begin{tabular}{p{2.0cm}cccc}
\toprule
\multicolumn{5}{c}{\textbf{Dataset Distribution (CUAD + NLI)}} \\
\midrule
& \multicolumn{2}{c}{\underline{\textbf{InText}}} & \multicolumn{2}{c}{\underline{\textbf{Legal}}} \\
\textbf{Category} & \textbf{Files} & \textbf{Pert.} & \textbf{Files} & \textbf{Pert.} \\
\midrule
Ambiguity          & 741 & 2,417 & 805 & 2,472 \\
Inconsistency    & 749 & 2,357 & 830 & 2,711 \\
Misalign. Ter.   & 772 & 2,636 & 834 & 2,864 \\
Omission           & 754 & 2,498 & 698 & 2,111 \\
Structural   & 760 & 2,540 & 571 & 1,349 \\
\midrule
\textbf{Totals}   & \textbf{3,776} & \textbf{12,448} & \textbf{3,738} & \textbf{11,507} \\
\midrule
\multicolumn{3}{l}{\textbf{Dataset Totals}} & \textbf{CUAD} & \textbf{NLI} \\
\multicolumn{3}{l}{Total Contracts} & 4,711 & 2,803 \\
\multicolumn{3}{l}{Total Perturbations} & 12,869 & 11,086 \\
\bottomrule
\end{tabular}
\caption{\small Comprehensive dataset statistics showing five perturbation categories across InText and Legal variants, with dataset-wise breakdowns.}
\label{tab:combined_dataset_summary}
\end{table}

\paragraph{\datasetName Statistics}
\vspace{-0.5em}
Table~\ref{tab:combined_dataset_summary} presents comprehensive statistics for both components, showing the distribution of JSON files and perturbations across all ten categories. The dataset exhibits several notable characteristics. First, each JSON file contains multiple perturbations, with CUAD averaging 2.73 perturbations per file and the NLI split averaging 3.96 perturbations per file. Second, the distribution across categories is relatively balanced, though some variation exists due to the inherent complexity of generating specific perturbation types for different contract structures. Third, the legal categories generally contain fewer perturbations than their inText counterparts, reflecting the additional constraints imposed by statutory validation requirements. Extended data statistics for text length before and after perturbation is given in Table~\ref{tab:text_length_stats}.
\vspace{-0.5em}

\paragraph{\datasetName Human Verification}

\begin{table}[h]
\setlength{\aboverulesep}{0pt}
\setlength{\belowrulesep}{0pt}
\setlength{\tabcolsep}{2.5pt}
\small
\centering
\begin{tabular}{ccc}
\toprule
\textbf{Expert Pair} & \textbf{Cohen's Kappa} & \textbf{Jaccard Coefficient} \\
\midrule
$\alpha_{1,2}$ & 0.97 & 0.97 \\
$\alpha_{2,3}$ & 0.99 & 0.99 \\
$\alpha_{1,3}$ & 0.97 & 0.97 \\
$\alpha_{1,2,3}$ & 0.98 & 0.98 \\
\bottomrule
\end{tabular}
\caption{\small Inter-rater agreement metrics. Cohen's Kappa measures agreement between raters accounting for chance agreement, while Jaccard Coefficient measures similarity between finite sample sets of contradiction identifications.}
\label{tab:agreement}
\vspace{-0.5em}
\end{table}

To validate quality and legal accuracy, we first engaged a legal expert to confirm the real-world relevance of our perturbation categories. Subsequently, 3 NLP experts evaluated 25\% of our dataset (13,275 perturbation instances) using structured binary classification to determine meaningful legal contradictions. Our human verification demonstrates exceptional quality with 98.58\% contradiction validation rate and strong inter-rater reliability shown in Table~\ref{tab:agreement}. The evaluation examined semantic significance, contradiction strength, and contextual coherence, ensuring our benchmark contains high-quality examples for testing LLM legal reasoning capabilities. Legal expert verification statsin appendix Table~\ref{tab:expert_scores_contradiction}.

\section{Discrepancy Detection Tasks}\label{sec:tasks}

To rigorously assess the ability of large language models (LLMs) to detect and reason about legal discrepancies in contracts, we define a structured suite consisting of four hierarchical tasks over a dataset $\mathcal{D} = \{d_1, d_2, \dots, d_N\}$ of perturbed legal documents. Each $d_i$ may or may not contain a deliberately introduced discrepancy. We denote:
\begin{itemize}[leftmargin=*,itemsep=0.1em]
\item $y_i \in \{0, 1\}$. Binary label indicating whether any discrepancy is present in document $d_i$.
    
    \item $t_i \in \mathcal{T}$, where
    {
    \small
    \[
    \mathcal{T} = \{\mathrm{InText},\ \mathrm{Legal}\}.
    \]
    Type of contradiction: within the document or against external law.
    }
    \item $s_i \in \mathcal{S}$, where
    {
    \small
    \[
    \mathcal{S}
    = \left\{
    \begin{array}{@{}l@{}}
    \mathrm{ambiguity},\ \mathrm{omission},\\
    \mathrm{misaligned\_terminology},\\
    \mathrm{inconsistency},\ \mathrm{structural\_flaw}
    \end{array}
    \right\}.
    \]
    }

    Fine-grained discrepancy subtype.
    
    \item $\ell_i$. External legal reference violated by the discrepancy (only present if $t_i = \mathrm{legal}$).
    
    \item $x_i$. Text span in $d_i$ where the discrepancy occurs.
\end{itemize}

We now define four levels of evaluation:

\paragraph{Task\_1:}
 Binary Discrepancy Detection. Given $d_i$, predict $y_i \in \{0, 1\}$ indicating whether any discrepancy is present.

\paragraph{Task\_2:}
 Contradiction Type Classification. Given $d_i$, predict $c_i \in \mathcal{T} \cup \{\text{none}\}$ indicating whether the contradiction is in-text, legal (external), or absent.

\paragraph{Task\_3:}

Explanation-Based Discrepancy Detection. Given a document $d_i$, the model is required to identify a discrepancy span $x_i$, generate a natural language explanation $e_i$ justifying why $x_i$ constitutes a discrepancy, and the violated legal reference $l_i$ (if $t_i = \text{Legal}$ where $t_i \in \mathcal{T}$ is the contradiction type), or null otherwise. The predicted tuple $(x_i, e_i, l_i)$ is compared against ground truth annotations $(x_i^{*}, e_i^{*}, l_i^{*})$ for evaluation. 
\vspace{-0.2em}
\section{Modeling Approaches}
\vspace{-0.2em}
We design a tiered set of modeling strategies that correspond directly to the three evaluation tasks outlined in Section~\ref{sec:tasks}. Each task is implemented using tailored prompting strategies, with increasing granularity and reasoning complexity from Eval\_1 through Eval\_3. Below, we describe the modeling design for each evaluation level.
\vspace{-0.2em}
\paragraph{Discrepancy Detection}

 \textit{Eval\_1} is formulated as a binary classification task where the model must determine if a document contains any form of discrepancy, without specifying its type or location. To test this in a purely zero-shot setup, we use a minimal prompt that \textit{'asks whether there are any discrepancies present in the document'} and instructs the model to answer only with "Yes" or "No" . For this evaluation, models are provided with both the original and perturbed documents to facilitate the detection task.

\paragraph{Discrepancy Type Classification.}

\textit{Eval\_2} introduces a three-way classification task, where the model is asked to categorize the document as one of: (1) in-text contradiction, (2) outer-law contradiction, or (3) no contradiction. The prompt provides clear definitions of each category and explicitly instructs the model to choose one label. This task requires the model to reason over both internal document coherence and compatibility with U.S. legal standards. Compared to Eval\_1, this prompt includes richer task framing and imposes stricter answer constraints. For this evaluation we provide the models only the perturbed documents.

\paragraph{Discrepancy Span Detection with Explanation Generation:}

\textit{Eval\_3} requires the model to identify all discrepancy spans in a legal document and provide a natural language explanation for each. The model must output a structured list of objects, each containing two fields: the \texttt{text} of the problematic span and an \texttt{explanation} detailing why it is a discrepancy. For documents containing legal discrepancies, the model is also asked to output the contradicting \texttt{law citation}.
We divide \textit{Eval\_3} into zero-shot (\textbf{L1}) and one-shot (\textbf{L2}) evaluation levels. At both levels, models are tasked with identifying spans and providing explanatory justifications. Distinct prompts are designed for in-text and legal contradiction documents, each including definitions of the relevant contradiction type and the five primary discrepancy categories. The prompts for legal contradictions specifically request the law citation in addition to the span and explanation.


\section{Experimental Evaluation}

Our experiments are driven by a fundamental question that underpins the excitement around "AI in Law": Are today's models truly robust enough for reliable application in the high-stakes legal domain? To explore this, we investigate: (RQ1) the baseline performance of current LLMs on structured legal discrepancy detection; (RQ2) how prompt design affects a model's ability to identify and explain inconsistencies; (RQ3) the extent to which models generalize across different contradiction types ; and (RQ4) how reliably models can produce accurate, structured outputs such as legal citations and explanations.

\paragraph{Evaluation:} Our evaluation spans three tasks: discrepancy and it's type detection, and discrepancy span detection with explanation
generation. For evaluation we have used a temperature 0.2 across all the prompts. For Eval\_1 and Eval\_2, we use exact label matching. For Eval\_3, we define three evaluation metrics: \texttt{location\_alignment}, \texttt{explanation\_match}, and \texttt{law\_match}. Each of these evaluation metrics are explained below:

\texttt{\textbf{location\_alignment}}. First, let $D_{\text{GT}}$ be the set of all ground-truth discrepancy text spans and $D_{P}$ be the set of all predicted discrepancy text spans for a given document. For any discrepancy text span $d \in D_{\text{GT}} \cup D_{P}$, we define a function $f_{s}(d)$ that tokenizes the text into a set of normalized (lowercase, stripped) sentences.

We then construct an undirected graph $G = (V, E)$ where the vertices are the individual discrepancies, $V = D_{\text{GT}} \cup D_{P}$. An edge $(d_i, d_j)$ exists between a ground-truth span $d_i \in D_{\text{GT}}$ and a predicted span $d_j \in D_{P}$ if and only if their sentence sets have a non-empty intersection:
\begin{equation}
E = \left\{ (d_i, d_j) \mid
    \begin{aligned}
        & d_i \in D_{\text{GT}}, d_j \in D_{P}, \text{ and} \\
        & f_{s}(d_i) \cap f_{s}(d_j) \neq \emptyset
    \end{aligned}
\right\}
\end{equation}

The alignment is determined by finding the connected components of this graph. A "match group," which constitutes a \textbf{True Positive} alignment, is any connected component that contains at least one ground-truth span and one predicted span.

Here, a \textbf{False Negative} is a ground-truth span $d_{\text{GT}} \in D_{\text{GT}}$ that does not belong to any such match group and \textbf{False Positive} is a predicted span $d_{P} \in D_{P}$ that does not belong to any such match group.

\texttt{\textbf{explanation\_match}}. The LLM, acting as a legal analyst, receives the combined explanations and all relevant GT contexts. It returns a structured JSON object with scores between 0-5 for:
\textbf{Accuracy} (factual correctness),
\textbf{Completeness} (coverage of key points),
\textbf{Clarity} (ease of understanding),
\textbf{Legal Reasoning} (soundness of legal logic), and
a final binary \textbf{Match} flag that determines the alignment outcome. This process is adapted for two contradiction types. Ground-truth context are provided during the judging process for better context aware outputs: \textbf{a) In-Text Contradictions}: The GT context includes the \textit{original text}, \textit{changed text}, and \textit{contradicted text}; \textbf{b) Legal Contradictions}: The context is augmented to include the \textit{original text}, \textit{changed text}, \textit{contradicted text}, plus the specific \textit{contradicted law} and a \textit{law explanation}. Comparison between LLM-as-judge and human judge is in appendix Table~\ref{tab:llm_vs_human}.

\texttt{\textbf{law\_match}}. This metric evaluates the alignment of legal citations by using a Gemini model prompted as an expert legal paralegal. To determine if a predicted citation corresponds to the same underlying law as the ground truth, the model performs a semantic comparison using the ground-truth law explanations for context. This approach is designed to establish a fundamental match rather than a literal one, accounting for variations in textual phrasing and resulting in a binary score (0 or 1). All evaluations assume the current date is Sept. 2025.

For both \texttt{explanation\_match} \& \texttt{law\_match} we have used GPT 4o and Gemini-2.5-Flash as judges.   

\paragraph{Models:} We have evaluated four language models: GPT-4o-mini, Gemini 2.0 Flash \cite{geminiteam2025geminifamilyhighlycapable}, Gemini 2.5 Flash\cite{comanici2025gemini25pushingfrontier}, and LLaMa 3.3 70b Instruct\cite{grattafiori2024LLaMa3herdmodels}. For consistency and reproducibility, all models were evaluated with standardized parameters. Hyperparameters mentioned in Appendix Section A.

\begin{table*}[h!]
\setlength{\aboverulesep}{0pt}
\setlength{\belowrulesep}{0pt}
\centering
\small
\setlength{\tabcolsep}{2.5pt}
\begin{tabular*}{\textwidth}{@{\extracolsep{\fill}}l|cccc|cccc|cccc|cccc@{}}
\hline
\multirow{2}{*}{\textbf{Category}} & 
\multicolumn{4}{c|}{\textbf{GPT-4o-mini}} &
\multicolumn{4}{c|}{\textbf{Gemini-2.0}} &
\multicolumn{4}{c|}{\textbf{Gemini-2.5}} &
\multicolumn{4}{c@{}}{\textbf{LLaMa-3.3}} \\
\cline{2-17}
& \textbf{Acc} & \textbf{Pre} & \textbf{Rec} & \textbf{F1} &
  \textbf{Acc} & \textbf{Pre} & \textbf{Rec} & \textbf{F1} &
  \textbf{Acc} & \textbf{Pre} & \textbf{Rec} & \textbf{F1} &
  \textbf{Acc} & \textbf{Pre} & \textbf{Rec} & \textbf{F1} \\
\cmidrule{1-17}
\rowcolor{lightestgray}
\cellcolor{white} & \multicolumn{16}{c}{\textbf{CUAD}} \\
Amb$ _{\text{Leg}}$      & 32.9 & 36.4 & 47.6 & 41.2 & \textbf{51.2} & \textbf{50.4} & 76.5 & 60.8 & 48.1 & 48.7 & \textbf{91.8} & \textbf{63.7} & 40.2 & 43.5 & 69.4 & 53.5 \\
Amb$ _{\text{inT}}$     & 34.5 & 38.0 & 51.3 & 43.6 & \textbf{53.0} & \textbf{51.6} & 80.2 & 62.8 & 48.1 & 48.8 & \textbf{91.9} & \textbf{63.8} & 41.2 & 44.2 & 71.5 & 54.7 \\
Incon$ _{\text{Leg}}$      & 33.4 & 36.6 & 50.7 & 42.5 & \textbf{51.1} & \textbf{49.7} & 77.1 & 60.5 & 48.1 & 48.7 & \textbf{92.4} & \textbf{63.8} & 40.6 & 43.9 & 69.9 & 53.9 \\
Incon$ _{\text{inT}}$     & 34.5 & 38.4 & 51.7 & 44.0 & \textbf{53.4} & \textbf{52.1} & 80.5 & \textbf{63.3} & 47.1 & 48.2 & \textbf{90.1} & 62.8 & 39.9 & 43.3 & 68.8 & 53.1 \\
MisTerm$ _{\text{Leg}}$    & 36.0 & 39.3 & 53.8 & 45.4 & \textbf{53.7} & \textbf{52.0} & 81.4 & \textbf{63.5} & 47.4 & 48.6 & \textbf{90.3} & 63.2 & 40.6 & 43.8 & 70.0 & 53.9 \\
MisTerm$ _{\text{inT}}$   & 32.2 & 36.2 & 46.9 & 40.8 & \textbf{50.4} & \textbf{50.2} & 74.4 & 60.0 & 47.7 & 48.6 & \textbf{91.4} & \textbf{63.4} & 40.2 & 43.7 & 68.9 & 53.5 \\
Omis$ _{\text{Leg}}$       & 33.5 & 36.5 & 49.0 & 41.8 & \textbf{50.3} & \textbf{49.4} & 75.3 & 59.7 & 48.1 & 48.8 & \textbf{92.1} & \textbf{63.8} & 39.1 & 42.8 & 67.0 & 52.2 \\
Omis$ _{\text{inT}}$      & 33.7 & 36.9 & 48.7 & 42.0 & 49.0 & 48.8 & 72.1 & 58.2 & \textbf{49.8} & \textbf{49.6} & \textbf{95.8} & \textbf{65.4} & 39.4 & 43.0 & 67.5 & 52.5 \\
StrFlaw$ _{\text{Leg}}$    & 32.6 & 27.3 & 62.3 & 37.9 & 45.8 & 36.3 & 84.9 & 50.9 & \textbf{48.5} & \textbf{49.1} & \textbf{92.5} & \textbf{64.2} & 9.2 & 7.0 & 6.8 & 6.9 \\
StrFlaw$ _{\text{inT}}$   & 35.9 & 39.3 & 53.3 & 45.2 & \textbf{54.7} & \textbf{52.8} & 83.2 & \textbf{64.6} & 48.1 & 48.8 & \textbf{92.0} & 63.8 & 39.6 & 43.3 & 67.8 & 52.8 \\
\cmidrule{1-17}
\rowcolor{lightestgray}
\cellcolor{white} & \multicolumn{16}{c}{\textbf{NLI}} \\
Amb$ _{\text{Leg}}$      & 29.4 & 26.9 & 69.3 & 38.7 & \textbf{49.2} & \textbf{35.2} & 69.0 & 46.6 & 36.8 & 32.6 & \textbf{90.2} & \textbf{47.9} & 26.4 & 11.3 & 18.8 & 14.1 \\
Amb$ _{\text{inT}}$    & 33.6 & 29.0 & 91.7 & 44.1 & \textbf{54.0} & \textbf{37.3} & 89.7 & \textbf{52.7} & 36.0 & 30.5 & \textbf{97.1} & 46.4 & 28.3 & 12.0 & 24.0 & 16.0 \\
Incon$ _{\text{Leg}}$      & 32.7 & 30.0 & 74.4 & 42.8 & \textbf{52.9} & \textbf{40.0} & 78.6 & \textbf{53.0} & 40.0 & 35.6 & \textbf{95.8} & 51.9 & 28.2 & 15.2 & 24.6 & 18.8 \\
Incon$ _{\text{inT}}$     & 35.0 & 30.7 & 90.6 & 45.9 & \textbf{54.8} & \textbf{39.3} & 89.1 & \textbf{54.5} & 38.0 & 32.8 & \textbf{98.5} & 49.1 & 26.9 & 10.9 & 19.6 & 14.0 \\
MisTerm$ _{\text{Leg}}$    & 32.6 & 30.4 & 76.6 & 43.6 & \textbf{53.4} & \textbf{40.6} & 79.8 & \textbf{53.8} & 40.0 & 35.6 & \textbf{95.2} & 51.9 & 26.7 & 12.9 & 20.2 & 15.8 \\
MisTerm$ _{\text{inT}}$  & 34.1 & 30.1 & 88.6 & 45.0 & \textbf{51.7} & \textbf{36.4} & 79.2 & \textbf{49.9} & 37.4 & 32.3 & \textbf{97.4} & 48.5 & 27.8 & 12.4 & 22.7 & 16.0 \\
Omis$ _{\text{Leg}}$       & 24.9 & 20.7 & 66.4 & 31.6 & \textbf{45.7} & \textbf{26.9} & 62.6 & 37.6 & 32.2 & 26.7 & \textbf{91.6} & \textbf{41.3} & 26.0 & 6.8 & 14.5 & 9.3 \\
Omis$ _{\text{inT}}$      & 33.3 & 29.6 & 88.8 & 44.4 & \textbf{51.0} & \textbf{35.4} & 77.2 & \textbf{48.5} & 36.8 & 31.7 & \textbf{96.5} & 47.8 & 26.2 & 9.6 & 17.4 & 12.4 \\
StrFlaw$ _{\text{Leg}}$    & 34.8 & 32.0 & 85.2 & 46.5 & \textbf{54.1} & \textbf{40.8} & 82.8 & \textbf{54.6} & 38.6 & 34.5 & \textbf{93.4} & 50.4 & 28.8 & 15.9 & 26.4 & 19.8 \\
StrFlaw$ _{\text{inT}}$   & 35.7 & 31.0 & \textbf{94.9} & 46.7 & \textbf{55.8} & \textbf{39.7} & 93.8 & \textbf{55.8} & 36.0 & 31.0 & 94.5 & 46.7 & 27.4 & 11.3 & 21.1 & 14.7 \\
\hline
\end{tabular*}
\caption{\small Evaluation 1 (binary discrepancy detection) results across all models and categories for CUAD and NLI datasets. Metrics: Accuracy (Acc), Precision (Pre), Recall (Rec), and F1-Score (F1). Category abbreviations: Amb $=$ Ambiguity, Incon $=$ Inconsistencies, MisTerm $=$ Misaligned Terminology, Omis $=$ Omission, StrFlaw $=$ Structural Flaws, $ _{\text{Leg}}= $Legal, $ _{\text{inT}}= $inText. All values are percentages.}
\label{tab:eval1_full_results}
\end{table*}

\vspace{-0.5em}
\paragraph{Metrics:} For the Eval\_1 and Eval\_2, we report standard metrics: Accuracy, Precision, Recall, and F1-Score. For Eval\_3, \texttt{location\_alignment} is evaluated using ROUGE, METEOR, and BERTScore, while \texttt{explanation\_match} is assessed via qualitative 0-5 scores for Accuracy, Clarity, Completeness, and Legal Reasoning.

\subsection{Findings: Results \& Analysis}
\datasetName poses a significant \textbf{challenge}: Discrepancy detection remains challenging across datasets and discrepancy categories, with low precision, recall, F1 score, and semantic and syntax matching scores. \\

\vspace{-0.75em}
\noindent\textbf{How does model performance on binary discrepancy detection vary across different model architectures and datasets?}

\noindent As shown in Table~\ref{tab:eval1_full_results}, a clear performance hierarchy emerges: Gemini-2.5 leads with the highest F1-scores (63\%+ on CUAD), followed by Gemini-2.0, while GPT-4o-mini and LLaMa-3.3 lag. Gemini-2.5's top performance is driven by a distinct high-recall (90\%+), low-precision strategy, whereas other models exhibit a more balanced profile. Consistently, all models perform better on the CUAD dataset than on the more complex NLI dataset, as exemplified by Gemini-2.5’s F1-score for Inconsistencies\_Legal dropping from 63.8\% (CUAD) to 51.9\% (NLI). \\

\vspace{-0.75em}
\noindent\textbf{Which contradiction types prove most challenging, and is there a consistent performance gap between \textit{intext} and \textit{legal} discrepancies?}

\noindent While no single category is universally hardest in Table~\ref{tab:eval1_full_results}, performance on \textit{legal} discrepancies is often marginally lower than on their \textit{in\_text} counterparts. A clear trend of difficulty is seen in the \textit{Omission\_Legal} category on the NLI dataset, which proves exceptionally challenging for GPT-4o-mini ($31\%$ F1) and LLaMa-3.3 ($9.3\%$ F1). A similar anomaly is LLaMa-3.3's $6.9\%$ F1-score on CUAD's \textit{Structural\_Flaws\_Legal}. This suggests that identifying legally significant absences and complex structural errors are key weaknesses for some architectures.
\vspace{-1em}
\begin{table}[h!]
\centering
\small
\renewcommand{\arraystretch}{1.15}
\setlength{\tabcolsep}{3pt}
\begin{tabular*}{\columnwidth}{@{\extracolsep{\fill}}l|c|c|c|c|c|c|c|c@{}}
\hline
\multirow{2}{*}{\textbf{Cat.}} & \multicolumn{2}{c|}{\textbf{GPT}} & \multicolumn{2}{c|}{\textbf{Gem2.0}} & \multicolumn{2}{c|}{\textbf{Gem2.5}} & \multicolumn{2}{c@{}}{\textbf{LLaMa}} \\
\cline{2-9}
& \textbf{C} & \textbf{N} & \textbf{C} & \textbf{N} & \textbf{C} & \textbf{N} & \textbf{C} & \textbf{N} \\
\hline
Amb. & \textbf{53.3} & 68.4 & 51.5 & \textbf{71.6} & 52.2 & 48.4 & 50.2 & 55.2 \\
Inc. & 54.0 & 66.2 & \textbf{55.1} & \textbf{73.3} & 51.4 & 51.0 & 50.9 & 56.4 \\
Str. & \textbf{63.9} & 56.5 & 52.3 & \textbf{58.9} & 47.3 & 47.9 & 50.5 & 52.2 \\
Mis. & \textbf{52.8} & 58.7 & 53.5 & \textbf{66.0} & 49.8 & 49.6 & 51.0 & 55.3 \\
Omi. & 51.4 & \textbf{67.4} & \textbf{52.3} & 63.0 & 50.0 & 54.8 & 50.5 & 51.1 \\
\hline
\end{tabular*}
\caption{\small Eval. 2 (contradiction type classification) accuracy. Cat.: Amb.=Ambiguity, Inc.=Inconsistencies, Str.=Structural Flaws, Mis.=Misaligned Terminology, Omi.=Omission. C=CUAD, N=NLI datasets. Values are percentages.}
\label{tab:eval2_results}
\end{table}

\noindent\textbf{How does the accuracy of LLMs in classifying contradiction types vary across different model architectures, datasets, and specific categories of legal discrepancies?}

\noindent The results from Eval\_2 in Table~\ref{tab:eval2_results} show significant variance, with performance depending on both the model and dataset. GPT-4o-mini performs strongest on CUAD (63.9\% in Structural Flaws), whereas Gemini-2.0 leads on the more difficult NLI dataset (73.3\% in Inconsistencies). The increased difficulty of NLI is a clear trend across models, illustrated by Gemini-2.5's accuracy on Ambiguity dropping from 52.2\% (CUAD) to 48.4\% (NLI). This highlights that while models possess aptitudes for certain discrepancy structures, robust generalization in classification remains a significant challenge.

\begin{figure}[h!]
\centering
\includegraphics[width=\linewidth]{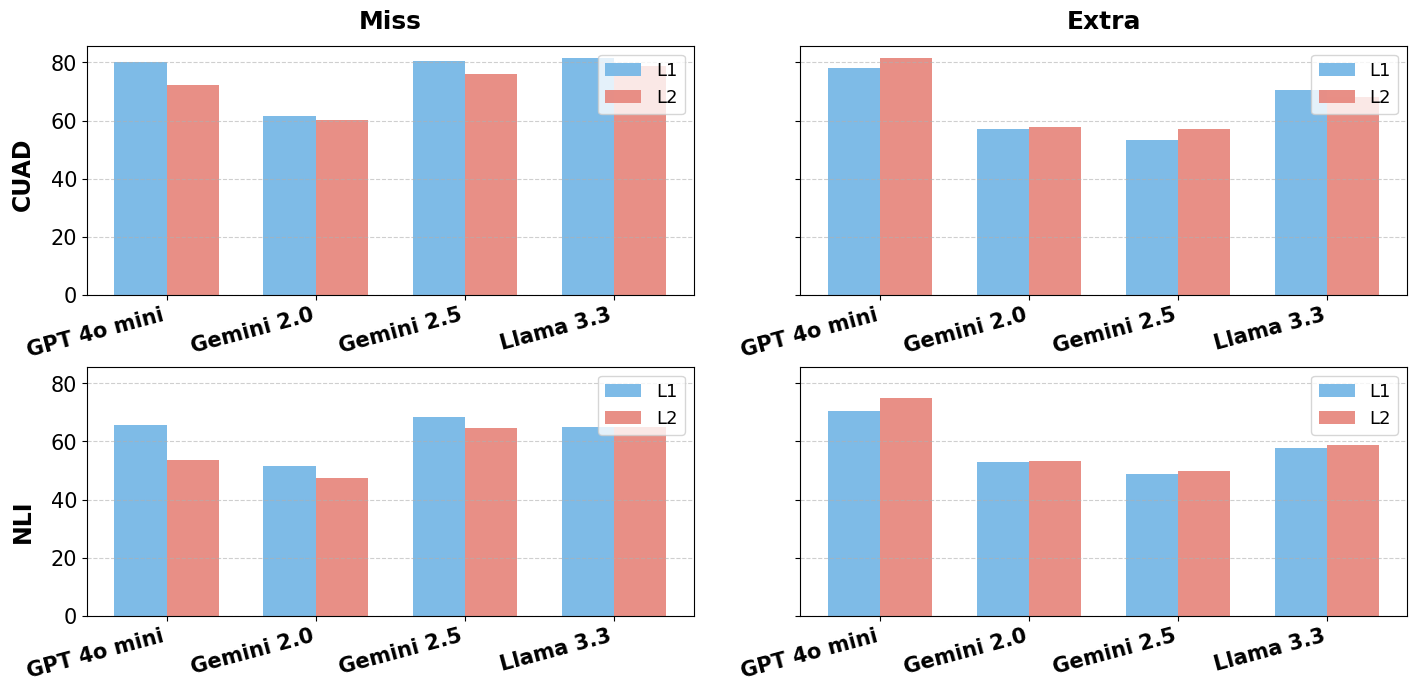}
\caption{ Comparison of model performance on CUAD and NLI datasets across L1 and L2 levels. The top two plots show Miss and Extra metrics for CUAD, while the bottom two correspond to NLI.}
\label{fig:my_image}
\end{figure}
\vspace{-0.5 em}
We analyze model performance using two key error metrics in Fig.~\ref{fig:my_image} : the Miss rate, which is the fraction of ground-truth perturbations models failed to get aligned with any predictions after the \texttt{location\_alignment}, the reverse logic goes for the Extra rate. 

\noindent\textbf{How does one-shot prompting (L2) versus zero-shot (L1) impact model error rates, and how do these patterns vary across models and datasets?}

\noindent A clear trade-off emerges in Fig.~\ref{fig:my_image}: L2 prompting lowers the Miss rate (better detection) but increases the Extra rate (more false positives) across all datasets. This sensitivity is most pronounced in GPT-4o-mini and Gemini-2.5, while Gemini-2.0 and LLaMa-3.3 show more stable performance. Crucially, the NLI dataset consistently results in higher error rates for all models, confirming its status as the most challenging benchmark for both detection and precision. For complete miss and extra rate across categories please refer to Table~\ref{tab:category_values} in the appendix.

\begin{table}[h!]
\setlength{\aboverulesep}{0pt}
\setlength{\belowrulesep}{0pt}
\centering
\small
\setlength{\tabcolsep}{0.6pt}
\begin{tabular*}{\columnwidth}{@{\extracolsep{\fill}}l|cccc@{\hspace{2pt}}|@{\hspace{2pt}}cccc@{}}

\hline
\multirow{2}{*}{\textbf{Data}} & \multicolumn{4}{c}{\textbf{L1}} & \multicolumn{4}{c@{}}{\textbf{L2}} \\
\cmidrule{2-9}
& \textbf{R$_1$} & \textbf{R$_2$} & \textbf{R$_L$} & \textbf{M} & \textbf{R$_1$} & \textbf{R$_2$} & \textbf{R$_L$} & \textbf{M} \\
\cmidrule{1-9}
\rowcolor{lightestgray}
\cellcolor{white} & \multicolumn{8}{c}{\textbf{GPT-4o-mini}} \\
CUAD & 61.4 & 60.7 & 60.6 & 52.0 & 62.4 & 61.6 & 61.6 & 53.1 \\
NLI & 67.6 & 66.6 & 66.7 & 59.8 & 68.9 & 67.9 & 67.9 & 61.4 \\
\cmidrule{1-9}
\rowcolor{lightestgray}
\cellcolor{white} & \multicolumn{8}{c}{\textbf{Gemini-2.0}} \\
CUAD & 72.8 & 71.9 & 71.5 & 66.5 & 72.6 & 71.7 & 71.4 & 66.2 \\
NLI & 74.9 & 73.5 & 73.6 & 71.3 & 75.4 & 74.0 & 74.0 & 71.0 \\
\cmidrule{1-9}
\rowcolor{lightestgray}
\cellcolor{white} & \multicolumn{8}{c}{\textbf{Gemini-2.5}} \\
CUAD & 67.1 & 65.3 & 65.0 & 60.5 & 64.2 & 62.1 & 62.5 & 57.4 \\
NLI & 71.6 & 69.5 & 69.5 & 68.3 & 72.1 & 69.9 & 69.9 & 68.9 \\
\cmidrule{1-9}
\rowcolor{lightestgray}
\cellcolor{white} & \multicolumn{8}{c}{\textbf{LLaMa-3.3}} \\
CUAD & 66.6 & 65.6 & 65.0 & 58.0 & 66.6 & 65.6 & 65.1 & 57.9 \\
NLI & 71.4 & 70.0 & 69.7 & 65.4 & 71.2 & 71.1 & 70.0 & 65.1 \\
\hline
\end{tabular*}
\vspace{-0.5em}
\small\caption{Eval. 3 location alignment metrics averaged across all categories. L1=Level 1 (zero-shot), L2=Level 2 (one-shot). R$_1$=ROUGE-1, R$_2$=ROUGE-2, R$_L$=ROUGE-L, M=METEOR. All values are percentages.}
\label{tab:eval3_location}
\vspace{-0.5em}
\end{table}

\noindent\textbf{How accurately do LLMs identify the textual location of legal discrepancies, and what does this location\_alignment performance reveal about their capabilities?}

\noindent In Table~\ref{tab:eval3_location}, the \texttt{location\_alignment} results show three key trends: 1) Gemini-2.0 is the clear leader (e.g., 75.4 R$_1$ on NLI), consistently outperforming even the more advanced Gemini-2.5. 2) All models universally score higher on the NLI dataset than on CUAD, suggesting NLI's discrepancy structures are more textually distinct. 3) One-shot (L2) prompting offers inconsistent and sometimes negative effects; most notably, it degrades Gemini-2.5's performance on CUAD (R$_1$ drops from 67.1 to 64.2). This counter-intuitive result indicates that single-example guidance does not reliably improve, and can even hinder, precise location detection. For complete \texttt{location\_alignment} results across categories refer to Table~\ref{tab:eval3_location_detailed} in the appendix.

\begin{table}[h!]
\centering
\setlength{\aboverulesep}{0pt}
\setlength{\belowrulesep}{0pt}
\setlength{\tabcolsep}{2.2pt}
\small
\begin{tabular}{l|cccc|cccc}
\hline
\multirow{3}{*}{\textbf{Model}} &
\multicolumn{4}{c|}{\textbf{CUAD}} &
\multicolumn{4}{c}{\textbf{NLI}} \\
\cline{2-9}
 & \multicolumn{2}{c|}{\textbf{L1}} &
  \multicolumn{2}{c|}{\textbf{L2}} &
  \multicolumn{2}{c|}{\textbf{L1}} &
  \multicolumn{2}{c}{\textbf{L2}} \\
\cline{2-9}
 & GPT & Gem & GPT & Gem &
  GPT & Gem & GPT & Gem \\
\hline
\rowcolor{lightestgray}
\cellcolor{white} & \multicolumn{8}{c}{\textbf{GPT-4o-mini}} \\
Accuracy & 2.6 & 2.1 & 2.9 & 2.5 & 2.9 & 2.5 & 2.8 & 2.5 \\
Clarity & 3.7 & 4.0 & 3.9 & 4.2 & 3.9 & 4.2 & 3.8 & 4.1 \\
Complete & 2.0 & 1.4 & 2.3 & 1.7 & 2.3 & 1.7 & 2.2 & 1.7 \\
Legal Rea. & 2.6 & 2.3 & 2.9 & 2.5 & 2.9 & 2.5 & 2.8 & 2.4 \\
\hline
\rowcolor{lightestgray}
\cellcolor{white} & \multicolumn{8}{c}{\textbf{Gemini-2.0}} \\
Accuracy & 3.0 & 2.6 & 2.9 & 2.5 & 2.9 & 2.5 & 3.0 & 2.7 \\
Clarity & 4.0 & 4.4 & 3.9 & 4.2 & 3.9 & 4.2 & 4.0 & 4.2 \\
Complete & 2.5 & 1.8 & 2.4 & 1.9 & 2.4 & 1.9 & 2.6 & 1.9 \\
Legal Rea. & 3.0 & 2.9 & 2.9 & 2.6 & 2.9 & 2.6 & 3.0 & 2.7 \\
\hline
\rowcolor{lightestgray}
\cellcolor{white} & \multicolumn{8}{c}{\textbf{Gemini-2.5}} \\
Accuracy & 3.0 & 2.4 & 2.9 & 2.4 & 2.9 & 2.4 & 2.9 & 2.4 \\
Clarity & 4.0 & 4.3 & 3.9 & 4.2 & 3.9 & 4.2 & 3.9 & 4.1 \\
Complete & 2.5 & 1.9 & 2.6 & 2.1 & 2.6 & 2.1 & 2.6 & 2.1 \\
Legal Rea. & 3.1 & 3.0 & 3.0 & 2.9 & 3.0 & 2.9 & 3.0 & 2.9 \\
\hline
\rowcolor{lightestgray}
\cellcolor{white} & \multicolumn{8}{c}{\textbf{LLaMa-3.3}} \\
Accuracy & 2.4 & 2.2 & 2.7 & 2.3 & 2.6 & 2.3 & 2.6 & 2.5 \\
Clarity & 3.5 & 4.1 & 3.8 & 4.0 & 3.7 & 4.1 & 3.7 & 4.1 \\
Complete & 1.8 & 1.4 & 2.1 & 1.6 & 2.1 & 1.5 & 2.3 & 1.6 \\
Legal Rea. & 2.4 & 2.2 & 2.6 & 2.2 & 2.6 & 2.2 & 2.5 & 2.3 \\
\hline
\end{tabular}
\caption{ Averaged explanation quality metrics across all categories. Judges: GPT4o (GPT), Gemini 2.5 (Gem). Here, the metrics Complete = Completeness, Legal Rea. = Legal Reasoning are abbreviated. All values on a scale of 0-5.}
\label{tab:eval3_explanation_collapsed}
\end{table}
\vspace{-0.7em}
\noindent\textbf{What is the overall quality of legal explanations generated by LLMs, and what factors influence this quality?}

\noindent As seen in Table~\ref{tab:eval3_explanation_collapsed}, the quality of legal explanations is defined by a critical trade-off between fluency and substance. Models universally generate text with high Clarity (often 4.0+) but low Completeness (<2.0), indicating fluent but shallow reasoning. Model architecture is a primary factor, establishing a clear performance hierarchy (Gemini > GPT-4o-mini > LLaMa-3.3) in substantive metrics like Accuracy and Legal Reasoning. Prompting strategy has a negligible impact, as L2 prompting offers minimal, inconsistent gains. For complete \texttt{explanation\_match} results across categories refer to Table~\ref{tab:eval3_explanation_quality} in the appendix. \\
\vspace{-1.75em}

\begin{table}[h!]
\centering
\small
\setlength{\aboverulesep}{0pt}
\setlength{\belowrulesep}{0pt}
\setlength{\tabcolsep}{2.8pt}
\begin{tabular}{l|cccc|cccc}
\hline
\multirowcell{3}{\textbf{Model}} & \multicolumn{4}{c|}{\textbf{CUAD}} & \multicolumn{4}{c}{\textbf{NLI}} \\ 
\cline{2-9}
 & \multicolumn{2}{c}{\textbf{GPT4o}} & \multicolumn{2}{c|}{\textbf{Gem2.5}} & \multicolumn{2}{c}{\textbf{GPT4o}} & \multicolumn{2}{c}{\textbf{Gem2.5}} \\
\cline{2-9}
 & \textbf{L1} & \textbf{L2} & \textbf{L1} & \textbf{L2} & \textbf{L1} & \textbf{L2} & \textbf{L1} & \textbf{L2} \\ 
\hline
\textbf{GPT4o mini} & 5.4 & 6.7 & 5.0 & 6.6 & 3.6 & 5.4 & 3.3 & 4.7 \\
\textbf{Gemini 2.0} & 7.9 & 7.1 & 7.6 & 6.6 & 3.2 & 3.5 & 3.4 & 3.6 \\
\textbf{Gemini 2.5} & 12.2 & 13.4 & 13.7 & 13.5 & 7.4 & 10.8 & 8.7 & 12.3 \\
\textbf{LLaMa 3.3} & 9.2 & 10.5 & 8.4 & 9.4 & 6.3 & 8.9 & 6.2 & 8.6 \\
\hline
\end{tabular}
\caption{Average \textit{Law\_Match} percentage averaged across categories where datasets (CUAD, NLI), levels (\textbf{L1}, \textbf{L2}), and judges (GPT4o vs Gemini 2.5 Flash).}
\label{tab:law_match_all}
\end{table}
\vspace{-0.75em}

\noindent\textbf{To what extent can current LLMs accurately perform semantic matching of legal citations, and how is this capability modulated by model architecture, dataset complexity, and prompting strategy?}

\noindent The ability of LLMs to semantically match legal citations is extremely limited, with even the best-performing model, Gemini 2.5, achieving scores below 14\% (Table~\ref{tab:law_match_all}. A clear performance hierarchy is evident, with Gemini 2.5 as the leader and Gemini 2.0 as the weakest, indicating highly specialized model capabilities. Performance consistently degrades from the CUAD to the more complex NLI dataset across all models. While L2 prompting provides a modest uplift, particularly for stronger models like Gemini 2.5 and LLaMa 3.3, and this is primarily because AI models struggle to extract the external knowledge needed to obtain law citations. 

Our analysis answers the four primary research questions through targeted empirical tests. We establish baseline performance and generalization (RQ 1, 3) by evaluating binary detection and contradiction classification accuracy across models and discrepancy types. The influence of prompt design (RQ2) is measured by comparing error rates between L1 and L2 settings. Finally, we assess the reliability of generating structured outputs (RQ4) through granular evaluations of location alignment, explanation quality, and law citation matching.

\section{Comparison with Related Works} 
\vspace{-0.5em}
Recent legal NLP advances have been propelled by datasets like CUAD \cite{hendrycks2021cuad} and LEDGAR \cite{tuggener2020ledgar}, which focus on clause labeling and categorization, as well as LexGLUE \cite{chalkidis2022lexglue}, ContractNLI \cite{koreeda2021contractnli}, and CaseHOLD \cite{zhong2020jec}, which target tasks such as entailment and case retrieval. However, these resources rely on manual annotation and do not assess model reasoning under targeted perturbations. While recent work explores hallucination in legal text \cite{shen2023hallucination} and LegalBench \cite{zheng2023legalbench} broadens task diversity, they do not simulate document-level inconsistencies or robustness to semantic edits. Retrieval-augmented generation (RAG) has shown promise in open-domain QA \cite{lin2022truthfulqa, ji2023haluEval}, but is rarely applied to legal contradiction detection. \datasetName is among the first fully automated, AI-driven legal benchmarks, using LLMs for both generation and validation. This scalable pipeline enables the creation of high-quality contradiction data, moving beyond static classification to simulate and verify realistic legal edits. Our approach uniquely bridges gaps in legal QA and hallucination detection, demonstrating the potential of automated dataset creation for complex legal reasoning tasks.
\vspace{-0.5em}
\section{Conclusion}
\vspace{-0.5em}
In this work, we present \datasetName, a benchmark designed to evaluate legal reasoning in large language models via systematic contradiction detection. By combining persona-driven generation with RAG-based validation, we show that automated methods can produce high-quality, legally-grounded datasets. Our evaluations highlight significant gaps in current LLMs’ abilities to detect and explain nuanced contractual inconsistencies, revealing key challenges for legal AI. \datasetName thus offers a foundation for advancing and validating legal AI systems, supporting the development of more reliable automated legal analysis tools. Through the four research questions explored in our experimental evaluation, covering performance, prompt sensitivity, generalization, and structured reliability, this study provides empirical evidence addressing the broader inquiry of whether current AI models are dependable enough for legal deployment. Our findings make this clear: despite impressive linguistic fluency, today’s LLMs lack the depth of legal reasoning required for high-stakes legal practice.

\section*{Acknowledgements}
We are grateful to Rui Heng Foo for his invaluable assistance with the data generation pipelines. We extend our sincere thanks to Arizona State University's Lincoln Center for Applied Ethics, along with their law professors and partners, for lending their expertise to advise on our perturbation categories and to authenticate our perturbed contracts for real-world relevance. We also thank the ASU Law and Pre-Law students for volunteering their time and expertise. Finally, we acknowledge the CoRAL Lab at Arizona State University for providing the computational resources that made this research possible.

\section*{Limitations}
As a foundational work in auditing the legal reasoning of LLMs, \textbf{CLAUSE} establishes a new paradigm for benchmark creation and evaluation. The deliberate choices made in its design define its current scope and naturally pave the way for exciting future extensions.

\begin{enumerate}[leftmargin=*,itemsep=0.1em]
    \item \textbf{Dataset Focus:} Our benchmark is built using U.S. commercial contracts from the well-regarded \textbf{CUAD} and \textbf{ContractNLI} datasets. This allows for a deep and relevant analysis of model performance in this specific, high-impact area. Consequently, our findings are most directly applicable to U.S. law, and future work could readily adapt our successful methodology to create similar benchmarks for international contracts or other legal document types.

    \item \textbf{AI-Driven Methodology:} A key strength of our work is the innovative AI pipeline that allows us to generate thousands of contract flaws automatically. This makes creating a large-scale benchmark feasible. The nature of these flaws is influenced by the capabilities of the generation model (Gemini 2.0 Flash), and our validation simplifies risk into a ``YES/NO'' decision. This scalable approach provides a powerful baseline, which can be complemented in the future by studies that focus on more complex, manually-crafted scenarios or the varying degrees of legal risk.

    \item \textbf{Evaluation Framework:} To assess the quality of the models' legal explanations at scale, we used other LLMs as judges. An interesting finding was that different AI judges showed slight variations in their scoring, which points to a promising research direction in developing robust, automated evaluation techniques. While our metrics confirm that models can identify the correct \textit{semantic area} of a flaw, future research can build on our work by developing metrics that also assess a model's ability to pinpoint the most legally critical part of that text.

    \item \textbf{Model Scope:} Our experiments provide a crucial baseline for the ``out-of-the-box'' reasoning capabilities of four major LLM families as of late 2025. We tested them in a direct, instruction-following format to measure their core abilities. This work sets the stage for future studies to explore the performance of specialized models that are fine-tuned on legal data or are integrated into practical, real-world systems with human oversight.

    \item \textbf{Scope of Human Verification:} To ensure the highest quality, a rigorous validation of our dataset was conducted by 3 NLP experts who reviewed a significant 25\% sample of all generated perturbations. The exceptional agreement rate (98.58\%) on this large sample provides high confidence in the overall quality of our automated pipeline. Future work could build on this by conducting even larger-scale human validation studies or by developing advanced AI verifiers that correlate more closely with human expert judgment.

    \item \textbf{Isolated Perturbation Analysis:} Our current methodology generates multiple, distinct perturbations for each source document, allowing us to precisely test a model's ability to detect each of the 10 types of flaws in isolation. Real-world contracts, however, can sometimes contain multiple, interacting issues (e.g., an ambiguous definition that in turn causes an inconsistency elsewhere). An exciting direction for future research is to create benchmarks with such compound discrepancies, which would test the deeper, holistic analysis capabilities of advanced LLMs.
\end{enumerate}

\section*{Ethics Statement}

\datasetName is intended to improve the safety and reliability of LLMs in legal settings by making their failure modes visible. We build on publicly available or appropriately licensed contract data, introduce no new personal identifiers, and retain only perturbations that pass automated checks and expert human validation to avoid propagating spurious signals. Any personally identifiable information (PII) present in source datasets (CUAD or ContractNLI) has been systematically redacted using the standardized placeholder ``(information\_redacted)''; we have verified that this redaction preserves textual meaning and does not degrade model performance on our evaluation tasks. The benchmark is not a source of legal advice; users are expected to combine model outputs with qualified human judgment. We document limitations including dataset skew and inherited legal norms, and encourage future work to broaden coverage and address bias. Release materials include usage guidance to discourage adversarial or deceptive repurposing. Overall, the goal is diagnostic: enable more accountable deployment of legal AI.

We acknowledge risks and implement appropriate safeguards. To prevent adversarial misuse where the benchmark could be exploited to train evasion models or generate flawed legal systems, we employ multi-layered mitigation: (1) transparent documentation of our perturbation methodology to enable reproducibility while deterring exploitation, (2) controlled dataset access via a mandatory request form requiring users to detail their research purpose, allowing us to screen for malicious intent, and (3) distribution with a comprehensive datasheet containing explicit ethical usage guidelines that prohibit adversarial training, commercial deployment without human oversight, and use of perturbed contracts as legal templates. Strong performance on \datasetName should not be interpreted as deployment certification; the benchmark evaluates specific capabilities under controlled conditions and cannot capture real-world legal complexity. We emphasize in all documentation that this is a diagnostic research tool for capability assessment, not a deployment readiness guarantee. The perturbed contracts intentionally contain legal flaws and must never be used as reference documents; we include prominent warnings and exclude all contracts with sensitive or personally identifiable information. Our benchmark reflects U.S.-centric commercial contract law and may disadvantage models trained on international or specialized legal corpora; we explicitly acknowledge these limitations and encourage community extensions to diverse legal systems. Finally, to address over-reliance on automated legal analysis, we document fundamental gaps in current models' legal reasoning that preclude unsupervised deployment, emphasizing that LLM outputs require qualified human legal review for high-stakes applications. 

The validation of \datasetName was conducted entirely on a voluntary basis by one legal expert and three NLP researchers who contributed their expertise without compensation. These experts were identified based on their familiarity with demographic nuances and their language proficiency needed to evaluate this dataset. The experts assessing \datasetName recognized the significant societal value of this work in advancing safer AI deployment in high-stakes legal contexts, including improving access to justice through more reliable automated legal analysis tools, preventing catastrophic failures in contract review systems that could lead to substantial financial and legal harm, and establishing rigorous evaluation standards that protect vulnerable populations from flawed AI-driven legal services. Their pro bono contributions reflect a shared commitment to responsible AI development and the belief that transparent, community-driven benchmarks are essential for building trustworthy legal AI systems that serve the public interest. The NLP researchers were provided with comprehensive guidelines (Appendix ~\ref{sec:human-evaluation-rubric}) to evaluate the dataset with consistency and accuracy.

We utilized Large Language Models in a controlled environment to edit and refine some sections of the paper.

\bibliography{acl_latex}

\appendix

\section{\datasetName Generation and Evaluation Details}
\label{sec:appendix-details}

This appendix provides comprehensive details on our dataset generation pipeline, complete evaluation results across all experimental conditions, and real-world case examples that motivated our perturbation taxonomy. We organize this material in three parts: first, we present the prompts and methodology underlying our data generation and validation processes; second, we provide detailed breakdowns of evaluation results with in-depth analysis of model performance patterns; and third, we document real-world contractual failures that demonstrate the practical significance of each perturbation type in our benchmark.

\paragraph{Experiment Parameters \& Setup}For data generation generation we have used Gemini 2.0 Flash at temperature 0.7 for balanced coherence and speed. For evaluation we used a temperature of 0.2 for more deterministic and focused outputs suitable for analysis, disabled all safety filters to prevent content moderation from interfering with the legal text analysis, and set a maximum token limit of 8,192 to ensure explanations were not prematurely truncated.

\subsection{Source Dataset Documentation}
\label{sec:source-dataset-documentation}

\datasetName is built upon two foundational legal NLP datasets: CUAD (Contract Understanding Atticus Dataset) and ContractNLI. Both datasets are licensed under Creative Commons Attribution 4.0 International (CC BY 4.0).

\paragraph{CUAD Dataset} The CUAD dataset comprises over 510 commercial legal contracts from various industries, annotated with 13,000+ expert labels across 41 categories relevant to contract review. The dataset covers English-language commercial contracts and focuses on complex contractual language, legal jargon, clause extraction, and nuanced legal concepts. Created with input from legal experts, CUAD reflects realistic legal issues encountered in contract review tasks. The dataset does not include demographic annotations, as it is document-based rather than human subject-focused, and provides monolingual coverage (English only).

\paragraph{ContractNLI Dataset} ContractNLI is the largest corpus of annotated contracts for natural language inference, containing approximately 607 contracts (primarily non-disclosure agreements). The dataset addresses document-level entailment, contradiction, and neutral inference tasks in English-language commercial contracts. It covers linguistic phenomena including negations, exceptions, multi-span evidence detection, and complex legal reasoning patterns. Like CUAD, ContractNLI does not include demographic annotations due to its focus on contract semantics and formal logic rather than human subject information, and provides monolingual English coverage.

\paragraph{Coverage Summary} Both source datasets provide comprehensive coverage of English legal contract language and document structure relevant to legal NLP benchmarking. They capture complex linguistic phenomena specific to contractual agreements—including legal reasoning markers, clause variations, and formal semantics—but do not address demographic representation or multilingual aspects due to the nature of commercial contract sources. These characteristics are inherited by \datasetName, which extends the source datasets through systematic perturbation while maintaining their domain focus on U.S. commercial contract law.
\subsection{Implementation Details and Package Documentation}
\label{sec:implementation-packages}

This section documents all software packages, models, and parameter settings used throughout our pipeline for data generation, validation, and evaluation.

\paragraph{Data Generation Models} For perturbation generation, we used Gemini 2.0 Flash (model ID: \texttt{gemini-2.0-flash-exp}) via Google's Gemini API with temperature 0.7, top-p 0.95, and maximum output tokens 8192. All safety filters were disabled to prevent content moderation interference with legal text generation. For LLM-based validation of generated perturbations, we used Gemini 2.5 Flash (model ID: \texttt{gemini-2.5-flash-002}) with temperature 0.2 for more deterministic outputs, top-p 0.95, and maximum output tokens 8192.

\paragraph{Evaluation Models} We evaluated four language models: (1) GPT-4o-mini (model ID: \texttt{gpt-4o-mini-2024-07-18}) accessed via OpenAI API with temperature 0.2, top-p 1.0, and max tokens 8192; (2) Gemini 2.0 Flash (model ID: \texttt{gemini-2.0-flash-exp}) with temperature 0.2, top-p 0.95, and max tokens 8192; (3) Gemini 2.5 Flash (model ID: \texttt{gemini-2.5-flash-002}) with temperature 0.2, top-p 0.95, and max tokens 8192; and (4) LLaMa 3.3 70B Instruct (model ID: \texttt{meta-LLaMa/LLaMa-3.3-70B-Instruct}) accessed via Groq API with temperature 0.2, top-p 1.0, and max tokens 8192. All models used identical generation parameters for fair comparison.

\paragraph{LLM Judge Models} For \texttt{explanation\_match} and \texttt{law\_match} evaluation metrics, we employed two judge models: GPT-4o (model ID: \texttt{gpt-4o-2024-08-06}) and Gemini 2.5 Flash (model ID: \texttt{gemini-2.5-flash-002}), both with temperature 0.1 for maximum consistency in scoring, top-p 1.0, and max tokens 4096. Judges received structured prompts with ground-truth context and evaluated outputs using predefined rubrics (0-5 scale for explanation quality dimensions, binary for law matching).

\paragraph{Packages} For \texttt{location\_alignment} evaluation, we used the following Python packages: (1) \texttt{rouge-score} version 0.1.2 (Google Research implementation) for ROUGE-1, ROUGE-2, and ROUGE-L metrics with default tokenization and stemming enabled; (2) \texttt{nltk} version 3.8.1 with the METEOR implementation (\texttt{nltk.translate.meteor\_score}) using default parameters and English language resources; and (3) \texttt{bert-score} version 0.3.13 for computing BERT F1 scores using the \texttt{microsoft/deberta-xlarge-mnli} model as the reference embedder with default rescaling. All metrics computed sentence-level alignments with normalization for case and punctuation.

\paragraph{Data Processing and Analysis} For data extraction and statistical analysis, we used \texttt{pandas} version 2.1.0 for DataFrame operations and CSV/JSON manipulation, \texttt{numpy} version 1.24.3 for numerical computations, and \texttt{openpyxl} version 3.1.2 for reading Excel-format evaluation results. Text preprocessing employed \texttt{re} (Python standard library) for regex-based cleaning and \texttt{json} (Python standard library) for structured data parsing. Graph-based alignment for \texttt{location\_alignment} used custom implementations leveraging \texttt{networkx} version 3.1 for connected component detection.

\paragraph{Reproducibility} All experiments were conducted using Python 3.10.12 on Ubuntu 22.04 LTS. Random seeds were not set for LLM API calls due to model-side non-determinism, but generation parameters were held constant across all experimental conditions. Complete implementation code, evaluation scripts, and detailed hyperparameter configurations are available in our public repository to ensure full reproducibility.

\subsection{Human Validation and Legal Expert Annotation}
\label{sec:expert-annotation}
\textbf{LLM-as-judge vs. Human Expert}. To see whether the LLM-as-judge truly applies the defined rubrics correctly and how this aligns with human expert judgment, we have conducted a comparison study between LLM judges’ outputs against human-curated ground-truth labels using a parallel evaluation process on a sample (15\% of the output sample) of the output generated. We used the human-validated ground-truth JSON file as the reference for scoring. Each file contains the corresponding ground-truth explanation, along with metadata linking them to specific discrepancy categories. Example snippet of such a json file is given at the end of the paper.

So, we use this human validated ground truth file explanation along with the other metadata and the predicted explanations and humans score the predicted explanations on the 4 parameters. The human experts assign the scores with all prior knowledge about the parameter definitions and the task. Here, legal experts are not required because we have to just match two different explanations and scoring them based on 4 different parameters which is an ideal task for NLP experts to take on.

Now below is the table~\ref{tab:llm_vs_human} where we have reported the average difference of human and LLM as a judge score for the 4 parameters for a sample of the predicted results:

\begin{table}[h]
\centering
\small
\begin{tabular}{lccc}
\hline
\textbf{Metric} & \textbf{LLM-as-Judge} & \textbf{Expert} & \textbf{Diff.} \\
\hline
Accuracy          & 3.12 & 3.28 & +0.16 \\
Completeness      & 2.52 & 2.72 & +0.20 \\
Clarity           & 4.04 & 3.76 & $-0.28$ \\
Legal Reasoning   & 3.16 & 2.92 & $-0.24$ \\
\hline
\end{tabular}
\caption{Comparison between LLM-as-Judge and Human Expert evaluations. All the reported values are mean values.}
\label{tab:llm_vs_human}
\end{table}

\textbf{Legal Expert Validation}. We engaged a licensed legal practitioner to audit approximately 10 perturbed documents from each of the 10 discrepancy categories ($\approx$100 total). Due to financial constraints, we were unable to extend the legal verification to a larger portion of the dataset. The expert reviewed these samples independently and confirmed that the perturbations were legally plausible, contextually coherent, and representative of real-world contractual risk patterns such as omissions, ambiguities, and misaligned terminology. The legal expert evaluated the perturbed documents based on the following four primary parameters:

\textbf{Legal Plausibility (LP)} -- Whether the perturbation represented a legally coherent and realistic modification that could occur in actual contract drafting or review scenarios.\\

\textbf{Contextual Coherence (CC)} -- Whether the changed clause remained semantically and contextually consistent within the overall document structure.\\

\textbf{Contradiction Strength and Significance (CSS)} -- Whether the modification created a clear, unambiguous conflict, either internal (in-text) or external (statutory), that could plausibly affect compliance.\\

\textbf{Representativeness of Legal Risk Type (RLR)} -- Whether the perturbation accurately reflected real-world contractual risk patterns such as omissions, ambiguities, misaligned terminology, inconsistencies, or structural flaws.\\

The expert rated each perturbation on a 0–5 scale across four above mentioned key parameters:
0 = Not acceptable, 5 = Fully satisfactory.
The scores shown in Table~\ref{tab:expert_scores_contradiction} represent the averaged ratings across the reviewed samples.

\begin{table}[h]
\centering
\small
\begin{tabular}{l l c}
\hline
\textbf{Parameter} & \textbf{Contradiction Type} & \textbf{Expert Score} \\
\hline
LP & Outer-law Contradiction & 4.83 \\
LP & In-text Contradiction   & 4.89 \\
CC & Outer-law Contradiction & 4.56 \\
CC & In-text Contradiction   & 4.43 \\
CSS & Outer-law Contradiction & 4.49 \\
CSS & In-text Contradiction   & 4.54 \\
RLR & Outer-law Contradiction & 4.79 \\
RLR & In-text Contradiction   & 4.85 \\
\hline
\end{tabular}
\caption{Average expert evaluation scores (0--5) across contradiction types. Expert scores are averaged out. The abbreviations used are Legal Plausibility (LP), Contextual Coherence (CC), Contradiction Strength and Significance (CSS), and Representativeness of Legal Risk Type (RLR).}
\label{tab:expert_scores_contradiction}
\end{table}

\subsection{Extended Data Statistics}
\label{sec:stats-extend}
In this sub-section we provide a detailed overview of the average text lenth before and after perturbation. We computed the character lengths of each original\_text and changed\_text entry across all ground-truth JSON files. The cumulative statistics for each dataset variant are summarized below in Table~\ref{tab:text_length_stats}.

\begin{table}[h]
\centering
\small
\begin{tabular}{l l c c c}
\hline
\textbf{Dataset} & \textbf{Cat} & \textbf{Org. TL} & \textbf{Changed TL} & \textbf{Diff.} \\
\hline
CUAD & In-Text & 546.2 & 496.7 & 49.5 \\
CUAD & Legal   & 510.7 & 503.8 & 6.90 \\
NLI  & In-Text & 470.5 & 414.0 & 56.5 \\
NLI  & Legal   & 431.6 & 387.3 & 44.3 \\
\hline
\end{tabular}
\caption{Average text length statistics before and after perturbation. TL denotes text length.}
\label{tab:text_length_stats}
\end{table}

\subsection{Data Generation Prompts and Validation}
\label{sec:appendix-generation}

Our dataset generation employs carefully designed prompts that guide language models to create realistic legal contradictions across ten perturbation categories. Each prompt incorporates persona-based instructions (e.g., ``You are a senior compliance officer''), domain-specific definitions, step-by-step modification guidelines, and few-shot examples demonstrating the target perturbation type. For in-text contradictions, prompts instruct the model to introduce internal document inconsistencies by modifying definitions, creating conflicting obligations, or disrupting structural coherence. For legal contradictions, prompts require the model to identify applicable jurisdiction, introduce modifications that violate specific statutes, and provide proper legal citations with two verified URLs to official government sources.

The validation pipeline employs specialized verification prompts that assess contradiction strength through retrieval-augmented analysis. Each perturbation undergoes binary classification (YES/NO) where the model evaluates three dimensions: semantic significance of the modification, strength and directness of the contradiction, and contextual coherence within the document. For legal contradictions, we scrape legal text from the provided URLs and augment the verification prompt with this statutory context, enabling the model to ground its assessment in actual legal language rather than parametric knowledge alone. Only perturbations receiving a YES classification—indicating strong, unambiguous contradictions—are retained in the final benchmark. This multi-stage approach, combining generation, retrieval-enhanced verification, and human validation (Section 2), ensures our benchmark contains high-quality contradictions that reflect genuine legal risks rather than superficial or arguable inconsistencies.

The following pages present representative prompts from our pipeline, illustrating how we guide models to create specific perturbation types while maintaining legal fidelity. We include examples for ambiguity generation (both in-text and legal), inconsistency generation (both types), and the validation prompts used in our quality control process. Additional prompts for omissions, misaligned terminology, and structural flaws follow similar patterns with category-specific adaptations.

\section{Example Prompts and Outputs}
\label{sec:appendix-prompts}

This section provides example prompts and model outputs for each evaluation task.

\begin{tcolorbox}[breakable,colback=lightestgray,colframe=black,title=\textbf{Ambiguity In-Text Data Generation Prompt}]
You are a contract analyst ensuring consistency in legal agreements. Your task is to modify a document by introducing conflicting definitions of the same term within different sections.

\subsection*{\textbf{Definition:}}
Ambiguities also occur when key terms are \textbf{vaguely and inconsistently defined within the document itself}, creating internal contradictions. This type of \textbf{in-text contradiction} confuses contract enforcement by allowing multiple interpretations of the same term in different sections, leading to potential legal disputes over meaning.

\subsection*{\textbf{Step-by-Step Instructions:}}
\begin{enumerate}
    \item Identify a \textbf{key term} in the contract.
    \item Modify its definition in different sections so that they \textbf{conflict}.
    \item Ensure the contradiction creates \textbf{uncertainty in enforcement}.
    \item For that perturbation, make sure in the file there should be \textbf{6} of them.
    \item Output the modified contract in structured JSON format.
    \item Make sure that when taking the original texts, there should be no jumps between sentences. Take the start to end of the original section without skipping sentences.
    \item Ensure that the In-Text contradiction being made, due to the inserted perturbation, is being referred in the explanation on where the contradictions are taking place, sure the the inserted perturbation.
    \item For each perturbation:
\end{enumerate}
\begin{itemize}
    \item Record the \textbf{location} (section or paragraph number) where the term was changed.
    \item Also identify and record the \textbf{contradicted\_location} — this is where the contradiction appears or becomes contradictory due to the change. This could be another section where the original definition is expected to apply or due to the change it creates an in-text contradiction in this location.
    \item If the contradiction affects multiple sections, include the \textbf{most directly impacted} one.
    \item Extract the \textbf{full contradicted\_text} from that location — this should be the complete section or paragraph that reflects the conflict caused by the intext contradiction perturbation. Include the entire section to preserve legal context.
    \item Also keep in mind the contradicted\_location, doesn't have to be the same location where the perturbation was made. A perturbation can be made in one location and contradict somewhere in the text in another location
\end{itemize}

\subsection*{\textbf{Examples of Conflicting Definitions of a Term:}}

\textbf{Example 1:}
\begin{itemize}
    \item \textbf{Section 2:} ``The term `Client' refers to any party who purchases our services.''
    \item \textbf{Section 8:} ``Clients are defined as parties who hold an active contract for at least six months.''
    \item \textbf{Explanation:} The first definition includes all buyers, while the second restricts ``Clients'' to long-term customers, leading to disputes over \textbf{contractual obligations}.
\end{itemize}

\textbf{Example 2:}
\begin{itemize}
    \item \textbf{Section 1:} ``Employees are eligible for health benefits upon hiring.''
    \item \textbf{Section 6:} ``Employees are eligible for health benefits after 90 days.''
    \item \textbf{Explanation:} The contradiction makes it unclear when benefits apply.
\end{itemize}

\textbf{Example 3:}
\begin{itemize}
    \item \textbf{Section 4:} ``The vendor is responsible for product warranties.''
    \item \textbf{Section 9:} ``Warranty claims shall be processed by the manufacturer.''
    \item \textbf{Explanation:} It's unclear \textbf{who holds warranty responsibility}.
\end{itemize}

\textbf{Example 4:}
\begin{itemize}
    \item \textbf{Section 3:} ``Data shall be stored for five years.''
    \item \textbf{Section 12:} ``Data must be deleted within two years.''
    \item \textbf{Explanation:} A \textbf{conflicting data retention period} could violate regulations.
\end{itemize}

\textbf{Example 5:}
\begin{itemize}
    \item \textbf{Section 2:} ``A shareholder is anyone holding 5\% or more equity.''
    \item \textbf{Section 7:} ``A shareholder is anyone with voting rights.''
    \item \textbf{Explanation:} A shareholder under one section may \textbf{not qualify in another}.
\end{itemize}

\textbf{Example 6:}
\begin{itemize}
    \item \textbf{Original:} ``The Supplier shall deliver all goods in accordance with the specifications outlined in Schedule A. Any failure to meet these specifications will be considered a material breach of contract. In such cases, the Buyer reserves the right to request a full refund or demand replacement goods within 30 days. The Supplier guarantees that all goods will conform to the quality and performance standards described in the agreement.''
    \item \textbf{Modified:} ``The Supplier shall make every effort to deliver goods in accordance with the specifications outlined in Schedule A. If specifications are not fully met, the Buyer and Supplier will engage in discussions to determine an appropriate resolution, which may include partial refunds or adjusted delivery timelines.''
    \item \textbf{Explanation:} This change introduces ambiguity by replacing `shall deliver' with `shall make every effort,' making performance non-binding. The original contract clearly stated that failure to meet specifications was a material breach with predefined consequences, whereas the modified version introduces an uncertain resolution process. This creates an \textbf{internal contradiction} if another part of the contract enforces strict adherence to Schedule A's specifications.
\end{itemize}

\subsection*{\textbf{Return JSON Format}}

{\small
\texttt{"file\_name": \{file\_name\},\\\\
"perturbation": \\\\
"type": "Ambiguities - In Text Contradiction",\\\\
"original\_text": "EXCERPT BEFORE CHANGE",\\\\
"changed\_text": "EXCERPT AFTER CHANGE", \\\\
"explanation": "WHY THIS CHANGE INTRODUCES A PERTURBATION AND SPECIFIC PLACE WHERE IT IS CONTRADICTING WITHIN THE TEXT",\\\\
"location": "SECTION OR PARAGRAPH NUMBER WHERE THE CHANGE WAS MADE",\\\\
"contradicted\_location": "SECTION OR PARAGRAPH NUMBER WHERE THE CONFLICT OCCURS DUE TO THIS CHANGE",\\\\
"contradicted\_text": "FULL TEXT OF THE SECTION THAT IS CONTRADICTED DUE TO THE CHANGE"\\\\}
}

Below is the original legal text:
\texttt{-------------------}\\
\texttt{\{original\_text\}}\\
\texttt{-------------------}

Now, return ONLY the structured JSON object with the modified text and explanation.
\end{tcolorbox}

\begin{tcolorbox}[breakable,colback=lightestgray,colframe=black,title=\textbf{Ambiguity Legal Data Generation Prompt}]
You are a senior compliance officer reviewing a legal contract. Your task is to modify a section of the contract by introducing an ambiguous legal obligation while ensuring that the ambiguity contradicts a state or national law.

Before modifying the text:
\begin{itemize}
    \item \textbf{Read the file} to determine what city, state, or country the contract applies to.
    \item If the jurisdiction is unclear, default to \textbf{United States law}.
    \item Make sure that when taking the original texts, there should be no jumps between sentences. Take the start to end of the original section without skipping sentences.
\end{itemize}

\subsection*{\textbf{Definition:}}
Ambiguities occur when a legal statement is vague, leading to multiple interpretations. A \textbf{legal contradiction} under this category happens when an obligation is introduced ambiguously, making it difficult to enforce under state or national law. This can result in non-compliance with regulatory requirements, leaving legal obligations open to dispute.

\subsection*{\textbf{Step-by-Step Instructions:}}
\begin{enumerate}
    \item Identify a clear legal obligation in the contract.
    \item Modify the wording to make it \textbf{vague or open to multiple interpretations}.
    \item Ensure that this ambiguity creates \textbf{non-compliance with a specific law} in the identified jurisdiction.
    \item Use online legal databases or verified government resources to \textbf{find a real law that is contradicted}.
    \item For that contradiction, extract:
\end{enumerate}
\begin{itemize}
    \item \textbf{The citation/title of the law} (e.g., ``29 CFR § 516.2'' or ``CA Labor Code § 201'')
    \item \textbf{Two direct links (URLs)} to official or government sources for the law (\textbf{these links must be accessible. The relevant citation should be inside this URL. The page should be found - we should not get a Page Not Found or similar error.})
    \item \textbf{The first URL must come from an official government or legislature domain (ends in .gov, .mil, .state.<XX>.us, or one of uscode.house.gov, ecfr.gov, govinfo.gov, law.cornell.edu). Use the most recent version of the statute/regulation.}
    \item \textbf{The second URL should be a secondary link to serve as a backup in case the first link falters, but we wish for that to not happen.}
    \item \textbf{A brief explanation} of how the modified text contradicts the law (\textbf{law\_explanation})
    \item Make 5 such perturbations for the file.
    \item Output only the structured JSON object as shown below.
\end{itemize}

\subsection*{\textbf{Examples of Ambiguous Legal Obligations:}}

\textbf{Example 1:}
\begin{itemize}
    \item \textbf{Original:} ``The company shall provide necessary accommodations for disabled employees.''
    \item \textbf{Modified:} ``The company shall provide accommodations for disabled employees as deemed appropriate.''
    \item \textbf{Explanation:} The term ``as deemed appropriate'' introduces ambiguity, conflicting with \textbf{ADA (Americans with Disabilities Act)}, which mandates \textbf{clear, non-discretionary accommodations}.
\end{itemize}

\textbf{Example 2:}
\begin{itemize}
    \item \textbf{Original:} ``All contractors must comply with local zoning laws.''
    \item \textbf{Modified:} ``All contractors must make reasonable efforts to comply with zoning laws.''
    \item \textbf{Explanation:} ``Reasonable efforts'' is vague—some zoning laws require strict adherence.
\end{itemize}

\textbf{Example 3:}
\begin{itemize}
    \item \textbf{Original:} ``The landlord shall ensure habitable living conditions in compliance with state law.''
    \item \textbf{Modified:} ``The landlord shall make efforts to maintain habitable conditions.''
    \item \textbf{Explanation:} ``Make efforts'' does not guarantee habitability, which violates \textbf{tenant protection laws}.
\end{itemize}

\textbf{Example 4:}
\begin{itemize}
    \item \textbf{Original:} ``The company shall maintain data security measures that meet industry standards.''
    \item \textbf{Modified:} ``The company shall maintain data security measures that it deems sufficient.''
    \item \textbf{Explanation:} ``Deems sufficient'' is subjective and contradicts \textbf{GDPR and CCPA} requirements for \textbf{specific security standards}.
\end{itemize}

\textbf{Example 5:}
\begin{itemize}
    \item \textbf{Original:} ``Employees shall be provided meal breaks as required by law.''
    \item \textbf{Modified:} ``Employees shall be encouraged to take meal breaks.''
    \item \textbf{Explanation:} Some states require \textbf{mandatory} meal breaks (e.g., California).
\end{itemize}

\textbf{Example 6:}
\begin{itemize}
    \item \textbf{Original:} ``The Contractor shall comply with all federal and state regulations governing workplace safety, ensuring all necessary precautions are taken to protect employees from occupational hazards. The Contractor must conduct quarterly safety inspections and submit reports to regulatory authorities. Any violations of safety standards shall result in corrective actions and potential penalties. The company's leadership is responsible for ensuring full compliance at all levels.''
    \item \textbf{Modified:} ``The Contractor should make reasonable efforts to comply with applicable federal and state regulations governing workplace safety. The Contractor may conduct periodic safety inspections and submit reports when deemed necessary. Violations of safety standards will be assessed on a case-by-case basis, and corrective actions may be recommended where appropriate.''
    \item \textbf{Explanation:} This change weakens the legal obligation by replacing `shall comply' with `should make reasonable efforts,' making compliance discretionary rather than mandatory. The removal of `quarterly safety inspections' eliminates a clear legal requirement, and replacing `shall result in corrective actions' with `may be recommended' creates uncertainty. This contradicts \textbf{OSHA regulations}, which mandate strict compliance and routine reporting on workplace safety violations.
\end{itemize}

\subsection*{\textbf{Return JSON Format}}

{\small
\texttt{"file\_name": \{file\_name\},\\\\
"perturbation": \\\\
"type": "Ambiguities - Ambiguous Legal Obligation",\\\\
"original\_text": "EXCERPT BEFORE CHANGE",\\\\
"changed\_text": "EXCERPT AFTER CHANGE",\\\\
"explanation": "WHY THIS CHANGE INTRODUCES A PERTURBATION",\\\\
"contradicted\_law": "SPECIFIC LAW OR REGULATION BEING VIOLATED",\\\\
"law\_citation": "TITLE OR SECTION OF THE LAW (e.g., '29 CFR § 516.2')",\\\\
"law\_url1": ["OFFICIAL\_LEGAL\_REFERENCE\_URL\_1"],\\\\
"law\_url2": ["OFFICIAL\_LEGAL\_REFERENCE\_URL\_2"],\\\\
"law\_explanation": "HOW AND WHY THIS MODIFICATION CONTRADICTS THAT LAW",\\\\
"location": "SECTION OR PARAGRAPH NUMBER"\\\\}
}

Below is the original legal text:
\texttt{-------------------}\\
\texttt{\{original\_text\}}\\
\texttt{-------------------}

Now, return ONLY the structured JSON object with the modified text and explanation.
\end{tcolorbox}

\begin{tcolorbox}[breakable,colback=lightestgray,colframe=black,title=\textbf{Inconsistency In-Text Data Generation Prompt}]
You are a financial contract auditor ensuring consistency in payment terms. Your task is to introduce a conflict between different payment deadline clauses.

\subsection*{\textbf{Definition:}}
Inconsistencies also lead to \textbf{in-text contradictions} when \textbf{different sections of a contract provide conflicting deadlines, obligations, or penalties}. This creates in-text contradiction regarding which terms should be enforced, leading to disputes over contractual obligations.

\subsection*{\textbf{Step-by-Step Instructions:}}
\begin{enumerate}
    \item Identify a \textbf{key term} in the contract.
    \item Modify its definition in different sections so that they \textbf{conflict}.
    \item Ensure the contradiction creates \textbf{uncertainty in enforcement}.
    \item For that perturbation, make sure in the file there should be \textbf{6} of them.
    \item Output the modified contract in structured JSON format.
    \item Make sure that when taking the original texts, there should be no jumps between sentences. Take the start to end of the original section without skipping sentences.
    \item Ensure that the In-Text contradiction being made, due to the inserted perturbation, is being referred in the explanation on where the contradictions are taking place, sure the the inserted perturbation.
    \item For each perturbation:
\end{enumerate}
\begin{itemize}
    \item Record the \textbf{location} (section or paragraph number) where the term was changed.
    \item Also identify and record the \textbf{contradicted\_location} — this is where the contradiction appears or becomes contradictory due to the change. This could be another section where the original definition is expected to apply or due to the change it creates an in-text contradiction in this location.
    \item If the contradiction affects multiple sections, include the \textbf{most directly impacted} one.
    \item Extract the \textbf{full contradicted\_text} from that location — this should be the complete section or paragraph that reflects the conflict caused by the omission intext contradiction perturbation. Include the entire section to preserve legal context.
    \item Also keep in mind the contradicted\_location, doesn't have to be the same location where the perturbation was made. A perturbation can be made in one location and contradict somewhere in the text in another location
\end{itemize}

\subsection*{\textbf{Examples of Conflicting Payment Terms:}}

\textbf{Example 1:}
\begin{itemize}
    \item \textbf{Section 4:} ``Invoices must be paid within 45 days.''
    \item \textbf{Section 12:} ``Late fees apply if invoices are unpaid after 30 days.''
    \item \textbf{Explanation:} One section allows \textbf{45 days}, but the other \textbf{charges late fees after 30 days}, creating ambiguity.
\end{itemize}

\textbf{Example 2:}
\begin{itemize}
    \item \textbf{Section 6:} ``Security deposits are fully refundable.''
    \item \textbf{Section 9:} ``Security deposits are non-refundable.''
    \item \textbf{Explanation:} It is unclear whether \textbf{deposits must be returned}.
\end{itemize}

\textbf{Example 3:}
\begin{itemize}
    \item \textbf{Section 3:} ``Early termination fees apply if the contract is ended before 12 months.''
    \item \textbf{Section 10:} ``Clients may terminate at any time without penalty.''
    \item \textbf{Explanation:} Conflicting clauses \textbf{cause uncertainty about termination fees}.
\end{itemize}

\textbf{Example 4:}
\begin{itemize}
    \item \textbf{Section 5:} ``Payment plans must be completed within 6 months.''
    \item \textbf{Section 11:} ``Customers may extend payment plans up to 12 months.''
    \item \textbf{Explanation:} Different sections \textbf{contradict the maximum payment period}.
\end{itemize}

\textbf{Example 5:}
\begin{itemize}
    \item \textbf{Section 8:} ``Interest on late payments accrues at 5\% per month.''
    \item \textbf{Section 14:} ``No interest will be charged on late payments.''
    \item \textbf{Explanation:} \textbf{Conflicting interest policies} could lead to legal disputes.
\end{itemize}

\textbf{Example 6:}
        \begin{itemize}
    \item \textbf{Original:} ``For the purposes of this Agreement, the term `Affiliate' shall refer to any entity that is controlled by, controls, or is under common control with either party. The term `control' shall mean ownership of at least fifty-one percent (51\%) of the voting equity in such entity. Affiliates shall be entitled to the same rights and obligations as the contracting party.''
    \item \textbf{Modified:} ``For the purposes of this Agreement, the term `Affiliate' shall refer to any entity that is affiliated with, works in conjunction with, or has a strategic partnership with either party. Affiliates may be granted certain rights and obligations as determined by the contracting party on a case-by-case basis.''
    \item \textbf{Explanation:} The modified text \textbf{creates an in-text contradiction} by \textbf{changing the definition of "Affiliate"} to include strategic partnerships and loosely affiliated entities, rather than just entities \textbf{under majority control}. This contradicts other contract sections that \textbf{rely on the original definition} (51\% ownership) to determine rights and obligations. Additionally, \textbf{allowing case-by-case determination} undermines the original intent of treating affiliates the same as contracting parties, introducing \textbf{legal uncertainty} and \textbf{potential disputes over obligations}.
\end{itemize}

\subsection*{\textbf{Return JSON Format}}

{\small
\texttt{"file\_name": \{file\_name\},\\\\
"perturbation": \\\\
"type": "Inconsistencies - In Text Contradiction",\\\\
"original\_text": "EXCERPT BEFORE CHANGE",\\\\
"changed\_text": "EXCERPT AFTER CHANGE",\\\\
"explanation": "WHY THIS CHANGE INTRODUCES A PERTURBATION AND SPECIFIC PLACE WHERE IT IS CONTRADICTING WITHIN THE TEXT",\\\\
"location": "SECTION OR PARAGRAPH NUMBER",\\\\
"contradicted\_location": "SECTION OR PARAGRAPH NUMBER WHERE THE CONFLICT OCCURS DUE TO THIS CHANGE",\\\\
"contradicted\_text": "FULL TEXT OF THE SECTION THAT IS CONTRADICTED DUE TO THE CHANGE"\\\\}
}

Below is the original legal text:
\texttt{-------------------}\\
\texttt{\{original\_text\}}\\
\texttt{-------------------}

Now, return ONLY the structured JSON object with the modified text and explanation.
\end{tcolorbox}

\begin{tcolorbox}[breakable,colback=lightestgray,colframe=black,title=\textbf{Inconsistency Legal Data Generation Prompt}]
You are an employment law specialist ensuring that contractual deadlines comply with legal regulations. Your task is to modify a timeline in the contract so that it contradicts state or federal laws.

Before modifying the text:
\begin{itemize}
    \item \textbf{Read the file} to determine what city, state, or country the contract applies to.
    \item If the jurisdiction is unclear, default to \textbf{United States law}.
    \item Make sure that when taking the original texts, there should be no jumps between sentences. Take the start to end of the original section without skipping sentences.
\end{itemize}

\subsection*{\textbf{Definition:}}
Inconsistencies arise when \textbf{time-sensitive obligations} in a contract do not align with legal requirements. A \textbf{legal contradiction} in this category happens when a contract sets \textbf{a deadline or requirement that violates federal or state law}, making the contractual terms unenforceable or illegal.

\subsection*{\textbf{Step-by-Step Instructions:}}
\begin{enumerate}
    \item Identify a \textbf{contractual deadline} that is regulated by law (e.g., payment terms, claims deadlines, notice periods).
    \item Modify the deadline to \textbf{conflict with state or federal legal requirements}.
    \item Ensure that the change \textbf{creates non-compliance} with regulatory standards.
    \item Use \textbf{online legal databases or verified government resources} to \textbf{find a real law} that is contradicted by this new timeline.
    \item For that contradiction, extract:
\end{enumerate}
\begin{itemize}
    \item \textbf{The citation/title of the law} (e.g., ``29 CFR § 516.2'' or ``CA Labor Code § 201'')
    \item \textbf{Two direct links (URLs)} to official or government sources for the law (\textbf{these links must be accessible. The relevant citation should be inside this URL. The page should be found - we should not get a Page Not Found or similar error.})
    \item \textbf{The first URL must come from an official government or legislature domain (ends in .gov, .mil, .state.<XX>.us, or one of uscode.house.gov, ecfr.gov, govinfo.gov, law.cornell.edu). Use the most recent version of the statute/regulation.}
    \item \textbf{The second URL should be a secondary link to serve as a backup in case the first link falters, but we wish for that to not happen.}
    \item \textbf{A brief explanation} of how the modified text contradicts the law (\textbf{law\_explanation})
    \item Make 5 such perturbations for the file.
    \item Output only the structured JSON object as shown below.
	\end{itemize}

\subsection*{\textbf{Examples of Conflicts with Regulatory Timelines:}}

\textbf{Example 1:}
\begin{itemize}
    \item \textbf{Original:} ``Employees must submit harassment claims within 15 days.''
    \item \textbf{Modified:} ``Employees must submit harassment claims within 5 days.''
    \item \textbf{Explanation:} Some states (e.g., California) require \textbf{at least 30 days} for harassment claims.
\end{itemize}

\textbf{Example 2:}
\begin{itemize}
    \item \textbf{Original:} ``Landlords must return security deposits within 21 days of lease termination.''
    \item \textbf{Modified:} ``Landlords must return security deposits within 60 days of lease termination.''
    \item \textbf{Explanation:} Many states \textbf{mandate 14-30 days} for deposit refunds.
\end{itemize}

\textbf{Example 3:}
\begin{itemize}
    \item \textbf{Original:} ``Customers have the right to cancel a contract within 10 days of signing.''
    \item \textbf{Modified:} ``Customers may cancel contracts within 48 hours.''
    \item \textbf{Explanation:} Some \textbf{consumer protection laws} require \textbf{at least 7-10 days} for cancellations.
\end{itemize}

\textbf{Example 4:}
\begin{itemize}
    \item \textbf{Original:} ``Workers must receive final wages within 72 hours of termination.''
    \item \textbf{Modified:} ``Workers will receive final wages at the company's discretion.''
    \item \textbf{Explanation:} Federal and state laws \textbf{require clear final paycheck deadlines}.
\end{itemize}

\textbf{Example 5:}
\begin{itemize}
    \item \textbf{Original:} ``Loan repayment plans must allow at least 90 days for late payments before default.''
    \item \textbf{Modified:} ``Loan repayment plans may declare default after 30 days of non-payment.''
    \item \textbf{Explanation:} Some \textbf{loan regulations} require longer grace periods.
\end{itemize}

\textbf{Example 6:}
\begin{itemize}
    \item \textbf{Original:} ``The term `Confidential Information' shall refer to any proprietary business, financial, and technical data... Confidential Information shall be protected for a period of five (5) years from the date of disclosure.''
    \item \textbf{Modified:} ``The term `Confidential Information' shall refer to sensitive business information disclosed by one party to the other. Each party shall determine what constitutes Confidential Information based on its internal policies. Confidentiality obligations shall remain in effect for a commercially reasonable period.''
    \item \textbf{Explanation:} Replacing a \textbf{clear five-year obligation} with ``commercially reasonable period'' creates \textbf{uncertainty}. Allowing each party to define confidentiality based on internal policies \textbf{contradicts} standard contract law principles requiring \textbf{definitive} confidentiality terms.
\end{itemize}

\subsection*{\textbf{Return JSON Format}}

{\small
\texttt{"file\_name": \{file\_name\},\\\\
"perturbation": \\\\
"type": "Inconsistencies - Legal Contradiction",\\\\
"original\_text": "EXCERPT BEFORE CHANGE",\\\\
"changed\_text": "EXCERPT AFTER CHANGE",\\\\
"explanation": "WHY THIS CHANGE INTRODUCES A PERTURBATION",\\\\
"contradicted\_law": "SPECIFIC LAW OR REGULATION BEING VIOLATED",\\\\
"law\_citation": "TITLE OR SECTION OF THE LAW (e.g., '29 CFR § 516.2')",\\\\
"law\_url1": ["OFFICIAL\_LEGAL\_REFERENCE\_URL\_1"],\\\\
"law\_url2": ["OFFICIAL\_LEGAL\_REFERENCE\_URL\_2"],\\\\
"law\_explanation": "HOW AND WHY THIS MODIFICATION CONTRADICTS THAT LAW",\\\\
"location": "SECTION OR PARAGRAPH NUMBER"\\\\}
}

Below is the original legal text:
\texttt{-------------------}\\
\texttt{\{original\_text\}}\\
\texttt{-------------------}

Now, return \textbf{ONLY} the structured JSON object with the modified text and explanation.
\end{tcolorbox}

\begin{tcolorbox}[breakable,colback=lightestgray,colframe=black,title=\textbf{In-Text LLM Validation Prompt}]
You are a contract analyst. Your task is to determine whether a strong \textbf{in-text contradiction} was introduced by a modification to the document. An in-text contradiction occurs when one part of a contract directly or implicitly conflicts with another part of the same document, leading to uncertainty, inconsistency, or unenforceable terms.

You are given:
\begin{itemize}
    \item The \textbf{type} of perturbation introduced
    \item The \textbf{original text} that was modified
    \item The \textbf{changed version} of that text
    \item A short \textbf{explanation} of how the change creates contradiction
    \item A \textbf{contradicted section} from elsewhere in the document where the contradiction is triggered
    \item The definitions of each perturbation type
\end{itemize}

\subsection*{\textbf{Perturbation Information:}}

\textbf{Perturbation Type:} \{perturbation.get(``type'', ``'') \}\\

\textbf{Original Text:} \{perturbation.get(``original\_text'', ``'') \}\\

\textbf{Changed Text:} \{perturbation.get(``changed\_text'', ``'') \}\\

\textbf{Perturbation Explanation:} \{perturbation.get(``explanation'', ``'') \}\\

\textbf{Contradicted Text:} \{perturbation.get(``contradicted\_text'', ``'') \}

\subsection*{\textbf{Perturbation Definitions:}}
\begin{enumerate}
    \item Ambiguities – In Text Contradiction: Ambiguities occur when key terms or obligations are defined inconsistently in different parts of a contract, making enforcement difficult or interpretation unclear.
    \item Omissions – In Text Contradiction: Omissions occur when essential language is removed or altered in one section, creating a contradiction with obligations or constraints stated in other sections.
    \item Misaligned Terminology - In Text Contradiction: Misaligned terminology also leads to \textbf{in-text contradictions} when the contract \textbf{uses multiple terms interchangeably without defining them}, leading to conflicting obligations.
    \item Structural Flaws - In Text Contradiction: Structural flaws also cause \textbf{in-text contradictions} when a clause is placed in \textbf{multiple locations with conflicting implications}. When contract sections contradict each other based on placement, it creates uncertainty about the obligations and rights of each party.
    \item Inconsistencies - In Text Contradiction: Inconsistencies also lead to \textbf{in-text contradictions} when \textbf{different sections of a contract provide conflicting deadlines, obligations, or penalties}. This creates in-text contradiction regarding which terms should be enforced, leading to disputes over contractual obligations.
\end{enumerate}

\subsection*{\textbf{Your Task (Strict Instructions):}}

\begin{enumerate}
    \item \textbf{Compare the changed text} with the \textbf{contradicted text}.
    \item Determine if the change creates a \textbf{clear and direct conflict, inconsistency, or contradiction} within the document. This must be based solely on the provided excerpts and follow the definitions stated above.
    \item Only assign \texttt{``contradiction\_exists'': ``YES''} if the change:
\end{enumerate}
\begin{itemize}
    \item Alters or narrows a definition, obligation, or entitlement that \textbf{directly breaks compatibility} with the contradicted section
    \item Makes enforcement unclear or creates a logical inconsistency
    \item The contradiction must be strong enough that a reasonable reader would immediately recognize the sections as incompatible.
    \item Direct, clear, and impactful contradictions that would cause confusion, unenforceability, or a legal issue within the document.
\end{itemize}

\textbf{Assign \texttt{``NO''} if:}
\begin{itemize}
    \item The changed clause still aligns with the contradicted section
    \item Any difference is minor, interpretative, or does not meaningfully impact understanding or enforcement
    \item If the contradiction is weak, subtle, open to interpretation, or would require legal expertise to identify
    \item If ever in Doubt of whether if it is a strong contradiction or not, ALWAYS assign NO.
\end{itemize}

\textbf{Additional Guidelines:}
\begin{itemize}
    \item Keep in mind that the Changed Text and Contradiction text might be in the same location. That doesn't necessarily mean there wasn't a contradiction. Read the explanation given and determine whether the Changed Text contradicts the Contradicted Text.
    \item Also Compare the original text and Change Text and see how the Change causes a contradiction to the contradicted text, based on the given explanation, and provide a ``YES'' or ``NO''.
\end{itemize}

\subsection*{\textbf{Contradiction Classification Guidelines:}}

\subsubsection*{\textbf{YES (Strong In-Text Contradiction)}}
Assign \texttt{``YES''} if the modified text:
\begin{itemize}
    \item Removes key terms or items that are later required or referenced
    \item Changes timing, eligibility, or definitions in a way that breaks consistency
    \item Introduces ambiguity where the contradicted clause expects certainty
    \item Adheres to one of the definitions above.
\end{itemize}

\textbf{Examples:}
\begin{itemize}
    \item Section 2 defines ``PRODUCT'' excluding `bags'; Section 5 requires the use of bags → \textbf{YES}
    \item Section 3: ``Benefits start on Day 1'' → Changed to: ``May be eligible after probation''; Section 6: ``Enroll in benefits in Week 1'' → \textbf{YES}
\end{itemize}

\subsubsection*{\textbf{NO (No Meaningful Contradiction)}}
Assign \texttt{``NO''} if:
\begin{itemize}
    \item The change is additive or softens the language but does not conflict
    \item Both clauses can be reasonably interpreted as compatible
    \item The contradiction is only apparent with legal inference, not internal inconsistency
\end{itemize}

\textbf{Examples:}
\begin{itemize}
    \item Changed: ``Data must be stored securely'' → ``Data must be stored securely and backed up'' — does not contradict another section → \textbf{NO}
    \item Changed: ``Contractors comply with local codes'' → ``Contractors comply with local and state codes''; Section requires compliance → \textbf{NO}
\end{itemize}

\subsection*{\textbf{Output Format (Strictly Follow This):}}

{\small
\texttt{\{\\\\
``contradiction\_exists'': ``Assign NO or YES'',\\\\
``justification'': ``A grounded explanation of how (and if) the modified text creates an in-text contradiction, referencing everything that is provided to give a binary score''\\\\
\}}
}

\textbf{Return ONLY this JSON object. No markdown, no notes, no commentary — just valid JSON with double quotes.}
\end{tcolorbox}

\begin{tcolorbox}[breakable,colback=lightestgray,colframe=black,title=\textbf{Legal LLM Validation Prompt}]
You are a legal verification expert. Your task is to determine how strongly a legal contradiction was created when a contract was modified. This contradiction must be evaluated based on legal standards, using the information provided.

You are given:
\begin{itemize}
    \item The \textbf{type} of perturbation introduced
    \item The \textbf{original contract text}
    \item The \textbf{changed (perturbed) version} of that text
    \item A short \textbf{perturbation explanation}, describing how the legal obligation was altered
    \item A \textbf{law explanation}, generated by an LLM, describing why it believes this change contradicts a specific law
    \item \textbf{Two scraped law snippets}, pulled from official legal sources (e.g., .gov/.org sites)
    \item The definitions for each type of Perturbation.
\end{itemize}

\subsection*{\textbf{Perturbation Information:}}

\textbf{Perturbation Type:} \{perturbation.get(``type'', ``'') \}\\

\textbf{Original Text:} \{perturbation.get(``original\_text'', ``'') \}\\

\textbf{Changed Text:} \{perturbation.get(``changed\_text'', ``'') \}\\

\textbf{Perturbation Explanation:} \{perturbation.get(``explanation'', ``'') \}\\

\textbf{Law Explanation (Generated by LLM):} \{perturbation.get(``law\_explanation'', ``'') \}\\

\textbf{Scraped Law Snippet 1:} \{perturbation.get(``scraped\_snippet\_1'', ``'') \}\\

\textbf{Scraped Law Snippet 2:} \{perturbation.get(``scraped\_snippet\_2'', ``'') \}

\subsection*{\textbf{Perturbation Definitions:}}
\begin{enumerate}
    \item Ambiguity Legal Contradiction: Ambiguities occur when a legal statement is vague, leading to multiple interpretations. A \textbf{legal contradiction} under this category happens when an obligation is introduced ambiguously, making it difficult to enforce under state or national law. This can result in non-compliance with regulatory requirements, leaving legal obligations open to dispute.
    \item Omission Legal Contradiction: Omissions occur when a contract \textbf{removes essential information}, creating legal loopholes. A \textbf{legal contradiction} in this category happens when a contract omits \textbf{a legally mandated consumer protection}, making it non-compliant.
    \item Misaligned Terminalogy Legal Contradiction: Misaligned terminology occurs when a contract \textbf{uses terms that do not match their statutory meaning}, leading to non-compliance. A \textbf{legal contradiction} in this category arises when contractual language deviates from legally recognized definitions, making the contract unenforceable.
    \item Structural Flaws Legal Contradiction: Structural flaws occur when the \textbf{organization of a contract affects its clarity or enforceability}. A \textbf{legal contradiction} in this category arises when a required legal disclosure is \textbf{placed in an irrelevant or misleading section}, making it difficult for parties to locate or interpret.
    \item Inconsistencies Legal Contradiction: Inconsistencies arise when \textbf{time-sensitive obligations} in a contract do not align with legal requirements. A \textbf{legal contradiction} in this category happens when a contract sets \textbf{a deadline or requirement that violates federal or state law}, making the contractual terms unenforceable or illegal.
\end{enumerate}

\subsection*{\textbf{Your Task (Strict Instructions):}}

\begin{enumerate}
    \item Analyze all the provided inputs thoroughly, along with the definitions to understand what perturbation type was used.
    \item Focus your contradiction assessment on the following 5 core elements:
\end{enumerate}
\begin{itemize}
    \item Scraped Law Snippet 1
    \item Scraped Law Snippet 2
    \item Perturbation Explanation
    \item Law Explanation
    \item The comparison between the Original and Changed Text, and how the Changed text creates a legal contradiction based on the explanation.
\end{itemize}

\textbf{Additional Guidelines:}
\begin{itemize}
    \item Use \textbf{both scraped law snippets}, even if one is missing, short, or incomplete. If both are valid, use both to strengthen your understanding of the law's scope and language. If one is missing, use the other one to strengthen your understanding of the law. If both are missing, use whatever else is given to produce your output.
\end{itemize}

\subsection*{\textbf{Contradiction Exists Category (Choose ONE: ``NO'' or ``YES'')}}

You must determine if a contradiction exists based on how significantly the modified contract text violates or undermines the legal obligation stated in the law.

This decision should consider:
\begin{itemize}
    \item The \textbf{importance and clarity} of the legal requirement in the scraped law
    \item How much the \textbf{modified text weakens, removes, or reverses} that legal requirement
    \item Whether the contradiction is \textbf{explicit and impactful}, or \textbf{subtle and arguable}
\end{itemize}

\subsubsection*{\textbf{Contradiction does NOT EXIST}}

Assign \texttt{NO} when:
\begin{itemize}
    \item The change introduces \textbf{minor weakening}, \textbf{softening}, or \textbf{interpretive ambiguity}
    \item The modified text still \textbf{preserves the legal intent}, even if it's more flexible or vague
    \item The contradiction is \textbf{context-dependent}, \textbf{non-obvious}, or requires legal interpretation to identify
    \item The contract could \textbf{reasonably be interpreted} as still being compliant
    \item If in doubt assign NO
\end{itemize}

\textbf{Examples:}
\begin{itemize}
    \item Adjustments that make language more discretionary or subjective
    \item Modifications that reduce precision, but not legal force
    \item Contradictions that rely on tone or nuance
\end{itemize}

\subsubsection*{\textbf{Contradiction EXISTS}}

Assign \texttt{YES} when:
\begin{itemize}
    \item The modification results in a \textbf{clear non-compliance} with the law
    \item The changed text \textbf{removes or contradicts core legal obligations}
    \item The contradiction is \textbf{direct, enforceable, and serious} from a legal or regulatory perspective
    \item The modified clause would likely \textbf{fail legal scrutiny} or \textbf{nullify required protections or standards}
    \item The contradiction falls under one of the definitions above.
\end{itemize}

\textbf{Examples:}
\begin{itemize}
    \item Contract change eliminates required action or accountability
    \item Legal mandates are replaced by optional or conflicting obligations
    \item Any change that would trigger a \textbf{legal violation} or breach of compliance
\end{itemize}

\textbf{Additional Requirements:}
\begin{itemize}
    \item Then provide a \textbf{justification}, explaining clearly:
    \begin{itemize}
        \item How the law was contradicted (or not)
        \item Which parts of the scraped snippets support your reasoning
        \item Why the contradiction score is appropriate, without simply repeating the score
    \end{itemize}
    \item Do not infer legal meaning beyond what's in the snippets or explanations. Only use what's provided.
    \item The contradiction\_exists and justification must be \textbf{independently valid} — don't base one on the other.
\end{itemize}

\subsection*{\textbf{Output Format (Strictly Follow This):}}

{\small
\texttt{\{\\\\
``contradiction\_exists'': ``Assign NO or YES'',\\\\
``justification'': ``A legally grounded explanation of how (and if) the modified text contradicts the law, referencing the scraped snippets if helpful.''\\\\
\}}
}

\textbf{Return ONLY this JSON object. No markdown, no notes, no commentary — just valid JSON with double quotes.}
\end{tcolorbox}

\begin{tcolorbox}[breakable,colback=lightestgray,colframe=black,title=\textbf{Eval\_1: Binary Discrepancy Detection Prompt}]
You are a U.S. contract attorney who answers concisely.

Please read the legal document below in full.

\textbf{Document:}
\texttt{``````\\
\{doc\}\\
``````}

Does this document contain \textbf{any} discrepancy? Reply with \textbf{only} `Yes' or `No'.
\end{tcolorbox}

\begin{tcolorbox}[breakable,colback=lightestgray,colframe=black,title=\textbf{Eval\_2: Document Classification Prompt}]
You are a legal expert specializing in U.S. law. You will read a legal document very carefully and classify it into one of the following two categories:

\textbf{1. In-text contradiction:} The document contains one or more statement or clause that contradicts another part of the same document.

\textbf{2. Outer-law contradiction:} The document contains one or more statement or clause that contradicts existing U.S. federal, state or city laws (or any other U.S. laws).

Classify the document below into one of the three categories. Respond with only one of the labels: `In-text contradiction' or `Outer-law contradiction' Do not provide any explanation. Write `In-text contradiction' if the document contains In-text contradiction and write `Outer-law contradiction' if the document contains Outer-law contradiction.

\textbf{Document:}
\texttt{``````\\
\{doc\_text\}\\
``````}

What type of contradiction does this document contain? Reply with only one of the two labels: `In-text contradiction' or `Outer-law contradiction'
\end{tcolorbox}

\begin{tcolorbox}[breakable,colback=lightestgray,colframe=black,title=\textbf{Eval\_3 L1: In-Text Discrepancy Detection Prompt}]
You are a senior U.S. attorney and contract analyst.

\subsection*{\textbf{Task:}}
Read the following legal document in full. Check and detect whether the document contains any intext discrepancies/contradictions in it.

\subsection*{\textbf{Definitions of intext discrepancy/contradiction:}}
\textbf{In Text Contradiction:} a statement, clause, term, or definition that conflicts with, negates, or is irreconcilable with another part of the \textbf{same} document.

There are five primary categories of intext discrepancies. Their definitions are provided below:

\subsection*{\textbf{Discrepancy Categories:}}

\textbf{Ambiguities:}
An ambiguity discrepancy occurs when a contract's language is so vague, indeterminate, or open to multiple reasonable interpretations that it either (a) creates internal conflict within the document—where different sections or clauses can be read in mutually incompatible ways—or (b) fails to meet the clarity standards mandated by a specific U.S. statute, regulation, law, or controlling case, thereby placing the clause in conflict with applicable legal requirements.

\textbf{Inconsistencies:}
An inconsistency discrepancy arises when a contract contains conflicting deadlines, obligations, or penalties—either within different sections of the document itself or between the contract and applicable legal standards. These contradictions can create uncertainty about which terms should be enforced or may render certain provisions unenforceable if they violate federal or state laws.

\textbf{Misaligned Terminology:}
A misaligned terminology discrepancy occurs when a contract uses different or undefined terms inconsistently—either by employing multiple terms interchangeably within the document, creating internal contradictions and conflicting obligations, or by deviating from legally established definitions, resulting in non-compliance with statutory or regulatory standards that can render the contract unenforceable.

\textbf{Omissions:}
An omission discrepancy arises when a required term, clause, definition, or disclosure is missing, incomplete, or left undefined—either when it is referenced elsewhere in the contract, causing internal inconsistency, or when its absence violates specific U.S. legal requirements, such as statutory mandates, regulatory disclosures, or protective safeguards.

\textbf{Structural Flaws:}
A structural flaw discrepancy occurs when the organization or placement of clauses within a contract undermines its clarity, coherence, or legal validity—such as when the same clause appears in multiple sections with conflicting implications, or when a legally required disclosure is positioned in a misleading or irrelevant part of the document, making it difficult for parties to locate, interpret, or enforce the intended obligations.

The document may contain in-text discrepancies belonging to any of the five defined types. Your task is to identify all such discrepancies and present them strictly in the output format specified below:

\subsection*{\textbf{Output Format:}}
Return \textbf{exactly} one JSON array—no markdown, no commentary—where each object includes:
\begin{itemize}
    \item \texttt{"text"}: the exact sentence(s) or phrase that contains the issue
    \item \texttt{"explanation"}: clear detailed reasoning—what it contradicts and why it is a discrepancy
\end{itemize}

\textbf{Example output shape:}
{\small
\texttt{[\\\\
\{\\\\
``text'': ``<exact sentence(s) or phrase with the issue>'',\\\\
``explanation'': ``<clear, detailed reason: what it contradicts and why it is a problem>''\\\\
\},\\\\
... (one object per discrepancy) ...\\\\
]}
}

\subsection*{\textbf{Rules:}}
\begin{itemize}
    \item Include \textbf{all} discrepancies you find.
    \item Assign the most appropriate \texttt{``main\_type''} and \texttt{``sub\_type''} to each discrepancy.
    \item If no discrepancy exists, return an empty array: \texttt{[]}.
    \item Do NOT wrap the JSON in markdown and do NOT add any extra keys or commentary.
\end{itemize}
\end{tcolorbox}

\begin{tcolorbox}[breakable,colback=lightestgray,colframe=black,title=\textbf{Eval\_3 L1: Legal Discrepancy Detection Prompt}]
You are a senior U.S. attorney and contract analyst.

\subsection*{\textbf{Task:}}
Read the following legal document in full. Check and detect whether the document contains any legal discrepancies/contradictions in it.

\subsection*{\textbf{Definitions of legal discrepancy/contradiction:}}
\textbf{Legal Contradiction:} any clause, section, or text that conflicts with, violates, or is unenforceable under existing U.S. law.

There are five primary categories of legal discrepancies. Their definitions are provided below:

\subsection*{\textbf{Discrepancy Categories:}}

\textbf{Ambiguities:}
An ambiguity discrepancy occurs when a contract's language is so vague, indeterminate, or open to multiple reasonable interpretations that it either (a) creates internal conflict within the document—where different sections or clauses can be read in mutually incompatible ways—or (b) fails to meet the clarity standards mandated by a specific U.S. statute, regulation, law, or controlling case, thereby placing the clause in conflict with applicable legal requirements.

\textbf{Inconsistencies:}
An inconsistency discrepancy arises when a contract contains conflicting deadlines, obligations, or penalties—either within different sections of the document itself or between the contract and applicable legal standards. These contradictions can create uncertainty about which terms should be enforced or may render certain provisions unenforceable if they violate federal or state laws.

\textbf{Misaligned Terminology:}
A misaligned terminology discrepancy occurs when a contract uses different or undefined terms inconsistently—either by employing multiple terms interchangeably within the document, creating internal contradictions and conflicting obligations, or by deviating from legally established definitions, resulting in non-compliance with statutory or regulatory standards that can render the contract unenforceable.

\textbf{Omissions:}
An omission discrepancy arises when a required term, clause, definition, or disclosure is missing, incomplete, or left undefined—either when it is referenced elsewhere in the contract, causing internal inconsistency, or when its absence violates specific U.S. legal requirements, such as statutory mandates, regulatory disclosures, or protective safeguards.

\textbf{Structural Flaws:}
A structural flaw discrepancy occurs when the organization or placement of clauses within a contract undermines its clarity, coherence, or legal validity—such as when the same clause appears in multiple sections with conflicting implications, or when a legally required disclosure is positioned in a misleading or irrelevant part of the document, making it difficult for parties to locate, interpret, or enforce the intended obligations.

The document may contain legal discrepancies belonging to any of the five defined types. Your task is to identify all such discrepancies and present them strictly in the output format specified below:

\subsection*{\textbf{Output Format:}}
Return \textbf{exactly} one JSON array—no markdown, no commentary—where each object includes:
\begin{itemize}
    \item \texttt{"text"}: the exact sentence(s) or phrase that contains the issue
    \item \texttt{"explanation"}: clear detailed reasoning—what it contradicts and why it is a discrepancy
    \item \texttt{"law"}: citation of the U.S. law / regulation / case it violates, or "N/A" if the conflict is only in-text or no law found.
\end{itemize}

\textbf{Example output shape:}
{\small
\texttt{[\\\\
\{\\\\
``text'': ``<exact sentence(s) or phrase with the issue>'',\\\\
``explanation'': ``<clear, detailed reason: what it contradicts and why it is a problem>'',\\\\
``law'': ``Cal. Com. Code § 2201''  // or ``N/A'' only if law not found\\\\
\},\\\\
... (one object per discrepancy) ...\\\\
]}
}

\subsection*{\textbf{Rules:}}
\begin{itemize}
    \item Include \textbf{all} discrepancies you find.
    \item Assign the most appropriate \texttt{``main\_type''} and \texttt{``sub\_type''} to each discrepancy.
    \item If no discrepancy exists, return an empty array: \texttt{[]}.
    \item Do NOT wrap the JSON in markdown and do NOT add any extra keys or commentary.
\end{itemize}
\end{tcolorbox}

\begin{tcolorbox}[breakable,colback=lightestgray,colframe=black,title=\textbf{Eval\_3 L2: In-Text Discrepancy Detection Prompt}]
You are a senior U.S. attorney and contract analyst.

\subsection*{\textbf{Task:}}
Read the following legal document in full. Check and detect whether the document contains any intext discrepancies/contradictions in it.

\subsection*{\textbf{Definitions of intext discrepancy/contradiction:}}
\textbf{In-Text Contradiction:} a statement, clause, term, or definition that conflicts with, negates, or is irreconcilable with another part of the \textbf{same} document.

There are five primary categories of intext discrepancies. Their definitions are provided below:

\subsection*{\textbf{Discrepancy Categories:}}

\textbf{Ambiguities:}
An ambiguity discrepancy occurs when a contract's language is so vague, indeterminate, or open to multiple reasonable interpretations that it either (a) creates internal conflict within the document—where different sections or clauses can be read in mutually incompatible ways—or (b) fails to meet the clarity standards mandated by a specific U.S. statute, regulation, law, or controlling case, thereby placing the clause in conflict with applicable legal requirements.

\textbf{Inconsistencies:}
An inconsistency discrepancy arises when a contract contains conflicting deadlines, obligations, or penalties—either within different sections of the document itself or between the contract and applicable legal standards. These contradictions can create uncertainty about which terms should be enforced or may render certain provisions unenforceable if they violate federal or state laws.

\textbf{Misaligned Terminology:}
A misaligned terminology discrepancy occurs when a contract uses different or undefined terms inconsistently—either by employing multiple terms interchangeably within the document, creating internal contradictions and conflicting obligations, or by deviating from legally established definitions, resulting in non-compliance with statutory or regulatory standards that can render the contract unenforceable.

\textbf{Omissions:}
An omission discrepancy arises when a required term, clause, definition, or disclosure is missing, incomplete, or left undefined—either when it is referenced elsewhere in the contract, causing internal inconsistency, or when its absence violates specific U.S. legal requirements, such as statutory mandates, regulatory disclosures, or protective safeguards.

\textbf{Structural Flaws:}
A structural flaw discrepancy occurs when the organization or placement of clauses within a contract undermines its clarity, coherence, or legal validity—such as when the same clause appears in multiple sections with conflicting implications, or when a legally required disclosure is positioned in a misleading or irrelevant part of the document, making it difficult for parties to locate, interpret, or enforce the intended obligations.

\subsection*{\textbf{Example of in-text contradiction/discrepancy}}

{\footnotesize
\texttt{\{\\\\
``original\_text'': ``1. DEFINITIONS. (a) `CONTENT' means all content or information, in any medium, provided by a party to the other party for use in conjunction with the performance of its obligations hereunder, including without limitation any text, music, sound, photographs, video, graphics, data or software. Content provided by 2TheMart is referred to herein as `2TheMart Content' and content provided by i-Escrow is referred to herein as `i-Escrow Content.'',\\\\
``changed\_text'': ``1. DEFINITIONS. (a) `CONTENT' means all content or information, in any medium, provided by a party to the other party for use in conjunction with the performance of its obligations hereunder, **including only text and images**. Content provided by 2TheMart is referred to herein as `2TheMart Content' and content provided by i-Escrow is referred to herein as `i-Escrow Content.'',\\\\
``explanation'': ``The revised definition limits `CONTENT' to text and images, whereas Section 6.1 later grants rights to `publicly perform' and `digitally perform' 2TheMart Content—language that presumes audio, video, or other media. The conflicting scopes create an ambiguity-driven in-text contradiction.'',\\\\
\}}
}

\textbf{Expected JSON entry:}
{\small
\texttt{[\\\\
\{\\\\
``text'': ``\`CONTENT\` means all content or information, in any medium, provided by a party to the other party for use in conjunction with the performance of its obligations hereunder, **including only text and images**. Content provided by 2TheMart is referred to herein as \`2TheMart Content\` and content provided by i-Escrow is referred to herein as \`i-Escrow Content.\`,\\\\
``explanation'': ``Definition excludes audio/video, yet Section 6.1 authorizes public/digital performance—an implicit reference to multimedia—so the provisions conflict.''\\\\
\}\\\\
]}
}

The document may contain in-text discrepancies belonging to any of the five defined types. Your task is to identify all such discrepancies and present them strictly in the output format specified below:

\subsection*{\textbf{Output Format:}}
Return \textbf{exactly} one JSON array—no markdown, no commentary—where each object includes:
\begin{itemize}
    \item \texttt{"text"}: the exact sentence(s) or phrase that contains the issue
    \item \texttt{"explanation"}: clear detailed reasoning—what it contradicts and why it is a discrepancy
\end{itemize}

\subsection*{\textbf{Rules:}}
\begin{itemize}
    \item Include \textbf{all} discrepancies you find.
    \item If no discrepancy exists, return an empty array: \texttt{[]}.
    \item Do NOT wrap the JSON in markdown and do NOT add any extra keys or commentary.
\end{itemize}
\end{tcolorbox}

\begin{tcolorbox}[breakable,colback=lightestgray,colframe=black,title=\textbf{Eval\_3 L2: Legal Discrepancy Detection Prompt}]
You are a senior U.S. attorney and contract analyst.

\subsection*{\textbf{Task:}}
Read the following legal document in full. Check and detect whether the document contains any legal discrepancies/contradictions in it.

\subsection*{\textbf{Definitions of legal discrepancy/contradiction:}}
\textbf{Legal Contradiction:} any clause, section, or text that conflicts with, violates, or is unenforceable under existing U.S. law.

There are five primary categories of legal discrepancies. Their definitions are provided below:

\subsection*{\textbf{Discrepancy Categories:}}

\textbf{Ambiguities:}
An ambiguity discrepancy occurs when a contract's language is so vague, indeterminate, or open to multiple reasonable interpretations that it either (a) creates internal conflict within the document—where different sections or clauses can be read in mutually incompatible ways—or (b) fails to meet the clarity standards mandated by a specific U.S. statute, regulation, law, or controlling case, thereby placing the clause in conflict with applicable legal requirements.

\textbf{Inconsistencies:}
An inconsistency discrepancy arises when a contract contains conflicting deadlines, obligations, or penalties—either within different sections of the document itself or between the contract and applicable legal standards. These contradictions can create uncertainty about which terms should be enforced or may render certain provisions unenforceable if they violate federal or state laws.

\textbf{Misaligned Terminology:}
A misaligned terminology discrepancy occurs when a contract uses different or undefined terms inconsistently—either by employing multiple terms interchangeably within the document, creating internal contradictions and conflicting obligations, or by deviating from legally established definitions, resulting in non-compliance with statutory or regulatory standards that can render the contract unenforceable.

\textbf{Omissions:}
An omission discrepancy arises when a required term, clause, definition, or disclosure is missing, incomplete, or left undefined—either when it is referenced elsewhere in the contract, causing internal inconsistency, or when its absence violates specific U.S. legal requirements, such as statutory mandates, regulatory disclosures, or protective safeguards.

\textbf{Structural Flaws:}
A structural flaw discrepancy occurs when the organization or placement of clauses within a contract undermines its clarity, coherence, or legal validity—such as when the same clause appears in multiple sections with conflicting implications, or when a legally required disclosure is positioned in a misleading or irrelevant part of the document, making it difficult for parties to locate, interpret, or enforce the intended obligations.

\subsection*{\textbf{Example of legal contradiction/discrepancy}}

{\footnotesize
\texttt{\{\\\\
``original\_text'': ``The parties agree to act in good faith with respect to each provision of this Agreement and any dispute that may arise related hereto.'',\\\\
``changed\_text'': ``The parties agree to attempt to act in good faith with respect to each provision of this Agreement and any dispute that may arise related hereto.'',\\\\
``explanation'': ``Changing `agree to act in good faith' to `agree to attempt to act in good faith' weakens the obligation. Good faith is a standard implied in contracts under California law, requiring honesty and fair dealing. The word `attempt' introduces ambiguity, implying that full good-faith efforts are not strictly required, thus undermining the implied covenant.'',\\\\
``contradicted\_law'': ``California Commercial Code § 1304'',\\\\
\}}
}

\textbf{Expected JSON entry:}
{\small
\texttt{[\\\\
\{\\\\
``text'': ``The parties agree to attempt to act in good faith with respect to each provision of this Agreement and any dispute that may arise related hereto.'',\\\\
``explanation'': ``The clause downgrades the statutory obligation of good faith to a mere attempt, conflicting with California Commercial Code § 1304.'',\\\\
``law'': ``Cal. Com. Code § 1304''\\\\
\}\\\\
]}
}

The document may contain legal discrepancies belonging to any of the five defined types. Your task is to identify all such discrepancies and present them strictly in the output format specified below:

\subsection*{\textbf{Output Format:}}
Return \textbf{exactly} one JSON array—no markdown, no commentary—where each object includes:
\begin{itemize}
    \item \texttt{"text"}: the exact sentence(s) or phrase that contains the issue
    \item \texttt{"explanation"}: clear detailed reasoning—what it contradicts and why it is a discrepancy
    \item \texttt{"law"}: citation of the U.S. law / regulation / case it violates, or "N/A" if the conflict is only in-text.
\end{itemize}

\subsection*{\textbf{Rules:}}
\begin{itemize}
    \item Include \textbf{all} discrepancies you find.
    \item If no discrepancy exists, return an empty array: \texttt{[]}.
    \item Do NOT wrap the JSON in markdown and do NOT add any extra keys or commentary.
\end{itemize}
\end{tcolorbox}

\section{Complete Evaluation Results}
\label{sec:appendix-evaluation}

This section presents comprehensive evaluation results across all three evaluation levels, organized by task complexity. For each evaluation, we provide detailed performance breakdowns across all perturbation categories, datasets, models, and prompt levels, enabling fine-grained analysis of model capabilities and failure modes.

\subsection{Evaluation 2: Miss and Extra Rates Across Categories}
\label{sec:appendix-eval2-miss-extra}

Table~\ref{tab:category_values} presents miss and extra rates for Evaluation 3 at both zero-shot (L1) and one-shot (L2) prompt levels, broken down by all ten perturbation categories and both datasets. The miss rate quantifies the fraction of ground-truth discrepancy spans that models failed to detect, while the extra rate measures the fraction of predicted spans that did not align with any ground-truth annotation. These complementary metrics reveal the precision-recall trade-off inherent in different prompting strategies.

Several critical patterns emerge from this data. First, one-shot prompting (L2) consistently reduces miss rates across nearly all configurations, indicating that example-guided inference improves recall. However, this improvement comes at the cost of elevated extra rates, suggesting that models become more liberal in their predictions when provided with examples, leading to increased false positives. Second, the CUAD dataset exhibits systematically higher extra rates than NLI, particularly for GPT-4o-mini and Gemini-2.5, indicating that CUAD's legal

 complexity may trigger more spurious detections. Third, certain categories prove exceptionally challenging: Omission$_{Legal}$ on NLI shows miss rates exceeding 80\% for GPT-4o-mini and LLaMa-3.3, while Structural Flaws demonstrate high variability across models, with Gemini-2.0 showing markedly better detection than LLaMa-3.3. These category-specific patterns suggest that different architectures encode distinct inductive biases about legal document structure and the nature of legal violations.

The pronounced L1-to-L2 shift in the precision-recall trade-off varies substantially across models. Gemini-2.0 exhibits relatively stable behavior, with modest changes in both metrics, suggesting a more conservative generation strategy less susceptible to prompt perturbations. In contrast, GPT-4o-mini and Gemini-2.5 show dramatic swings, with extra rates sometimes doubling from L1 to L2, indicating high sensitivity to prompt design. LLaMa-3.3's performance on certain categories (e.g., Structural Flaws$_{Legal}$ on CUAD with 72.3\% extra rate and 57.1\% miss rate at L1) reveals fundamental brittleness in handling complex legal structural violations, likely due to training distribution gaps. Overall, this table underscores that no single prompting strategy dominates: practitioners must balance detection completeness against precision depending on their application's tolerance for false positives versus false negatives.

\begin{table*}[h!]
\setlength{\aboverulesep}{0pt}
\setlength{\belowrulesep}{0pt}
\centering
\small
\renewcommand{\arraystretch}{1.15}
\setlength{\tabcolsep}{4pt}
\begin{tabular}{l|cccc|cccc|cccc|cccc}
\hline
\multirowcell{3}{\textbf{Category}} & 
\multicolumn{4}{c|}{\textbf{GPT-4o mini}} &
\multicolumn{4}{c|}{\textbf{Gemini 2.0}} &
\multicolumn{4}{c|}{\textbf{Gemini 2.5}} &
\multicolumn{4}{c}{\textbf{LLaMa 3.3}} \\
\cline{2-17}
& \multicolumn{2}{c|}{\textbf{L1}} & \multicolumn{2}{c|}{\textbf{L2}} &
  \multicolumn{2}{c|}{\textbf{L1}} & \multicolumn{2}{c|}{\textbf{L2}} &
  \multicolumn{2}{c|}{\textbf{L1}} & \multicolumn{2}{c|}{\textbf{L2}} &
  \multicolumn{2}{c|}{\textbf{L1}} & \multicolumn{2}{c}{\textbf{L2}} \\
\cline{2-17}
& \textbf{E} & \textbf{M} & \textbf{E} & \textbf{M} &
  \textbf{E} & \textbf{M} & \textbf{E} & \textbf{M} &
  \textbf{E} & \textbf{M} & \textbf{E} & \textbf{M} &
  \textbf{E} & \textbf{M} & \textbf{E} & \textbf{M} \\
\cmidrule{1-17}
\rowcolor{lightestgray}
\cellcolor{white} & \multicolumn{16}{c}{\textbf{CUAD}} \\
Amb$_{\text{Leg}}$ & 89.7 & 90.4 & 70.9 & 91.5 & 67.9 & 55.6 & 65.6 & 56.8 & 82.6 & 50.4 & 77.8 & 43.3 & 91.8 & 83.4 & 86.4 & 79.4 \\
Amb$_{\text{inT}}$ & 76.7 & 72.1 & 73.7 & 72.4 & 57.2 & 58.6 & 56.0 & 57.2 & 77.8 & 48.7 & 71.5 & 53.5 & 76.8 & 63.8 & 72.6 & 60.6 \\
Incon$_{\text{Leg}}$ & 82.0 & 81.1 & 67.4 & 86.1 & 68.2 & 59.2 & 66.7 & 61.2 & 85.4 & 69.0 & 83.7 & 69.9 & 86.1 & 77.3 & 85.7 & 76.9 \\
Incon$_{\text{inT}}$ & 76.2 & 73.0 & 74.8 & 74.4 & 53.7 & 58.7 & 53.9 & 58.4 & 79.7 & 54.8 & 75.7 & 59.6 & 75.6 & 62.3 & 73.5 & 63.1 \\
MisTerm$_{\text{Leg}}$ & 79.3 & 76.8 & 64.5 & 81.5 & 53.9 & 41.9 & 50.6 & 43.1 & 76.9 & 41.8 & 69.5 & 65.8 & 80.4 & 67.4 & 80.9 & 68.4 \\
MisTerm$_{\text{inT}}$ & 83.1 & 80.0 & 83.4 & 81.3 & 69.6 & 68.9 & 70.0 & 69.6 & 80.8 & 55.3 & 78.7 & 60.9 & 82.9 & 74.0 & 80.4 & 73.0 \\
Omis$_{\text{Leg}}$ & 85.5 & 86.9 & 67.6 & 90.6 & 71.0 & 63.6 & 68.5 & 64.9 & 85.9 & 65.9 & 63.3 & 67.9 & 86.5 & 74.9 & 86.6 & 75.2 \\
Omis$_{\text{inT}}$ & 84.6 & 80.8 & 84.8 & 82.5 & 70.6 & 70.8 & 70.6 & 70.6 & 81.0 & 58.6 & 78.1 & 65.2 & 84.9 & 77.5 & 78.5 & 70.3 \\
StrFlaw$_{\text{Leg}}$ & 73.8 & 71.2 & 66.3 & 83.3 & 57.2 & 44.2 & 54.8 & 47.7 & 77.8 & 46.2 & 75.3 & 43.3 & 72.3 & 57.1 & 72.6 & 57.9 \\
StrFlaw$_{\text{inT}}$ & 72.1 & 68.3 & 70.0 & 70.7 & 45.0 & 48.8 & 46.4 & 49.0 & 75.6 & 42.9 & 66.9 & 47.1 & 79.2 & 67.9 & 69.3 & 56.0 \\
\cmidrule{1-17}
\rowcolor{lightestgray}
\cellcolor{white} & \multicolumn{16}{c}{\textbf{NLI}} \\
Amb$_{\text{Leg}}$ & 78.8 & 81.9 & 56.7 & 83.1 & 50.9 & 50.9 & 42.1 & 40.1 & 64.5 & 36.7 & 59.9 & 33.1 & 78.9 & 70.9 & 69.9 & 62.9 \\
Amb$_{\text{inT}}$ & 56.3 & 64.2 & 49.0 & 67.1 & 48.5 & 50.7 & 41.3 & 52.7 & 67.6 & 49.7 & 58.1 & 48.6 & 53.0 & 49.0 & 75.7 & 56.9 \\
Incon$_{\text{Leg}}$ & 57.1 & 70.7 & 43.3 & 75.5 & 36.1 & 45.8 & 33.9 & 45.4 & 61.7 & 50.6 & 61.1 & 52.8 & 62.3 & 58.9 & 62.0 & 61.1 \\
Incon$_{\text{inT}}$ & 59.0 & 66.0 & 52.6 & 70.9 & 50.2 & 52.9 & 40.8 & 53.4 & 66.0 & 49.1 & 62.0 & 51.8 & 49.3 & 47.7 & 48.3 & 48.6 \\
MisTerm$_{\text{Leg}}$ & 64.4 & 75.0 & 53.3 & 79.9 & 40.0 & 46.2 & 39.1 & 46.3 & 60.7 & 41.7 & 60.3 & 44.5 & 67.8 & 62.7 & 63.9 & 63.0 \\
MisTerm$_{\text{inT}}$ & 76.5 & 78.0 & 69.1 & 81.9 & 65.7 & 69.3 & 64.1 & 73.5 & 77.7 & 61.5 & 72.0 & 62.7 & 74.3 & 69.2 & 74.9 & 70.6 \\
Omis$_{\text{Leg}}$ & 75.4 & 81.2 & 52.1 & 86.0 & 56.3 & 62.8 & 54.9 & 63.9 & 71.9 & 62.4 & 70.3 & 63.6 & 74.3 & 67.7 & 72.2 & 68.0 \\
Omis$_{\text{inT}}$ & 73.0 & 75.5 & 59.5 & 79.2 & 66.7 & 67.7 & 61.5 & 72.4 & 78.0 & 63.4 & 74.6 & 65.8 & 71.0 & 63.2 & 71.3 & 65.9 \\
StrFlaw$_{\text{Leg}}$ & 65.3 & 49.9 & 56.3 & 57.8 & 56.9 & 33.2 & 57.7 & 37.8 & 70.7 & 27.2 & 68.8 & 29.0 & 71.5 & 40.9 & 68.7 & 41.6 \\
StrFlaw$_{\text{inT}}$ & 50.4 & 62.3 & 43.8 & 66.2 & 42.9 & 48.0 & 37.6 & 48.6 & 63.5 & 45.8 & 57.9 & 47.7 & 46.0 & 45.3 & 43.6 & 47.4 \\
\hline
\end{tabular}
\caption{\small Comparison across models and tasks for L1 (zero-shot) and L2 (one-shot) levels. Values represent extra (E) and miss (M) rates for each discrepancy category.}
\label{tab:category_values}
\end{table*}

\subsection{Evaluation 3: Location Alignment Metrics by Category}
\label{sec:appendix-eval3-location}

Table~\ref{tab:eval3_location_detailed} provides granular location alignment metrics for all category-model-dataset-level combinations, reporting ROUGE-1, ROUGE-2, ROUGE-L, METEOR, and BERT F1 scores. These metrics assess how accurately models identify the precise text spans containing discrepancies, with higher scores indicating better localization capability. This fine-grained analysis reveals nuanced performance patterns obscured by aggregated results presented in the main paper.

Three key insights emerge from the detailed breakdown. First, category difficulty varies systematically: Ambiguity and Inconsistency perturbations generally yield higher scores than Structural Flaws and certain Omissions, suggesting that models more readily identify discrepancies with explicit textual markers versus those involving absence or organizational defects. For example, Gemini-2.0 achieves 92.3 ROUGE-1 on Ambiguity$_{Legal}$ (CUAD, L1) but only 63.6 on Structural Flaws$_{Legal}$, a 28.7-point gap. Second, dataset effects are category-dependent: while most categories show higher scores on NLI than CUAD, Ambiguity$_{inText}$ exhibits the opposite pattern, with models performing 10-15 points better on CUAD across all architectures. This suggests that CUAD contracts contain more explicit definitional sections that facilitate detection of ambiguous terms, whereas NLI's clause-level structure makes in-text ambiguities harder to isolate. Third, one-shot prompting (L2) produces inconsistent effects: for some configurations (e.g., GPT-4o-mini on Ambiguity categories), L2 improves scores by 2-5 points, but for others (e.g., Gemini-2.5 on CUAD Omission$_{Legal}$), L2 actually degrades performance, likely because the example biases the model toward a specific localization pattern that does not generalize.

The BERT F1 scores, which measure semantic similarity rather than lexical overlap, reveal that models achieve surprisingly high alignment (>90\%) even when ROUGE scores are modest (60-70\%). This indicates that models often correctly identify the semantic region containing the discrepancy but struggle with exact span boundaries, a critical distinction for practical deployment, where approximate localization may suffice for human review but precise extraction is needed for automated processing. The consistent superiority of Gemini-2.0 across nearly all configurations (typically 5-10 points above other models) suggests architecture-specific strengths in span prediction, potentially related to its training on document-level tasks. Conversely, LLaMa-3.3's relatively uniform scores across categories (typically 60-65 on ROUGE-1 for inText categories) indicate a less differentiated understanding of perturbation types, treating diverse categories more homogeneously than other models.

\begin{table*}[p]
\centering
\scriptsize
\setlength{\tabcolsep}{3pt}
\renewcommand{\arraystretch}{1.15}
\begin{tabular}{l|l|ccccc|ccccc|ccccc|ccccc}
\hline
\multirow{3}{*}{\textbf{Category}} & \multirow{3}{*}{\textbf{Data}} & \multicolumn{10}{c}{\textbf{GPT-4o-m}} & \multicolumn{10}{c}{\textbf{Gem-2.0}} \\
\cline{3-22}
  &   & \multicolumn{5}{c|}{\textbf{L1}} & \multicolumn{5}{c|}{\textbf{L2}} & \multicolumn{5}{c|}{\textbf{L1}} & \multicolumn{5}{c|}{\textbf{L2}} \\
\cline{3-22}
  &   & \textbf{R$_1$} & \textbf{R$_2$} & \textbf{R$_L$} & \textbf{M} & \textbf{F1} & \textbf{R$_1$} & \textbf{R$_2$} & \textbf{R$_L$} & \textbf{M} & \textbf{F1} & \textbf{R$_1$} & \textbf{R$_2$} & \textbf{R$_L$} & \textbf{M} & \textbf{F1} & \textbf{R$_1$} & \textbf{R$_2$} & \textbf{R$_L$} & \textbf{M} & \textbf{F1} \\
\hline
\multirow{2}{*}{Amb$_{\text{Leg}}$} & CUAD & 83.9 & 83.4 & 83.9 & 77.6 & 97.2 & 81.7 & 81.2 & 81.6 & 74.9 & 96.7 & 92.3 & 92.0 & 90.3 & 90.9 & 98.3 & 90.9 & 90.6 & 89.1 & 89.5 & 98.0 \\
 & NLI & 87.7 & 87.3 & 86.9 & 83.4 & 97.8 & 85.8 & 85.3 & 85.2 & 81.6 & 97.5 & 96.6 & 96.4 & 94.5 & 95.8 & 99.1 & 95.0 & 94.8 & 93.5 & 93.4 & 98.9 \\
\hline
\multirow{2}{*}{Amb$_{\text{inT}}$} & CUAD & 52.5 & 51.5 & 51.7 & 41.5 & 91.7 & 56.3 & 55.5 & 55.1 & 45.2 & 92.4 & 69.7 & 68.6 & 68.6 & 61.1 & 94.5 & 69.0 & 68.1 & 68.1 & 61.1 & 94.5 \\
 & NLI & 59.7 & 58.6 & 58.7 & 51.1 & 92.8 & 61.8 & 60.5 & 60.2 & 53.2 & 93.1 & 66.8 & 65.7 & 65.9 & 61.5 & 93.6 & 69.5 & 68.2 & 68.2 & 63.6 & 93.8 \\
\hline
\multirow{2}{*}{Incon$_{\text{Leg}}$} & CUAD & 85.1 & 84.5 & 83.7 & 81.0 & 97.1 & 83.5 & 83.0 & 82.8 & 79.1 & 96.8 & 85.4 & 85.1 & 84.9 & 85.1 & 97.2 & 85.2 & 84.6 & 84.5 & 85.0 & 97.1 \\
 & NLI & 85.2 & 84.5 & 84.6 & 80.4 & 97.1 & 85.5 & 84.9 & 84.4 & 80.5 & 97.2 & 88.6 & 87.9 & 87.3 & 86.6 & 97.6 & 88.0 & 87.4 & 87.1 & 85.4 & 97.6 \\
\hline
\multirow{2}{*}{Incon$_{\text{inT}}$} & CUAD & 50.7 & 49.9 & 50.1 & 39.2 & 91.5 & 54.4 & 53.4 & 53.2 & 43.5 & 92.1 & 64.1 & 63.1 & 63.2 & 55.4 & 93.6 & 64.2 & 63.3 & 63.3 & 55.1 & 93.6 \\
 & NLI & 55.1 & 54.2 & 54.4 & 45.1 & 92.2 & 57.7 & 56.7 & 56.9 & 47.9 & 92.7 & 67.0 & 65.2 & 65.9 & 60.2 & 93.2 & 66.1 & 64.5 & 65.2 & 59.1 & 93.2 \\
\hline
\multirow{2}{*}{MisTerm$_{\text{Leg}}$} & CUAD & 77.9 & 77.2 & 76.8 & 70.8 & 95.9 & 77.1 & 76.4 & 75.8 & 70.0 & 95.7 & 87.1 & 86.6 & 85.9 & 84.3 & 97.4 & 86.1 & 85.6 & 85.0 & 83.1 & 97.2 \\
 & NLI & 82.5 & 81.5 & 81.2 & 76.9 & 96.6 & 83.3 & 82.5 & 80.9 & 77.9 & 96.6 & 89.1 & 88.6 & 87.9 & 87.0 & 97.7 & 89.0 & 88.6 & 87.6 & 86.4 & 97.7 \\
\hline
\multirow{2}{*}{MisTerm$_{\text{inT}}$} & CUAD & 54.0 & 53.3 & 53.2 & 43.8 & 91.7 & 59.1 & 58.3 & 58.3 & 48.2 & 92.8 & 68.5 & 67.4 & 67.2 & 61.6 & 94.2 & 68.4 & 67.2 & 67.0 & 60.6 & 94.0 \\
 & NLI & 58.6 & 57.8 & 57.7 & 49.5 & 92.9 & 62.4 & 61.5 & 61.8 & 54.1 & 93.3 & 68.3 & 66.6 & 66.8 & 63.9 & 93.5 & 69.5 & 67.7 & 67.7 & 64.5 & 93.8 \\
\hline
\multirow{2}{*}{Omis$_{\text{Leg}}$} & CUAD & 81.4 & 80.8 & 80.8 & 75.0 & 96.4 & 79.5 & 78.7 & 79.1 & 72.7 & 95.9 & 85.0 & 84.3 & 83.4 & 83.1 & 96.9 & 86.4 & 85.9 & 85.3 & 85.2 & 97.2 \\
 & NLI & 91.4 & 91.1 & 90.9 & 88.3 & 98.3 & 90.1 & 89.7 & 89.8 & 86.7 & 98.1 & 92.5 & 92.1 & 91.8 & 92.3 & 98.3 & 92.5 & 92.1 & 91.8 & 91.5 & 98.4 \\
\hline
\multirow{2}{*}{Omis$_{\text{inT}}$} & CUAD & 55.3 & 54.4 & 55.0 & 44.4 & 92.2 & 58.7 & 57.6 & 58.0 & 48.9 & 92.7 & 69.6 & 68.4 & 68.6 & 63.2 & 94.3 & 69.5 & 68.0 & 68.4 & 63.5 & 94.3 \\
 & NLI & 62.2 & 60.9 & 61.2 & 54.3 & 93.1 & 62.2 & 60.9 & 61.5 & 56.4 & 93.2 & 67.8 & 65.3 & 66.3 & 66.2 & 92.9 & 68.5 & 66.3 & 67.1 & 65.6 & 93.2 \\
\hline
\multirow{2}{*}{StrFlaw$_{\text{Leg}}$} & CUAD & 47.6 & 46.9 & 46.3 & 34.6 & 90.3 & 39.8 & 39.2 & 39.6 & 27.7 & 88.9 & 63.6 & 62.1 & 59.8 & 52.2 & 92.5 & 64.1 & 62.6 & 60.5 & 51.6 & 92.6 \\
 & NLI & 56.8 & 54.9 & 54.4 & 45.1 & 91.8 & 57.6 & 55.7 & 56.1 & 45.6 & 91.9 & 68.4 & 65.8 & 65.5 & 65.9 & 93.6 & 69.7 & 67.7 & 66.8 & 65.2 & 93.8 \\
\hline
\multirow{2}{*}{StrFlaw$_{\text{inT}}$} & CUAD & 48.4 & 47.5 & 47.7 & 37.6 & 91.0 & 53.2 & 52.4 & 52.4 & 42.6 & 91.9 & 62.3 & 61.4 & 61.6 & 52.2 & 93.4 & 60.9 & 60.0 & 60.2 & 50.5 & 93.2 \\
 & NLI & 57.2 & 56.3 & 56.9 & 47.1 & 92.7 & 59.7 & 58.6 & 59.2 & 49.9 & 93.0 & 65.9 & 64.0 & 64.8 & 58.1 & 93.3 & 65.7 & 63.9 & 64.4 & 57.6 & 93.3 \\
\hline
\multirow{3}{*}{\textbf{Category}} & \multirow{3}{*}{\textbf{Data}} & \multicolumn{10}{c|}{\textbf{Gem-2.5}} & \multicolumn{10}{c|}{\textbf{LLaMa-3.3}} \\
\cline{3-22}
  &   & \multicolumn{5}{c|}{\textbf{L1}} & \multicolumn{5}{c|}{\textbf{L2}} & \multicolumn{5}{c|}{\textbf{L1}} & \multicolumn{5}{c|}{\textbf{L2}} \\
\cline{3-22}
  &   & \textbf{R$_1$} & \textbf{R$_2$} & \textbf{R$_L$} & \textbf{M} & \textbf{F1} & \textbf{R$_1$} & \textbf{R$_2$} & \textbf{R$_L$} & \textbf{M} & \textbf{F1} & \textbf{R$_1$} & \textbf{R$_2$} & \textbf{R$_L$} & \textbf{M} & \textbf{F1} & \textbf{R$_1$} & \textbf{R$_2$} & \textbf{R$_L$} & \textbf{M} & \textbf{F1} \\
\hline
\multirow{2}{*}{Amb$_{\text{Leg}}$} & CUAD & 89.4 & 88.9 & 88.4 & 89.7 & 98.0 & 90.5 & 90.1 & 89.2 & 90.2 & 98.2 & 87.3 & 86.8 & 86.0 & 83.3 & 97.4 & 87.6 & 87.2 & 86.3 & 83.7 & 97.7 \\
 & NLI & 93.1 & 92.7 & 91.4 & 93.1 & 98.6 & 93.5 & 93.1 & 91.6 & 92.9 & 98.7 & 91.2 & 90.9 & 89.7 & 88.8 & 98.2 & 92.9 & 92.6 & 90.0 & 89.9 & 98.6 \\
\hline
\multirow{2}{*}{Amb$_{\text{inT}}$} & CUAD & 62.0 & 59.8 & 59.6 & 53.9 & 92.7 & 61.8 & 59.4 & 59.9 & 53.9 & 92.7 & 61.6 & 60.4 & 60.7 & 51.6 & 93.2 & 60.9 & 59.7 & 59.6 & 50.6 & 93.0 \\
 & NLI & 63.7 & 60.9 & 61.0 & 58.1 & 92.4 & 66.2 & 63.5 & 63.6 & 61.4 & 92.9 & 63.9 & 62.6 & 62.4 & 56.0 & 93.5 & 62.2 & 70.1 & 65.4 & 61.3 & 66.1 \\
\hline
\multirow{2}{*}{Incon$_{\text{Leg}}$} & CUAD & 79.5 & 78.7 & 78.7 & 79.0 & 96.2 & 80.8 & 80.0 & 80.0 & 78.9 & 96.3 & 86.1 & 85.5 & 84.2 & 83.7 & 97.1 & 86.4 & 85.6 & 85.0 & 84.6 & 97.1 \\
 & NLI & 85.4 & 84.5 & 84.3 & 84.4 & 97.0 & 85.4 & 84.4 & 83.7 & 84.2 & 97.0 & 86.3 & 85.4 & 84.3 & 84.7 & 97.1 & 85.9 & 85.2 & 84.5 & 82.6 & 97.2 \\
\hline
\multirow{2}{*}{Incon$_{\text{inT}}$} & CUAD & 58.2 & 56.0 & 56.0 & 48.6 & 92.0 & 58.2 & 55.7 & 56.1 & 48.9 & 92.0 & 58.7 & 57.5 & 57.8 & 48.0 & 92.7 & 59.1 & 58.0 & 57.9 & 48.4 & 92.8 \\
 & NLI & 64.3 & 61.7 & 62.0 & 58.5 & 92.5 & 65.2 & 62.3 & 62.8 & 59.9 & 92.5 & 64.0 & 62.6 & 63.4 & 55.5 & 93.5 & 62.5 & 61.2 & 61.8 & 53.4 & 93.3 \\
\hline
\multirow{2}{*}{MisTerm$_{\text{Leg}}$} & CUAD & 82.1 & 81.2 & 80.8 & 80.0 & 96.4 & 49.5 & 56.7 & 54.2 & 48.8 & 49.5 & 84.6 & 84.0 & 82.7 & 79.2 & 97.0 & 84.2 & 83.4 & 81.9 & 78.7 & 96.9 \\
 & NLI & 83.2 & 82.3 & 81.9 & 82.3 & 96.8 & 83.7 & 83.0 & 82.1 & 81.5 & 96.8 & 88.7 & 87.9 & 86.6 & 85.9 & 97.5 & 86.6 & 85.8 & 83.9 & 83.0 & 97.2 \\
\hline
\multirow{2}{*}{MisTerm$_{\text{inT}}$} & CUAD & 62.3 & 59.9 & 59.9 & 54.1 & 92.6 & 62.6 & 60.7 & 60.9 & 55.5 & 92.8 & 58.2 & 56.9 & 56.9 & 47.5 & 92.5 & 57.8 & 56.7 & 56.9 & 46.7 & 92.5 \\
 & NLI & 66.1 & 63.7 & 64.0 & 60.8 & 93.1 & 66.8 & 64.3 & 64.2 & 63.6 & 93.0 & 62.1 & 60.4 & 61.3 & 55.0 & 93.0 & 62.2 & 60.8 & 61.0 & 53.4 & 93.2 \\
\hline
\multirow{2}{*}{Omis$_{\text{Leg}}$} & CUAD & 77.3 & 75.9 & 75.8 & 74.9 & 95.6 & 55.2 & 56.8 & 59.2 & 60.1 & 57.5 & 82.5 & 81.9 & 80.4 & 78.4 & 96.5 & 82.5 & 81.8 & 81.0 & 78.4 & 96.5 \\
 & NLI & 89.3 & 88.6 & 88.5 & 90.1 & 97.9 & 89.0 & 88.6 & 88.4 & 89.2 & 97.9 & 89.2 & 88.6 & 87.9 & 89.1 & 97.8 & 92.0 & 91.6 & 90.7 & 90.9 & 98.3 \\
\hline
\multirow{2}{*}{Omis$_{\text{inT}}$} & CUAD & 62.0 & 59.7 & 59.5 & 55.4 & 92.5 & 63.9 & 61.2 & 61.8 & 58.2 & 92.8 & 61.4 & 60.1 & 60.3 & 51.8 & 93.2 & 60.9 & 59.6 & 59.9 & 50.7 & 93.1 \\
 & NLI & 62.0 & 59.0 & 60.0 & 61.1 & 92.2 & 61.7 & 58.3 & 59.1 & 61.1 & 92.0 & 64.7 & 63.0 & 63.6 & 58.0 & 93.5 & 65.9 & 64.4 & 64.5 & 57.8 & 93.8 \\
\hline
\multirow{2}{*}{StrFlaw$_{\text{Leg}}$} & CUAD & 63.1 & 61.1 & 58.8 & 51.5 & 92.2 & 65.4 & 61.1 & 57.8 & 52.7 & 47.7 & 52.1 & 51.1 & 48.5 & 38.6 & 91.0 & 51.8 & 50.7 & 48.9 & 38.5 & 91.0 \\
 & NLI & 65.4 & 63.2 & 62.1 & 60.8 & 93.1 & 65.8 & 63.3 & 62.9 & 61.2 & 93.0 & 61.7 & 59.2 & 56.8 & 51.7 & 92.2 & 60.8 & 59.4 & 56.3 & 50.1 & 92.6 \\
\hline
\multirow{2}{*}{StrFlaw$_{\text{inT}}$} & CUAD & 57.6 & 55.2 & 55.7 & 47.5 & 92.0 & 57.9 & 55.7 & 56.3 & 48.7 & 92.0 & 54.3 & 53.2 & 53.5 & 43.4 & 92.1 & 56.0 & 54.8 & 55.1 & 44.9 & 92.4 \\
 & NLI & 64.5 & 61.5 & 61.4 & 58.4 & 92.6 & 64.7 & 61.6 & 62.2 & 58.2 & 92.7 & 62.0 & 60.3 & 61.0 & 53.2 & 93.2 & 62.5 & 61.1 & 61.5 & 53.3 & 93.4 \\
\hline
\end{tabular}
\caption{\small Detailed Location Alignment Metrics (Eval 3) for all categories, models, and datasets. Category abbreviations: Amb=Ambiguity, Incon=Inconsistencies, MisTerm=Misaligned Terminology, Omis=Omission, StrFlaw=Structural Flaws, $_{\text{Leg}}$=Legal, $_{\text{inT}}$=inText. Metrics: ROUGE-1 (R$_1$), ROUGE-2 (R$_2$), ROUGE-L (R$_L$), METEOR (M), BERT F1 Score (F1). Models: GPT-4o-m=GPT-4o mini, Gem-2.0=Gemini 2.0, Gem-2.5=Gemini 2.5, LLaMa-3.3=LLaMa 3.3. Levels: L1 (zero-shot), L2 (one-shot). All values shown as percentages.}
\label{tab:eval3_location_detailed}
\end{table*}

\subsection{Evaluation 3: Explanation Quality Metrics by Category}
\label{sec:appendix-eval3-explanation}

Table~\ref{tab:eval3_explanation_quality} presents LLM-judge-based quality assessments of generated explanations across four dimensions: Accuracy (factual correctness of the identified issue), Clarity (ease of understanding), Completeness (coverage of relevant legal points), and Legal Reasoning (soundness of legal logic). Each metric ranges from 0-5, with scores evaluated by both GPT-4o and Gemini-2.5 judges to control for evaluator bias. This table reveals striking patterns in explanation generation capabilities that differ substantially from detection performance. The most pronounced finding is the universal weakness in Completeness: across all models, datasets, and categories, Completeness scores rarely exceed 3.0 and often fall below 2.0, while Clarity scores consistently reach 4.0+. This dissociation indicates that models generate fluent, well-structured explanations that nonetheless omit critical legal context—they articulate \textit{what} the discrepancy is but fail to fully explain \textit{why} it matters legally or \textit{what specific risks} it creates. For instance, GPT-4o-mini achieves Clarity of 4.2 but Completeness of only 1.8 (Gemini-2.5 judge, NLI Ambiguity$_{inText}$, L1), suggesting surface-level coherence without substantive depth. This pattern appears consistent with models' tendency to generate confident but shallow legal analysis, a known limitation in domain-specific reasoning tasks. Judge variability reveals systematic evaluator biases: Gemini-2.5 judges consistently assign lower Completeness and Accuracy scores than GPT-4o judges (often by 0.5-1.0 points), particularly for GPT-generated explanations. This suggests Gemini judges apply stricter standards or that cross-architecture evaluation introduces systematic bias. However, the \textit{relative ranking} of models remains consistent across judges: Gemini-2.5 (the evaluated model) consistently achieves the highest Legal Reasoning scores, followed by Gemini-2.0, then GPT-4o-mini and LLaMa-3.3. This ranking holds regardless of which judge evaluates, lending confidence to the robustness of the performance hierarchy. Category-specific patterns reveal that legal contradictions generally elicit higher-quality explanations than in-text contradictions: Legal Reasoning scores for Ambiguity$_{Legal}$ average 0.5-1.0 points higher than Ambiguity$_{inText}$ across all models. This likely reflects that legal violations require explicit citation of violated statutes, which provides additional structure and grounding to explanations, whereas in-text contradictions demand more abstract reasoning about internal consistency without external anchors. The Omission category shows particularly weak performance across all metrics, with Completeness scores often below 1.5, indicating that models struggle to articulate what is missing and why its absence creates legal risk—a fundamentally harder task than critiquing present text. The minimal improvement from L1 to L2 (typically <0.3 points across all metrics) suggests that explanation quality is primarily limited by models' intrinsic legal reasoning capacity rather than prompt engineering, marking a clear bottleneck for practical deployment.

\section{Contractual Pitfalls: Real-World Case Examples}
\label{sec:contractual_pitfalls}

Table~\ref{tab:contractual_pitfalls} documents real-world U.S. legal cases illustrating the five core perturbation types in our benchmark. These cases demonstrate that the contradictions in our dataset are not merely theoretical constructs but reflect genuine failure modes that have led to litigation, substantial financial penalties, and precedent-setting rulings. Each entry connects a perturbation category to a specific case, detailing the contractual error, the legal dispute it triggered, and the resulting consequences. This grounding in case law validates our taxonomy and underscores the practical significance of evaluating LLM robustness to such flaws.

The selected cases span multiple domains—entertainment licensing, construction, financial services, insurance, and consumer contracts—illustrating that contractual pitfalls are pervasive across industries. The \textit{Perini} case exemplifies Omission vulnerabilities: a missing consequential damages waiver resulted in a \$14.5 million judgment, far exceeding the contract's value. The \textit{Coinbase} case demonstrates Inconsistency risks: conflicting arbitration clauses across related agreements required Supreme Court intervention to determine enforceability. The \textit{Fetch Interactive} case highlights Ambiguity: a cure period stated as ``fifteen (30) days'' created direct textual contradiction, leading to disputed obligations. These examples collectively show that even single-word changes (``shall'' vs. ``should'') or missing clauses can fundamentally alter enforceability, liability exposure, and contractual obligations—precisely the risks our benchmark is designed to stress-test in LLMs deployed for legal contract review.

\begin{sidewaystable*}[h!]
\centering
\scriptsize
\renewcommand{\arraystretch}{1.1}
\setlength{\tabcolsep}{2pt}
\begin{tabular}{|l|l|cccc|cccc|cccc|cccc|cccc|cccc|cccc|cccc|}
\hline
\multirow{3}{*}{\textbf{Category}} & \multirow{3}{*}{\textbf{Judge}} &
\multicolumn{8}{c|}{\textbf{GPT-4o mini}} &
\multicolumn{8}{c|}{\textbf{Gemini 2.0}} &
\multicolumn{8}{c|}{\textbf{Gemini 2.5}} &
\multicolumn{8}{c|}{\textbf{LLaMa 3.3}} \\
\cline{3-34}
 & & \multicolumn{4}{c|}{\textbf{L1}} &
  \multicolumn{4}{c|}{\textbf{L2}} &
  \multicolumn{4}{c|}{\textbf{L1}} &
  \multicolumn{4}{c|}{\textbf{L2}} &
  \multicolumn{4}{c|}{\textbf{L1}} &
  \multicolumn{4}{c|}{\textbf{L2}} &
  \multicolumn{4}{c|}{\textbf{L1}} &
  \multicolumn{4}{c|}{\textbf{L2}} \\
\cline{3-34}
 & & \textbf{Acc} & \textbf{Cla} & \textbf{Com} & \textbf{LR} & \textbf{Acc} & \textbf{Cla} & \textbf{Com} & \textbf{LR} & \textbf{Acc} & \textbf{Cla} & \textbf{Com} & \textbf{LR} & \textbf{Acc} & \textbf{Cla} & \textbf{Com} & \textbf{LR} & \textbf{Acc} & \textbf{Cla} & \textbf{Com} & \textbf{LR} & \textbf{Acc} & \textbf{Cla} & \textbf{Com} & \textbf{LR} & \textbf{Acc} & \textbf{Cla} & \textbf{Com} & \textbf{LR} & \textbf{Acc} & \textbf{Cla} & \textbf{Com} & \textbf{LR} \\
\hline
\rowcolor{lightgray}
\multicolumn{34}{|c|}{\textbf{CUAD}} \\
\hline
\multirow{2}{*}{Amb$_{\text{Leg}}$} & \textbf{GPT4o} & 3.3 & 4.2 & 2.5 & 3.2 & 3.3 & 4.2 & 2.6 & 3.3 & 4.3 & 4.8 & 3.6 & 4.1 & 4.2 & 4.8 & 3.6 & 4.0 & 4.1 & 4.7 & 3.6 & 4.1 & 4.3 & 4.8 & 3.9 & 4.3 & 3.0 & 3.9 & 2.3 & 2.9 & 3.0 & 3.9 & 2.4 & 3.0 \\
& \textbf{Gem2.5} & 3.5 & 4.6 & 1.8 & 3.2 & 3.5 & 4.7 & 2.0 & 3.4 & 4.4 & 4.9 & 3.2 & 4.2 & 4.3 & 4.9 & 3.2 & 4.2 & 3.8 & 4.7 & 3.2 & 4.0 & 4.1 & 4.8 & 3.6 & 4.4 & 3.2 & 4.5 & 1.8 & 2.7 & 2.8 & 4.2 & 1.8 & 2.7 \\
\hline
\multirow{2}{*}{Amb$_{\text{inT}}$} & \textbf{GPT4o} & 2.5 & 3.6 & 1.8 & 2.5 & 3.0 & 4.0 & 2.4 & 3.1 & 3.1 & 4.0 & 2.4 & 3.0 & 3.0 & 3.9 & 2.4 & 3.0 & 3.0 & 3.9 & 2.5 & 3.1 & 2.8 & 3.7 & 2.5 & 2.9 & 2.4 & 3.5 & 1.8 & 2.4 & 2.9 & 3.8 & 2.3 & 2.8 \\
& \textbf{Gem2.5} & 2.0 & 3.8 & 1.2 & 2.2 & 2.7 & 4.1 & 1.9 & 2.8 & 2.6 & 4.3 & 1.8 & 2.9 & 2.6 & 4.1 & 1.8 & 2.6 & 2.4 & 4.3 & 1.8 & 3.3 & 2.3 & 3.9 & 1.9 & 2.9 & 2.1 & 4.1 & 1.2 & 2.2 & 2.7 & 4.1 & 1.8 & 2.6 \\
\hline
\multirow{2}{*}{Incon$_{\text{Leg}}$} & \textbf{GPT4o} & 3.1 & 4.0 & 2.5 & 3.0 & 3.1 & 4.1 & 2.7 & 3.1 & 3.3 & 4.2 & 2.9 & 3.3 & 3.5 & 4.3 & 3.0 & 3.4 & 3.2 & 4.1 & 2.8 & 3.2 & 3.3 & 4.2 & 3.1 & 3.5 & 2.7 & 3.8 & 2.2 & 2.6 & 3.1 & 4.0 & 2.5 & 3.0 \\
& \textbf{Gem2.5} & 2.5 & 4.3 & 1.9 & 2.4 & 2.8 & 4.5 & 2.1 & 2.6 & 2.9 & 4.5 & 2.3 & 2.8 & 3.0 & 4.6 & 2.3 & 2.7 & 2.6 & 4.4 & 2.2 & 2.8 & 2.8 & 4.6 & 2.4 & 3.1 & 2.3 & 4.3 & 1.7 & 2.0 & 2.3 & 4.2 & 1.7 & 2.2 \\
\hline
\multirow{2}{*}{Incon$_{\text{inT}}$} & \textbf{GPT4o} & 2.4 & 3.6 & 1.9 & 2.5 & 3.3 & 4.1 & 2.7 & 3.3 & 2.8 & 3.8 & 2.3 & 2.8 & 2.9 & 3.8 & 2.5 & 2.8 & 2.7 & 3.9 & 2.3 & 2.9 & 2.8 & 3.7 & 2.5 & 2.9 & 2.2 & 3.4 & 1.7 & 2.2 & 3.2 & 4.1 & 2.8 & 3.1 \\
& \textbf{Gem2.5} & 1.7 & 3.8 & 1.1 & 2.1 & 2.8 & 4.1 & 2.0 & 2.9 & 2.2 & 4.2 & 1.4 & 2.5 & 2.3 & 3.8 & 1.8 & 2.4 & 2.2 & 4.3 & 1.7 & 3.1 & 2.1 & 3.8 & 2.0 & 2.9 & 1.7 & 3.9 & 1.0 & 1.8 & 3.1 & 4.4 & 2.4 & 2.8 \\
\hline
\multirow{2}{*}{MisTerm$_{\text{Leg}}$} & \textbf{GPT4o} & 3.7 & 4.5 & 3.1 & 3.7 & 3.7 & 4.5 & 3.1 & 3.7 & 4.0 & 4.6 & 3.6 & 4.0 & 4.1 & 4.7 & 3.7 & 4.1 & 3.9 & 4.5 & 3.6 & 4.0 & 3.9 & 4.5 & 3.6 & 4.1 & 3.1 & 4.0 & 2.6 & 3.0 & 2.3 & 3.5 & 1.8 & 2.4 \\
& \textbf{Gem2.5} & 3.6 & 4.6 & 2.7 & 3.5 & 3.6 & 4.6 & 2.7 & 3.5 & 4.0 & 4.8 & 3.2 & 3.9 & 4.0 & 4.8 & 3.3 & 3.8 & 3.5 & 4.6 & 3.1 & 3.8 & 3.5 & 4.6 & 3.2 & 3.8 & 3.0 & 4.4 & 2.1 & 2.7 & 1.9 & 4.0 & 1.2 & 2.0 \\
\hline
\multirow{2}{*}{MisTerm$_{\text{inT}}$} & \textbf{GPT4o} & 1.9 & 3.3 & 1.4 & 2.0 & 2.5 & 3.7 & 1.9 & 2.6 & 2.4 & 3.6 & 1.9 & 2.5 & 2.2 & 3.3 & 1.8 & 2.2 & 2.6 & 3.8 & 2.2 & 2.7 & 2.5 & 3.6 & 2.2 & 2.6 & 1.7 & 3.1 & 1.2 & 1.9 & 2.3 & 3.5 & 1.8 & 2.3 \\
& \textbf{Gem2.5} & 1.4 & 3.8 & 0.8 & 1.8 & 2.1 & 4.1 & 1.4 & 2.4 & 2.0 & 4.1 & 1.3 & 2.6 & 1.8 & 3.8 & 1.3 & 1.9 & 2.2 & 4.2 & 1.7 & 3.1 & 2.1 & 4.0 & 1.9 & 2.8 & 1.4 & 3.9 & 0.7 & 1.7 & 1.9 & 4.0 & 1.2 & 1.7 \\
\hline
\multirow{2}{*}{Omission$_{\text{Leg}}$} & \textbf{GPT4o} & 2.6 & 3.7 & 2.1 & 2.7 & 2.5 & 3.7 & 1.9 & 2.5 & 2.9 & 3.9 & 2.4 & 3.0 & 2.7 & 3.8 & 2.2 & 2.7 & 2.8 & 3.9 & 2.3 & 2.9 & 2.9 & 4.0 & 2.4 & 3.0 & 2.4 & 3.5 & 1.8 & 2.4 & 1.8 & 3.1 & 1.2 & 1.8 \\
& \textbf{Gem2.5} & 2.2 & 4.1 & 1.3 & 2.0 & 2.0 & 4.2 & 1.3 & 1.9 & 2.5 & 4.4 & 1.6 & 2.5 & 2.3 & 4.4 & 1.4 & 2.3 & 2.2 & 4.4 & 1.5 & 2.6 & 2.5 & 4.4 & 1.7 & 2.7 & 2.0 & 4.0 & 1.1 & 1.7 & 1.1 & 3.6 & 0.5 & 1.1 \\
\hline
\multirow{2}{*}{Omis$_{\text{inT}}$} & \textbf{GPT4o} & 1.6 & 3.1 & 1.1 & 1.8 & 1.7 & 3.0 & 1.1 & 1.8 & 2.0 & 3.4 & 1.5 & 2.1 & 1.5 & 2.7 & 1.0 & 1.4 & 2.2 & 3.6 & 1.8 & 2.4 & 1.4 & 2.8 & 1.0 & 1.5 & 1.8 & 3.1 & 1.2 & 1.9 & 1.6 & 3.0 & 1.2 & 1.6 \\
& \textbf{Gem2.5} & 0.9 & 3.5 & 0.5 & 1.3 & 1.0 & 3.4 & 0.5 & 1.1 & 1.5 & 4.0 & 0.8 & 2.1 & 0.9 & 3.4 & 0.5 & 1.1 & 1.6 & 4.1 & 1.1 & 2.5 & 0.8 & 3.6 & 0.5 & 1.3 & 1.3 & 3.9 & 0.6 & 1.5 & 0.9 & 3.5 & 0.5 & 1.0 \\
\hline
\multirow{2}{*}{StrFlaws$_{\text{Leg}}$} & \textbf{GPT4o} & 1.7 & 3.1 & 1.3 & 1.9 & 1.9 & 3.2 & 1.5 & 1.9 & 2.0 & 3.3 & 1.5 & 2.1 & 2.0 & 3.3 & 1.5 & 2.0 & 1.9 & 3.3 & 1.5 & 2.0 & 2.1 & 3.3 & 1.7 & 2.1 & 1.7 & 3.1 & 1.2 & 1.8 & 3.3 & 4.1 & 2.8 & 3.3 \\
& \textbf{Gem2.5} & 1.0 & 4.0 & 0.5 & 1.5 & 1.1 & 3.7 & 0.8 & 1.4 & 1.3 & 4.1 & 0.8 & 2.0 & 1.4 & 3.9 & 0.9 & 1.6 & 1.3 & 4.0 & 0.9 & 1.8 & 1.3 & 4.0 & 1.1 & 1.7 & 2.4 & 4.1 & 1.5 & 2.4 & 3.1 & 4.2 & 2.1 & 3.0 \\
\hline
\multirow{2}{*}{StrFlaws$_{\text{inT}}$} & \textbf{GPT4o} & 2.9 & 3.9 & 2.3 & 3.0 & 3.6 & 4.3 & 2.9 & 3.6 & 3.3 & 4.1 & 2.7 & 3.3 & 3.4 & 4.1 & 2.8 & 3.3 & 3.2 & 4.1 & 2.8 & 3.4 & 3.2 & 3.9 & 2.9 & 3.3 & 2.6 & 3.7 & 2.1 & 2.6 & 3.2 & 4.6 & 2.0 & 2.9 \\
& \textbf{Gem2.5} & 2.4 & 3.9 & 1.7 & 2.6 & 3.2 & 4.3 & 2.4 & 3.3 & 2.8 & 4.3 & 1.9 & 3.1 & 3.0 & 4.1 & 2.3 & 3.0 & 2.5 & 4.3 & 2.0 & 3.5 & 2.6 & 4.0 & 2.5 & 3.4 & 3.1 & 4.0 & 2.5 & 2.9 & 3.2 & 4.1 & 2.6 & 3.2 \\
\hline
\rowcolor{lightgray}
\multicolumn{34}{|c|}{\textbf{NLI}} \\
\hline
\multirow{2}{*}{Amb$_{\text{Leg}}$} & \textbf{GPT4o} & 3.3 & 4.2 & 2.6 & 3.3 & 3.3 & 4.2 & 2.6 & 3.2 & 4.2 & 4.8 & 3.6 & 4.0 & 4.3 & 4.8 & 3.6 & 4.2 & 4.3 & 4.8 & 3.9 & 4.3 & 4.3 & 4.8 & 3.9 & 4.2 & 3.1 & 4.0 & 2.5 & 2.9 & 3.3 & 4.1 & 2.8 & 3.3 \\
& \textbf{Gem2.5} & 3.5 & 4.7 & 2.0 & 3.4 & 3.6 & 4.7 & 1.9 & 3.5 & 4.3 & 4.9 & 3.2 & 4.2 & 4.4 & 4.9 & 3.2 & 4.3 & 4.1 & 4.8 & 3.6 & 4.4 & 4.2 & 4.9 & 3.6 & 4.4 & 3.2 & 4.6 & 2.0 & 2.9 & 3.6 & 4.7 & 2.3 & 3.4 \\
\hline
\multirow{2}{*}{Amb$_{\text{inT}}$} & \textbf{GPT4o} & 3.0 & 4.0 & 2.4 & 3.1 & 3.0 & 3.9 & 2.3 & 2.9 & 3.0 & 3.9 & 2.4 & 3.0 & 3.1 & 4.0 & 2.6 & 3.0 & 2.8 & 3.7 & 2.5 & 2.9 & 2.8 & 3.7 & 2.6 & 2.9 & 3.0 & 3.9 & 2.4 & 3.0 & 2.6 & 3.1 & 3.6 & 1.3 \\
& \textbf{Gem2.5} & 2.7 & 4.1 & 1.9 & 2.8 & 2.8 & 4.0 & 1.8 & 2.5 & 2.6 & 4.1 & 1.8 & 2.6 & 2.7 & 4.1 & 1.9 & 2.8 & 2.3 & 3.9 & 1.9 & 2.9 & 2.2 & 3.9 & 1.9 & 2.9 & 2.8 & 4.2 & 1.8 & 2.7 & 3.2 & 4.6 & 1.5 & 2.0 \\
\hline
\multirow{2}{*}{Incon$_{\text{Leg}}$} & \textbf{GPT4o} & 3.1 & 4.1 & 2.7 & 3.1 & 3.4 & 4.3 & 2.9 & 3.3 & 3.5 & 4.3 & 3.0 & 3.4 & 3.6 & 4.4 & 3.1 & 3.5 & 3.3 & 4.2 & 3.1 & 3.5 & 3.4 & 4.2 & 3.1 & 3.5 & 2.9 & 3.8 & 2.3 & 2.8 & 3.0 & 4.0 & 2.5 & 3.0 \\
& \textbf{Gem2.5} & 2.8 & 4.5 & 2.1 & 2.6 & 3.2 & 4.6 & 2.3 & 2.9 & 3.0 & 4.6 & 2.3 & 2.7 & 3.1 & 4.6 & 2.4 & 2.9 & 2.8 & 4.6 & 2.4 & 3.1 & 2.9 & 4.6 & 2.5 & 3.1 & 2.7 & 4.1 & 1.8 & 2.6 & 2.7 & 4.3 & 2.1 & 2.6 \\
\hline
\multirow{2}{*}{Incon$_{\text{inT}}$} & \textbf{GPT4o} & 3.3 & 4.1 & 2.7 & 3.3 & 3.1 & 4.0 & 2.4 & 3.0 & 2.9 & 3.8 & 2.5 & 2.8 & 3.1 & 3.9 & 2.6 & 3.0 & 2.8 & 3.7 & 2.5 & 2.9 & 2.9 & 3.8 & 2.6 & 3.0 & 3.1 & 4.0 & 2.5 & 3.0 & 3.0 & 3.9 & 2.3 & 2.9 \\
& \textbf{Gem2.5} & 2.8 & 4.1 & 2.0 & 2.9 & 2.7 & 4.0 & 1.8 & 2.5 & 2.3 & 3.8 & 1.8 & 2.4 & 2.4 & 3.9 & 1.8 & 2.5 & 2.1 & 3.8 & 2.0 & 2.9 & 2.2 & 3.8 & 2.0 & 2.9 & 2.3 & 4.2 & 1.7 & 2.2 & 2.7 & 4.0 & 1.7 & 2.5 \\
\hline
\multirow{2}{*}{MisTerm$_{\text{Leg}}$} & \textbf{GPT4o} & 3.7 & 4.5 & 3.1 & 3.7 & 3.8 & 4.4 & 3.2 & 3.7 & 4.1 & 4.7 & 3.7 & 4.1 & 4.2 & 4.7 & 3.8 & 4.2 & 3.9 & 4.5 & 3.6 & 4.1 & 4.0 & 4.6 & 3.6 & 4.1 & 3.2 & 4.1 & 2.8 & 3.1 & 3.4 & 4.2 & 2.9 & 3.3 \\
& \textbf{Gem2.5} & 3.6 & 4.6 & 2.7 & 3.5 & 3.7 & 4.7 & 2.9 & 3.7 & 4.0 & 4.8 & 3.3 & 3.8 & 4.2 & 4.8 & 3.5 & 4.1 & 3.5 & 4.6 & 3.2 & 3.8 & 3.7 & 4.7 & 3.3 & 4.0 & 3.1 & 4.4 & 2.4 & 2.8 & 3.4 & 4.5 & 2.7 & 3.3 \\
\hline
\multirow{2}{*}{MisTerm$_{\text{inT}}$} & \textbf{GPT4o} & 2.5 & 3.7 & 1.9 & 2.6 & 2.1 & 3.4 & 1.6 & 2.2 & 2.2 & 3.3 & 1.8 & 2.2 & 2.2 & 3.3 & 1.8 & 2.2 & 2.5 & 3.6 & 2.2 & 2.6 & 2.5 & 3.5 & 2.2 & 2.6 & 2.3 & 3.5 & 1.8 & 2.4 & 2.1 & 3.4 & 1.5 & 2.2 \\
& \textbf{Gem2.5} & 2.1 & 4.1 & 1.4 & 2.4 & 1.7 & 3.7 & 1.1 & 2.0 & 1.8 & 3.8 & 1.3 & 1.9 & 1.9 & 3.8 & 1.3 & 1.9 & 2.1 & 4.0 & 1.9 & 2.8 & 2.0 & 3.8 & 1.8 & 2.6 & 1.9 & 4.0 & 1.2 & 2.0 & 1.7 & 3.7 & 1.0 & 1.8 \\
\hline
\multirow{2}{*}{Omis$_{\text{Leg}}$} & \textbf{GPT4o} & 2.5 & 3.7 & 1.9 & 2.5 & 2.7 & 3.8 & 2.1 & 2.7 & 2.7 & 3.8 & 2.2 & 2.7 & 2.9 & 4.0 & 2.4 & 2.9 & 2.9 & 4.0 & 2.4 & 3.0 & 2.9 & 4.0 & 2.4 & 3.1 & 2.3 & 3.5 & 1.8 & 2.3 & 2.4 & 3.6 & 1.8 & 2.4 \\
& \textbf{Gem2.5} & 2.0 & 4.2 & 1.3 & 1.9 & 2.3 & 4.2 & 1.5 & 2.2 & 2.3 & 4.4 & 1.4 & 2.3 & 2.4 & 4.4 & 1.6 & 2.5 & 2.5 & 4.4 & 1.7 & 2.7 & 2.5 & 4.4 & 1.8 & 2.7 & 1.9 & 4.0 & 1.2 & 1.7 & 2.0 & 4.1 & 1.2 & 1.9 \\
\hline
\multirow{2}{*}{Omis$_{\text{inT}}$} & \textbf{GPT4o} & 1.7 & 3.0 & 1.1 & 1.8 & 1.6 & 2.9 & 1.1 & 1.6 & 1.5 & 2.7 & 1.0 & 1.4 & 1.6 & 2.9 & 1.2 & 1.6 & 1.4 & 2.8 & 1.0 & 1.5 & 1.5 & 2.8 & 1.1 & 1.5 & 1.8 & 3.1 & 1.2 & 1.8 & 1.8 & 3.1 & 1.2 & 1.9 \\
& \textbf{Gem2.5} & 1.0 & 3.4 & 0.5 & 1.1 & 1.0 & 3.3 & 0.4 & 1.0 & 0.9 & 3.4 & 0.5 & 1.1 & 1.1 & 3.3 & 0.5 & 1.1 & 0.8 & 3.6 & 0.5 & 1.3 & 1.0 & 3.3 & 0.7 & 1.3 & 1.1 & 3.6 & 0.5 & 1.1 & 1.3 & 3.5 & 0.5 & 1.4 \\
\hline
\multirow{2}{*}{StrFlaw$_{\text{Leg}}$} & \textbf{GPT4o} & 1.9 & 3.2 & 1.5 & 1.9 & 1.9 & 3.2 & 1.5 & 2.0 & 2.0 & 3.3 & 1.5 & 2.0 & 2.2 & 3.5 & 1.7 & 2.3 & 2.1 & 3.3 & 1.7 & 2.1 & 2.0 & 3.3 & 1.5 & 2.0 & 1.6 & 3.0 & 1.2 & 1.6 & 1.7 & 3.1 & 1.3 & 1.8 \\
& \textbf{Gem2.5} & 1.1 & 3.7 & 0.8 & 1.4 & 1.1 & 3.7 & 0.8 & 1.3 & 1.4 & 3.9 & 0.9 & 1.6 & 1.4 & 3.9 & 1.0 & 1.8 & 1.3 & 4.0 & 1.1 & 1.7 & 1.3 & 3.8 & 1.0 & 1.6 & 0.9 & 3.5 & 0.5 & 1.0 & 0.9 & 3.7 & 0.5 & 1.1 \\
\hline
\multirow{2}{*}{StrFlaw$_{\text{inT}}$} & \textbf{GPT4o} & 3.6 & 4.3 & 2.9 & 3.6 & 3.3 & 4.1 & 2.6 & 3.2 & 3.4 & 4.1 & 2.8 & 3.3 & 3.3 & 4.2 & 2.9 & 3.3 & 3.2 & 3.9 & 2.9 & 3.3 & 3.2 & 3.9 & 2.9 & 3.2 & 3.3 & 4.1 & 2.8 & 3.3 & 3.3 & 4.1 & 2.6 & 3.3 \\
& \textbf{Gem2.5} & 3.2 & 4.3 & 2.4 & 3.3 & 2.9 & 4.0 & 2.1 & 2.9 & 3.0 & 4.1 & 2.3 & 3.0 & 3.0 & 4.1 & 2.2 & 3.0 & 2.6 & 4.0 & 2.5 & 3.4 & 2.5 & 3.9 & 2.3 & 3.3 & 3.1 & 4.2 & 2.1 & 3.0 & 3.2 & 4.2 & 2.1 & 3.1 \\
\hline
\end{tabular}
\caption{\small Explanation quality evaluation across models, datasets, and judges. Metrics: Accuracy (Acc), Clarity (Cla), Completeness (Com), Legal Reasoning (LR). Models: GPT-4o mini, Gemini 2.0, Gemini 2.5, LLaMa 3.3. Levels: L1 (zero-shot), L2 (one-shot). Judges: GPT4o, Gemini 2.5 (Gem2.5). All values on a scale of 0-5.}
\label{tab:eval3_explanation_quality}
\end{sidewaystable*}

\begin{table*}[h!]
\centering
\small
\renewcommand{\arraystretch}{2.0}
\setlength{\tabcolsep}{3pt}
\begin{tabular}{|>{\centering\arraybackslash}m{1.8cm}|>{\centering\arraybackslash}m{3.8cm}|>{\centering\arraybackslash}m{2.8cm}|>{\centering\arraybackslash}m{3.5cm}|>{\centering\arraybackslash}m{3cm}|}
\hline
\textbf{Pitfall} & \textbf{Description} & \textbf{U.S. Case} & \textbf{Contractual Error} & \textbf{Consequence} \\
\hline
\rowcolor{lightestgray}
\textbf{Ambiguity} & Vague, unclear, or contradictory language creating uncertainty in interpretation, making contract terms susceptible to multiple conflicting meanings. & \textit{Fetch Interactive Television, LLC v. Touchstream Technologies, Inc.} & Cure period stated as ``fifteen (30)'' days, creating direct textual contradiction and ambiguity in timeline. & Disputed deadlines, unexpected obligations based on court interpretation. \\
\hline
\textbf{Omission} & Deliberate removal or absence of critical information, clauses, or terms necessary for complete understanding or legal enforceability. & \textit{Perini Corp. v. Greate Bay Hotel \& Casino, Inc.} & Construction contract lacked waiver of consequential damages or clear limitation of liability. & Exposure to unforeseen catastrophic claims for consequential or punitive damages (multi-million dollar award). \\
\hline
\rowcolor{lightestgray}
\textbf{Inconsistency} & Direct contradictions between sections where statements, obligations, or terms are mutually exclusive or logically incompatible. & \textit{Coinbase, Inc. v. Suski} & Two agreements between same parties contained conflicting dispute resolution clauses---one delegating to arbitrator, other to courts. & Uncertainty, breach claims, difficulty determining true obligation, requiring court intervention. \\
\hline
\textbf{Misaligned Terminology} & Inconsistent use of defined terms, or terminology conflicting with established legal definitions or industry standards. & \textit{Lithko Contracting, LLC, et al. v. XL Insurance America, Inc.} & Waiver clause used ``no party shall be liable to another party'' but inconsistently referred to ``Tenant [Amazon]'' separately, creating ambiguity. & Critical clauses may not cover intended parties, leading to disputes over scope of protection. \\
\hline
\rowcolor{lightestgray}
\textbf{Structural Flaws} & Modifications disrupting logical organization, hierarchy, or cross-referencing, creating navigational confusion or broken dependencies. & \textit{Carnival Cruise Lines, Inc. v. Shute} & Forum selection clause placement and communication scrutinized for fundamental fairness to determine enforceability. & Legally significant clauses may be unenforceable if not reasonably communicated or if placement is unfair; uncertainty regarding prevailing terms. \\
\hline
\end{tabular}
\caption{\small Real-world U.S. case examples illustrating the five core perturbation types. Each case demonstrates how contractual flaws lead to disputes and adverse legal consequences.}
\label{tab:contractual_pitfalls}
\end{table*}

\section{Human Evaluation Guidelines and Rubric}
\label{sec:human-evaluation-rubric}

This section provides the evaluation guidelines used by expert annotators to validate contradiction quality in \datasetName. All evaluators were NLP researchers with legal document analysis experience who reviewed perturbations using structured binary classification. The rubric was designed to ensure consistent assessment of contradiction strength, legal relevance, and contextual coherence across all perturbation types.

\subsection{Evaluation Task Overview}

For each perturbation instance, evaluators receive:
\begin{itemize}[leftmargin=*,itemsep=0.1em]
    \item \textbf{Perturbation Type}: One of ten categories (Ambiguity, Inconsistency, Structural Flaw, Misaligned Terminology, Omission—each with Legal and InText variants)
    \item \textbf{Original Text}: The unmodified contract excerpt
    \item \textbf{Changed Text}: The perturbed contract excerpt
    \item \textbf{Explanation}: Model-generated justification for why this constitutes a contradiction
    \item \textbf{Contradiction Type}: Whether this is a legal (external law violation) or in-text (internal document inconsistency) contradiction
    \item \textbf{Additional Context}: For legal contradictions, includes law citation, scraped law text, and law explanation; for in-text contradictions, includes contradicted text location and full contradicted section
\end{itemize}

\textbf{Evaluation Goal}: Determine whether the modification creates a \textbf{strong, meaningful, and unambiguous contradiction} that would reasonably cause legal uncertainty, enforceability issues, or compliance violations.

\subsection{Binary Classification Criteria}

Evaluators assign a binary label:

\paragraph{YES (Strong Contradiction Exists)} Assign \textbf{YES} if the perturbation satisfies \textbf{all} of the following:
\begin{enumerate}[leftmargin=*,itemsep=0.1em]
    \item \textbf{Clarity}: The contradiction is immediately recognizable to a reasonably competent legal practitioner without requiring extensive legal research or interpretation
    \item \textbf{Strength}: The modification directly conflicts with, negates, or undermines a legal requirement (for legal contradictions) or creates mutually exclusive obligations (for in-text contradictions)
    \item \textbf{Significance}: The contradiction would plausibly lead to:
    \begin{itemize}
        \item Enforceability disputes or litigation
        \item Regulatory non-compliance or penalties
        \item Contractual ambiguity requiring judicial interpretation
        \item Material breach of obligations or rights
    \end{itemize}
    \item \textbf{Coherence}: The perturbation maintains linguistic naturalness and does not introduce obvious artifacts that would immediately reveal it as artificially generated
\end{enumerate}

\paragraph{NO (Insufficient Contradiction)} Assign \textbf{NO} if \textbf{any} of the following apply:
\begin{enumerate}[leftmargin=*,itemsep=0.1em]
    \item \textbf{Weak or Arguable}: The contradiction is subtle, speculative, or would require expert legal opinion to establish definitively
    \item \textbf{Interpretable}: Both versions can be reasonably reconciled through standard contract interpretation principles
    \item \textbf{Minor Modification}: The change is cosmetic, stylistic, or adds/removes non-essential details without affecting legal obligations
    \item \textbf{Context-Dependent}: The contradiction only manifests under specific factual scenarios not clearly established in the contract
    \item \textbf{Unnatural Language}: The perturbation introduces grammatical errors, nonsensical phrasing, or clearly artificial modifications that compromise linguistic quality
\end{enumerate}

\subsection{Category-Specific Guidance}

\paragraph{Legal Contradictions} For perturbations violating external legal standards:
\begin{itemize}[leftmargin=*,itemsep=0.1em]
    \item \textbf{Verify Law Applicability}: Check that the cited law/statute is genuinely applicable to the contract's jurisdiction and subject matter
    \item \textbf{Assess Violation Directness}: Ensure the modification directly violates the law rather than merely reducing compliance clarity
    \item \textbf{Consider Enforceability Impact}: Evaluate whether the violation would render the clause unenforceable or expose parties to legal liability
    \item \textbf{Example of YES}: Changing ``Employees shall receive overtime pay at 1.5× regular rate for hours exceeding 40 per week'' to ``Employees may receive overtime compensation as determined by management'' violates FLSA mandatory overtime provisions
    \item \textbf{Example of NO}: Changing ``The company will comply with all applicable employment laws'' to ``The company shall make reasonable efforts to comply with applicable employment laws'' weakens the commitment but does not create clear non-compliance
\end{itemize}

\paragraph{In-Text Contradictions} For perturbations creating internal document conflicts:
\begin{itemize}[leftmargin=*,itemsep=0.1em]
    \item \textbf{Verify Cross-Reference Impact}: Confirm that the modification actually conflicts with specific text elsewhere in the document (not just theoretically)
    \item \textbf{Assess Mutual Exclusivity}: Ensure the conflicting clauses cannot both be true or enforced simultaneously
    \item \textbf{Consider Materiality}: Evaluate whether the contradiction affects substantive rights, obligations, or financial terms (not just procedural details)
    \item \textbf{Example of YES}: Section 2 defines ``Net Revenue'' as ``gross revenue minus operating expenses,'' while Section 7 calculates royalties as ``15\% of Net Revenue (gross revenue only)''
    \item \textbf{Example of NO}: Section 3 refers to ``the Effective Date'' while Section 9 refers to ``the Commencement Date,'' where both terms clearly reference the same date in context
\end{itemize}

\subsection{Handling Edge Cases}

\paragraph{When in Doubt} If uncertain whether a contradiction meets the threshold for YES:
\begin{itemize}[leftmargin=*,itemsep=0.1em]
    \item Default to \textbf{NO} to maintain high benchmark quality
    \item Consider: ``Would a practicing attorney immediately flag this in contract review?''
    \item Prioritize false negatives over false positives—excluding borderline cases preserves benchmark integrity
\end{itemize}

\paragraph{Disagreement Resolution} When evaluators disagree:
\begin{itemize}[leftmargin=*,itemsep=0.1em]
    \item Majority vote determines classification (3 evaluators per perturbation)
    \item Significant disagreement (2-1 split on multiple perturbations from same category) triggers category-level review
    \item Systematic disagreement patterns are documented and analyzed for evaluator calibration
\end{itemize}

\subsection{Quality Assurance Procedures}

To ensure evaluation consistency and reliability:
\begin{enumerate}[leftmargin=*,itemsep=0.1em]
    \item \textbf{Calibration Phase}: All evaluators independently assess 100 pilot perturbations, followed by group discussion to align interpretation of criteria
    \item \textbf{Inter-Rater Reliability}: Cohen's Kappa and Jaccard Coefficient computed pairwise and across all evaluators (see Table~\ref{tab:agreement} in Section 2)
    \item \textbf{Blind Evaluation}: Evaluators do not see each other's assessments or know the generating model to prevent bias
    \item \textbf{Stratified Sampling}: Perturbations are sampled proportionally across all categories, datasets, and generation conditions to ensure representative coverage
\end{enumerate}

\clearpage

\begin{figure*}[t]
\centering
\includegraphics[width=\textwidth]{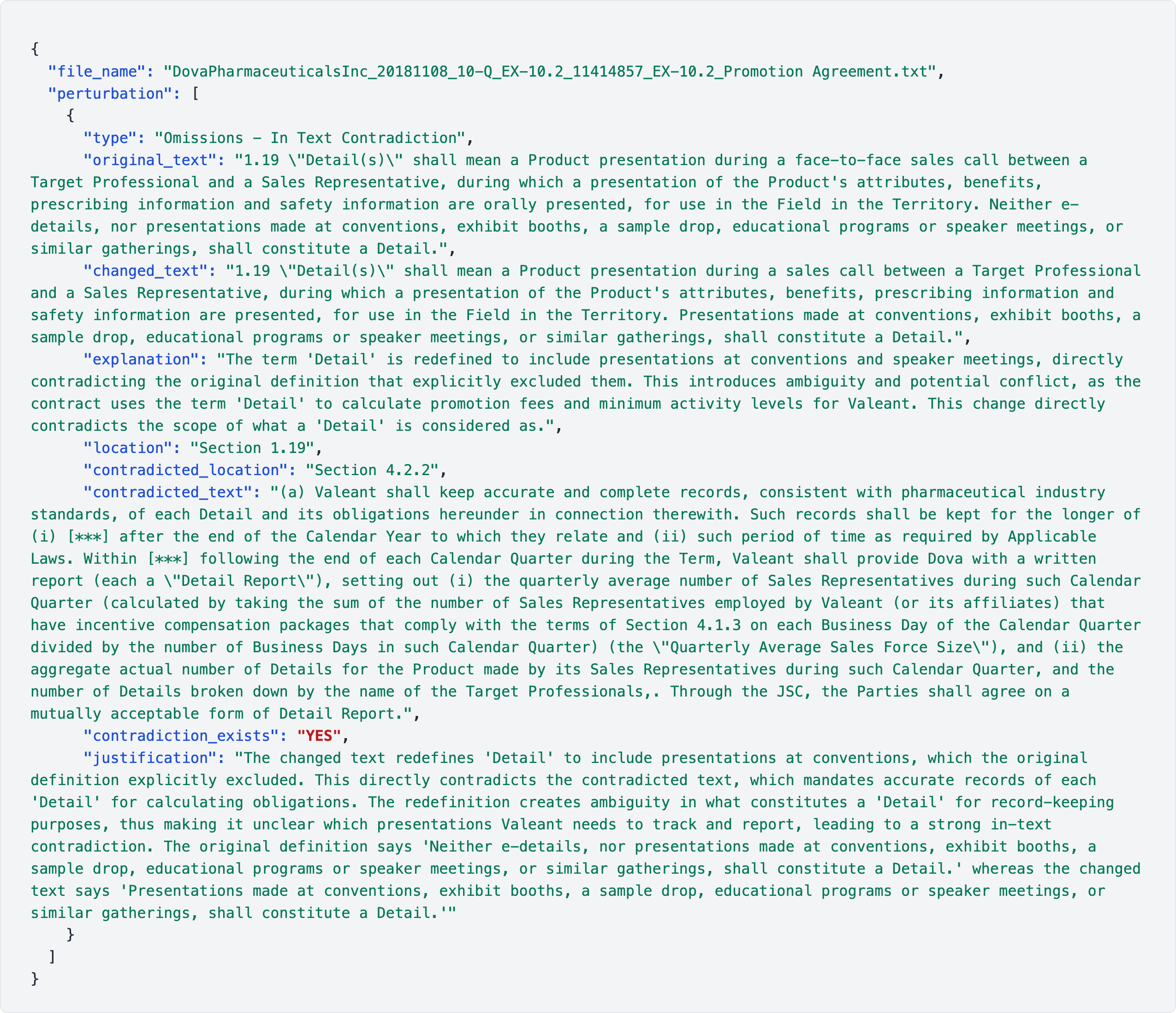}
\caption{\small Example JSON structure for in-text contradiction perturbations, showing the metadata schema used throughout the dataset.}
\label{fig:inText_JSON}
\end{figure*}

\begin{figure*}[t]
\centering
\includegraphics[width=\textwidth]{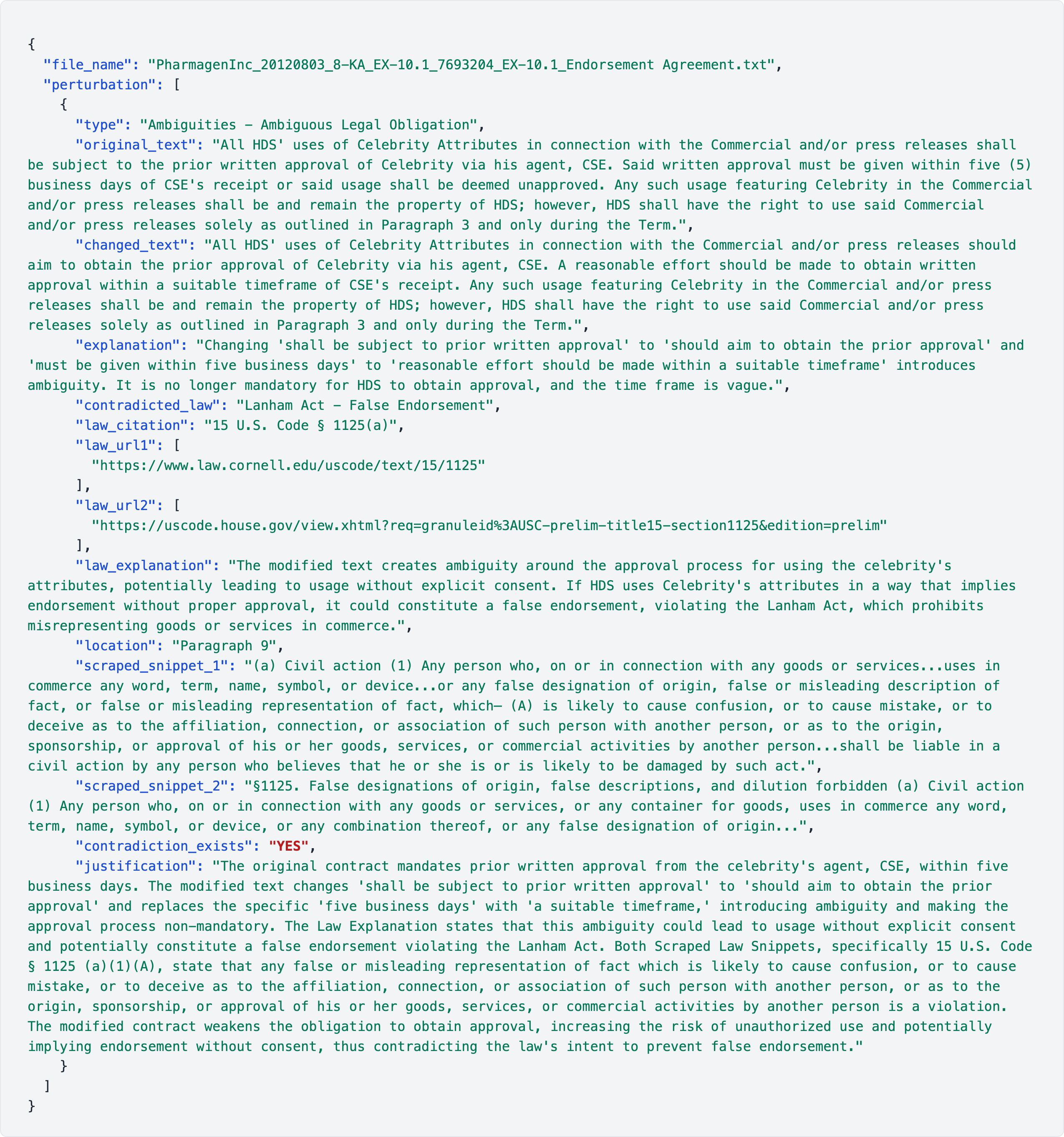}
\caption{\small Example JSON structure for legal contradiction perturbations, including law citations and statutory references.}
\label{fig:Legal_JSON}
\end{figure*}

\end{document}